\documentclass[journal,onecolumn,draftcls,12pt]{IEEEtran}

\usepackage{cite}

\ifCLASSINFOpdf
  \usepackage[pdftex]{graphicx}
  \graphicspath{{figures/}}
  \DeclareGraphicsExtensions{.pdf,.jpeg,.png}
\else
  \usepackage[dvips]{graphicx}
  \graphicspath{{figures/}}
  \DeclareGraphicsExtensions{.eps}
\fi

\usepackage{amsmath}
\usepackage{array}

\ifCLASSOPTIONcompsoc
  \usepackage[caption=false,font=normalsize,labelfont=sf,textfont=sf]{subfig}
\else
  \usepackage[caption=false,font=footnotesize]{subfig}
\fi

\usepackage{url}

\usepackage{threeparttable}
\usepackage{booktabs}
\usepackage{adjustbox}

\usepackage{bm} 
\usepackage{amsfonts}
\usepackage{amsmath}
\usepackage{amssymb}
\usepackage{amsthm}

\ifCLASSOPTIONcompsoc
  \usepackage[caption=false,font=normalsize,labelfont=sf,textfont=sf]{subfig}
\else
  \usepackage[caption=false,font=footnotesize]{subfig}
\fi

\usepackage{calc}

\usepackage{alphalph}

\usepackage{mathtools}

\DeclarePairedDelimiter\abs{\lvert}{\rvert}%
\DeclarePairedDelimiter\norm{\lVert}{\rVert}%

\newcommand{\ie}{\textit{i}.\textit{e}.,}
\newcommand{\eg}{\textit{e}.\textit{g}.,}

\begin{document}

\bstctlcite{IEEEexample:BSTcontrol}

\title{Incremental Cluster Validity Indices for Hard Partitions: Extensions and Comparative Study}

\author{Leonardo~Enzo~Brito~da~Silva,~\IEEEmembership{Member,~IEEE,} Niklas~M.~Melton,~\IEEEmembership{Member,~IEEE,}
Donald~C.~Wunsch~II,~\IEEEmembership{Fellow,~IEEE}%
\thanks{L. E. Brito da Silva is with the Applied Computational Intelligence Laboratory, Department of Electrical and Computer Engineering, Missouri University of Science and Technology, Rolla, MO 65409 USA, and also with the CAPES Foundation, Ministry of Education of Brazil, Bras\'{i}lia, DF 70040-020, Brazil (e-mail: leonardoenzo@ieee.org).}%
\thanks{N. M. Melton is with the Applied Computational Intelligence Laboratory, Department of Electrical and Computer Engineering, Missouri University of Science and Technology, Rolla, MO 65409 USA (e-mail: niklasmelton@ieee.org).}%
\thanks{D. C. Wunsch II is with the Applied Computational Intelligence Laboratory, Department of Electrical and Computer Engineering, Missouri University of Science and Technology, Rolla, MO 65409 USA (e-mail: wunsch@ieee.org).}%
}

\markboth{Preprint submitted to arXiv.org}{}

\maketitle

\begin{abstract}
Validation is one of the most important aspects of clustering, but most approaches have been batch methods. Recently, interest has grown in providing incremental alternatives. This paper extends the incremental cluster validity index (iCVI) family to include incremental versions of Calinski-Harabasz (iCH), I index and Pakhira-Bandyopadhyay-Maulik (iI and iPBM), Silhouette (iSIL), Negentropy Increment (iNI), Representative Cross Information Potential (irCIP) and Representative Cross Entropy (irH), and Conn\_Index (iConn\_Index). Additionally, the effect of under- and over-partitioning on the behavior of these six iCVIs, the Partition Separation (PS) index, as well as two other recently developed iCVIs (incremental Xie-Beni (iXB) and incremental Davies-Bouldin (iDB)) was examined through a comparative study. Experimental results using fuzzy adaptive resonance theory (ART)-based clustering methods showed that while evidence of most under-partitioning cases could be inferred from the behaviors of all these iCVIs, over-partitioning was found to be a more challenging scenario indicated only by the iConn\_Index. The expansion of incremental validity indices provides significant novel opportunities for assessing and interpreting the results of unsupervised learning. 
\end{abstract}


\IEEEpeerreviewmaketitle

\section{Introduction}  \label{Sec:intro}

\IEEEPARstart{C}{luster} validation~\cite{Gordon1998} is a critical topic in cluster analysis. It is crucial to assess the quality of the partitions detected by clustering algorithms when there is no class label information. Different clustering solutions may be found by distinct algorithms, or even by the same algorithm subjected to different hyper-parameters or a different input presentation order~\cite{xu2012,leonardo2018}. \textit{Cluster validity indices} (CVIs) perform the role of evaluators of such solutions. CVIs typically exhibit a trade-off between measures of compactness (within-cluster scatter) and isolation (between-cluster separation)~\cite{xu2012}. Numerous examples of such criteria have been presented in the literature; for comprehensive reviews and experimental studies the interested reader can go to~\cite{milligan1985, Bezdek1997, Halkidi2002a, Halkidi2002b, vendramin2010, Arbelaitz2013, xu2005, xu2009}. 

Recently, \textit{incremental cluster validity indices} (iCVIs) have been developed to track the effectiveness of online clustering methods over data streams~\cite{Moshtaghi2018, Moshtaghi2018b, Ibrahim2018, Keller2018}. To enable cluster validation in such applications, a recursive formulation of compactness was introduced in~\cite{Moshtaghi2018, Moshtaghi2018b}. This strategy has been used to develop incremental versions of four CVIs so far~\cite{Keller2018}: viz., incremental Davies-Bouldin (iDB)~\cite{Moshtaghi2018, Moshtaghi2018b}, incremental Xie-Beni (iXB)~\cite{Moshtaghi2018, Moshtaghi2018b} and modified Dunn's indices~\cite{Ibrahim2018b}. Particularly, the behavior of iXB and iDB are analyzed in both accurately and poorly partitioned data sets in~\cite{Moshtaghi2018, Moshtaghi2018b}, whereas the studies in~\cite{Ibrahim2018, Keller2018} only investigate the iDB's behavior in cases where online clustering algorithms accurately detect data structures, \ie~when they yield high performing experimental results.

Therefore, the contributions of this work are three-fold: (1) presenting incremental versions of six additional CVIs (thereby extending the family of iCVIs), (2) discussing the interpretation of these novel iCVIs in cases of accurately, under- and over-partitioning, and (3) performing a systematic comparative study among ten iCVIs. To explore such scenarios, fuzzy adaptive resonance theory (ART)-based clustering methods~\cite{Carpenter1991, Bartfai1994} were chosen for their simple parameterization of cluster granularity and other appealing properties~\cite{wunsch2009}.

The following, Section~\ref{Sec:theory}, provides a brief review of CVIs, iCVIs and ART; Section~\ref{Sec:method} presents this work's extensions of several other CVIs to the incremental family; Section~\ref{Sec:setup} details the set-up used in the numerical experiments; Section~\ref{Sec:results} describes and discusses the results; Section~\ref{Sec:inc_vs_batch} compares batch and incremental versions of CVIs; and Section~\ref{Sec:conclusion} summarizes this paper's findings.

\section{Background and related work}
\label{Sec:theory}

This section briefly recaps the theory regarding the CVIs, iCVIs and ART-based clustering algorithms used in this study. 

\subsection{Cluster Validity Indices (CVIs)}
\label{Sec:theory_CVI}

Consider a data set $\bm{X}=\{\bm{x}_i\}_{i=1}^N$ and its hard partition \mbox{$\Omega=\{\omega_i\}_{i=1}^k$} of $k$ disjointed clusters~$\omega_i$, such that \mbox{$\bigcup\limits_{i=1}^{k} \omega_i = \bm{X}$}. In the following CVI overview, $\bm{v}$~is a cluster prototype (centroid), $k$~is the number of clusters, $d$~is the dimensionality of the data ($ \bm{x}_i \in {\rm I\!R}^d$), $\| \cdot \|$ is the Euclidean norm, and $N$ and $n_i$ are the cardinalities of a data set and cluster $\omega_i$, respectively.

\subsubsection{Calinski-Harabasz (CH)~\cite{vrc}}

the CH index is defined as:
\begin{equation}
CH = \frac{BGSS/\left(k-1\right)}{WGSS/\left(N-k\right)},
\label{Eq:valind_ch1}
\end{equation}
\noindent where the between group sum of squares (BGSS) and within group sum of squares (WGSS) are computed as:
\begin{equation}
WGSS = \sum \limits_{i=1}^{k} \sum \limits_{
\substack{j=1 \\ \bm{x}_j\in \omega_i}}^{n_i} \| \bm{x}_j - \bm{v}_i \|^2,
\label{Eq:valind_ch2}
\end{equation}
\begin{equation}
BGSS = \sum \limits_{i=1}^{k} n_i \| \bm{v}_i - \bm{\mu}_{data} \|^2,
\label{Eq:valind_ch3}
\end{equation}
\begin{equation}
\bm{\mu}_{data} = \frac{1}{N} \sum\limits_{i=1}^N\bm{x}_i.
\label{Eq:valind_ch4}
\end{equation}
This is an optimization-like criterion~\cite{vendramin2010} such that larger values of CH indicate better clustering solutions (maximization).

\subsubsection{Davies-Bouldin (DB) ~\cite{db}}

the DB index averages the similarities $R$ of each cluster $i$ with respect to its maximally similar cluster $j \neq i$: 
\begin{equation}
DB = \frac{1}{k} \sum_{i=1}^k R_i,
\label{Eq:valind_db1}
\end{equation}
\noindent where
\begin{equation}
R_i = \max_{i \neq j}\left( \frac{S_i + S_j}{M_{i,j}} \right),
\label{Eq:valind_db2}
\end{equation}
\begin{equation}
S_l = \left[ \frac{1}{n_l} \sum \limits_{
\substack{m=1 \\ \bm{x}_m\in \omega_l}}^{n_l} \norm{\bm{x}_m - \bm{v}_l}^q \right]^{\frac{1}{q}},~l=\{1,...,k\},
\label{Eq:valind_db3}
\end{equation}
\begin{equation}
M_{i,j} = \left[\sum \limits_{t=1}^{d} \abs{v_{it} - v_{jt}}^p \right]^{\frac{1}{p}},~p \geq 1.
\label{Eq:valind_db4}
\end{equation}
The variables ($p$, $q$) are user-defined parameters, and $S_l$ and $M_{i,j}$ (Minkowski metric) measure compactness and separation, respectively. Smaller values of DB indicate better clustering solutions (minimization). 

\subsubsection{Xie-Beni (XB)~\cite{Xie1991}}

the XB index was originally designed to detect compact and separated clusters in fuzzy c-partitions. A hard partition version is given by the following ratio of compactness to separation~\cite{Lamirel2015, Lamirel2016}:
\begin{equation}
XB = \frac{WGSS / N}{\min\limits_{i \neq j} \| v_i - v_j \|^2 }.
\label{Eq:valind_xb1}
\end{equation}

Smaller values of XB indicate better clustering solutions (minimization). 

\subsubsection{Pakhira-Bandyopadhyay-Maulik (PBM)~\cite{Bandyopadhyay2001, pbm}}

consider the I~index~\cite{Bandyopadhyay2001} defined as:
\begin{equation}
I = \left(\frac{1}{k} \times \frac{E_1}{E_k} \times D_k \right)^p,~p \geq 1,
\label{Eq:valind_pbm1}
\end{equation}
\noindent where
\begin{equation}
E_1 = \sum\limits_{i=1}^N \norm{\bm{x}_i - \bm{\mu}_{data}},
\label{Eq:valind_pbm2}
\end{equation}
\begin{equation}
E_k = \sum\limits_{i=1}^k \sum \limits_{
\substack{j=1 \\ \bm{x}_j\in \omega_i}}^{n_i} \norm{\bm{x}_j - \bm{v}_i},
\label{Eq:valind_pbm3}
\end{equation}
\begin{equation}
D_k = \max_{i \neq j}\left( \norm{\bm{v}_i - \bm{v}_j} \right),
\label{Eq:valind_pbm4}
\end{equation}
\noindent The quantities $E_k$ and $D_k$ measure compactness and separation, respectively. This CVI comprises a trade-off among the three competing factors in Eq.~(\ref{Eq:valind_pbm1}): $\frac{1}{k}$ decreases with~$k$, whereas both $\frac{E_1}{E_k}$ and $D_k$ increase. By setting $p=2$ in Eq.~(\ref{Eq:valind_pbm1}), the I index reduces to the PBM index~\cite{pbm}. Larger values of PBM indicate better clustering solutions (maximization). 

\subsubsection{Silhouette (SIL)~\cite{sil}}

the SIL index is computed by averaging the silhouette coefficients $sc_i$ across all data samples $\bm{x}_i$:
\begin{equation}
SIL = \frac{1}{N}\sum_{i=1}^N sc_i,
\label{Eq:valind_sil1}
\end{equation}
\noindent where
\begin{equation}
sc_i= \frac{b_i-a_i}{\max\left(a_i, b_i\right)},
\label{Eq:valind_sil2}
\end{equation}
\begin{equation}
a_i = \frac{1}{n_i-1} \sum \limits_{\substack{j=1,j \neq i \\ \bm{x}_j \in \omega_i}}^{n_i} \| \bm{x}_j - \bm{x}_i \|,
\label{Eq:valind_sil3}
\end{equation}
\begin{equation}
b_i = \min\limits_{l, l \neq i}\left(\frac{1}{n_l}\sum \limits_{\substack{j=1 \\ \bm{x}_j \in \omega_l}}^{n_l} \| \bm{x}_j - \bm{x}_i \|\right),
\label{Eq:valind_sil4}
\end{equation}
\noindent the variables $a_i$ and $b_i$ measure compactness and separation, respectively. Larger values of SIL (close to 1) indicate better clustering solutions (maximization). To reduce computational complexity, some SIL variants, such as~\cite{Hruschka2004, Hruschka2006, Rawashdeh2012, Romera2016}, use a centroid-based approach. The simplified SIL~\cite{Hruschka2004, Hruschka2006} has been successfully used in clustering data streams processed in chunks, in which the silhouette coefficients are also used to make decisions regarding the centroids' incremental updates~\cite{Silva2016}.

\subsubsection{Partition Separation (PS)~\cite{Yang2001}} the PS index was originally developed for fuzzy clustering; its hard clustering version is given by~\cite{lughofer2008}:
\begin{equation}
PS = \sum\limits_{i=1}^{k} PS_i,
\label{Eq:valind_ps1}
\end{equation}
\noindent where
\begin{equation}
PS_i = \frac{n_i}{\max\limits_{j}(n_j)} - exp\left[ - \frac{\min\limits_{i \neq j} \left( \| \bm{v}_i - \bm{v}_j \|^2\right)}{\beta_T}  \right],
\label{Eq:valind_ps2}
\end{equation}
\begin{equation}
\beta_T = \frac{1}{k} \sum\limits_{l=1}^{k} \| \bm{v}_l - \bar{\bm{v}} \|^2,
\label{Eq:valind_ps3}
\end{equation}
\begin{equation}
\bar{\bm{v}} =\frac{1}{k} \sum\limits_{l=1}^{k} \bm{v}_l,
\label{Eq:valind_ps4}
\end{equation}

The PS index only comprises a measure of separation between prototypes. Therefore, this CVI can be readily used to evaluate the partitions identified by unsupervised incremental learners that model clusters using centroids (\eg~\cite{lughofer2008}). Larger values of PS indicate better clustering solutions (maximization).

\subsubsection{Negentropy Increment (NI)~\cite{fernandez2010,fernandez2009}}

the NI index measures the average normality of the clusters of a given partition $\Omega$ via negentropy~\cite{Comon1994} while avoiding the direct computation of the clusters' differential entropies. Unlike the other CVIs discussed so far, the NI is not explicitly constructed using measures of compactness and separation~\cite{fernandez2010,Arbelaitz2013}, thereby being defined as:
\begin{equation}
NI = \frac{1}{2}\sum\limits_{i=1}^{k} p_i\ln | \bm{\Sigma}_i | - \frac{1}{2} \ln |\bm{\Sigma}_{data}| -  \sum\limits_{i=1}^{k} p_i\ln p_i,
\label{Eq:valind_ni1}
\end{equation}
\noindent where $| \cdot |$ denotes the determinant. The probabilities ($p$), means ($\bm{v}$) and covariance matrices ($\bm{\Sigma}$) are estimated as:
\begin{equation}
p_i = \frac{n_i}{N},
\label{Eq:valind_ni2}
\end{equation}
\begin{equation}
\bm{v}_i =\frac{1}{n_i} \sum\limits_{
\substack{j=1 \\ \bm{x}_j\in \omega_i}}^{n_i} \bm{x}_j,
\label{Eq:valind_ni3}
\end{equation}
\begin{equation}
\bm{\Sigma}_i = \frac{1}{n_i-1} \sum\limits_{
\substack{j=1 \\ \bm{x}_j\in \omega_i}}^{n_i} (\bm{x}_j-\bm{v}_i)(\bm{x}_j-\bm{v}_i)^T,
\label{Eq:valind_ni4}
\end{equation}
\begin{equation}
\bm{\Sigma}_{data} = \frac{1}{N-1}\left( \bm{X}^T\bm{X} - N\bm{\mu}_{data} \bm{\mu}_{data} ^T\right),
\end{equation}
\noindent and $\bm{\mu}_{data}$ is estimated using Eq.~(\ref{Eq:valind_ch4}). Smaller values of NI indicate better clustering solutions (minimization).

\subsubsection{Representative Cross Information Potential (rCIP)~\cite{araujo20131, araujo20132}}
cluster evaluation functions (CEFs) based on cross information potential (CIP)~\cite{gokcay2000, gokcay2002} have been consistently used in the literature to evaluate partitions and drive optimization algorithms searching for data structure~\cite{gokcay2000, gokcay2002, araujo20131, araujo20132}, thus this work includes these CEFs under the CVI category. Precisely, representative approaches~\cite{araujo20131, araujo20132} replace the sample-by-sample estimation of Renyi's quadratic Entropy~\cite{renyi1961} using the Parzen-window method~\cite{duda2000} (original CIP~\cite{gokcay2000, gokcay2002}) via prototypes and the statistics of their associated Voronoi polyhedron. The rCIP was devised for prototype-based clustering (\ie~two-step methods: vector quantization followed by clustering of the prototypes)~\cite{Cottrell1997, Karypis1999, tyree1999, Vesanto2000, Ana2003}. The CEF used here is defined as~\cite{araujo20132}:
\begin{equation}
CEF = \sum\limits_{i=1}^{k-1} \sum\limits_{j=i+1}^k rCIP(\omega_i, \omega_j),
\label{Eq:valind_rCIP1}
\end{equation}
\noindent where
\begin{equation}
rCIP(\omega_i, \omega_j) = \frac{1}{M_i M_j} \sum\limits_{l=1}^{M_i}\sum\limits_{m=1}^{M_j}  G(\bm{v}_l - \bm{v}_m, \bm{\Sigma}_{l,m}),
\label{Eq:valind_rCIP2}
\end{equation}
\begin{equation}
G(\bm{v}_l - \bm{v}_m, \bm{\Sigma}_{l,m}) =  \frac{e^{  -\frac{1}{2} \left( \bm{v}_l - \bm{v}_m \right)^T \bm{\Sigma}_{l,m}^{-1} \left( \bm{v}_l - \bm{v}_m \right)}}{\sqrt{\left( 2 \pi \right)^{d} | \bm{\Sigma}_{l,m}| }},
\label{Eq:valind_rCIP3}
\end{equation}

\noindent $\bm{\Sigma}_{l,m}=\bm{\Sigma}_l + \bm{\Sigma}_m$, $\{\bm{v}_l,\Sigma_l\} \in \omega_i$, $\{\bm{v}_m,\Sigma_m\} \in \omega_j$, $M_i$ and $M_j$ are the number of prototypes used to represent clusters $\omega_i$ and $\omega_j$, respectively. The prototypes and covariance matrices are estimated using Eqs.~(\ref{Eq:valind_ni3}) and~(\ref{Eq:valind_ni4}), respectively. Smaller values of CEF indicate better clustering solutions (minimization). Recently, the information potential (IP)~\cite{principe2010} measure has been used to define a system's state when modeling and analyzing dynamic processes~\cite{Oliveira2017, Oliveira2018}. 

\subsubsection{Conn\_Index~\cite{tasdemir2007,tasdemir2011}}
the Conn\_Index was also developed for prototype-based clustering. It is formulated using the connectivity strength matrix (CONN), which is a symmetric square similarity matrix that represents local data densities between neighboring prototypes~\cite{tasdemir2006, tasdemir2009}. Its $(i,j)^{th}$ entry is formally given by:
\begin{equation}
CONN(i,j) = CADJ(i,j) + CADJ(j,i),
\label{Eq:valind_conn1}
\end{equation}
\noindent where the $(i,j)^{th}$ entry of the non-symmetric cumulative adjacency matrix (CADJ) corresponds to the number of samples for which $\bm{v}_i$ and $\bm{v}_j$ are, simultaneously, the first and second closest prototypes (according to some measure), respectively. The Conn\_Index is defined as:
\begin{equation}
Conn\_Index = Intra\_Conn \times \left( 1 - Inter\_Conn \right),
\label{Eq:valind_conn2}
\end{equation}
\noindent where the intra-cluster ($Intra\_Conn$) and inter-cluster ($Inter\_Conn$) connectivities are:
\begin{equation}
Intra\_Conn = \frac{1}{k}\sum\limits_{l=1}^k Intra\_Conn(\omega_l),
\label{Eq:valind_conn3}
\end{equation}
\begin{equation}
Intra\_Conn(\omega_l) =  \frac{1}{n_l} \sum\limits_{\substack{i,j \\ \bm{v}_i,\bm{v}_j \in \omega_l}}^{P} CADJ(i,j),
\label{Eq:valind_conn4}
\end{equation}
\begin{equation}
Inter\_Conn = \frac{1}{k} \sum\limits_{l=1}^k \max_{\substack{m \\ m \neq l}} \left[ Inter\_Conn(\omega_l,\omega_m) \right],
\label{Eq:valind_conn5}
\end{equation}
\begin{equation}
Inter\_Conn(\omega_l,\omega_m) =
\frac{\sum\limits_{\substack{i,j \\ \bm{v}_i \in \omega_l, \bm{v}_j \in \omega_m}}^P CONN(i,j)}{\sum\limits_{\substack{i,j \\ \bm{v}_i \in V_{l,m}}}^P CONN(i,j)},
\label{Eq:valind_conn7}
\end{equation}
\begin{equation}
V_{l,m} = \{ \bm{v}_i : \bm{v}_i  \in \omega_l, \exists \bm{v}_j \in \omega_m : CADJ(i,j)>0 \},
\label{Eq:valind_conn8}
\end{equation}
\noindent the variable $P$ is the total number of prototypes, and \mbox{$Inter\_Conn(\omega_l,\omega_m)=0$} if $V_{l,m} = \{ \emptyset \}$. Naturally, the quantities $Intra\_Conn$ and $Inter\_Conn$ measure compactness and separation, respectively. Larger values of the Conn\_Index (close to 1) indicate better clustering solutions (maximization).

\subsection{Incremental Cluster Validity Indices (iCVIs)}

The compactness and separation terms commonly found in CVIs are generally computed using data samples and prototypes, respectively~\cite{Moshtaghi2018, Ibrahim2018}. 
In order to handle online clustering applications demands (\ie~data streams), an incremental CVI (iCVI) formulation that recursively estimates the compactness term was introduced in~\cite{Moshtaghi2018, Moshtaghi2018b} in the context of fuzzy clustering.

Specifically, consider the hard clustering version of cluster~$i$'s compactness $CP$ (\ie~by setting the fuzzy memberships in~\cite{Moshtaghi2018, Moshtaghi2018b} to binary indicator functions): 
\begin{equation}
CP_i = \sum \limits_{
\substack{j=1 \\ \bm{x}_j\in \omega_i}}^{n_i} \| \bm{x}_j - \bm{v}_i \|^2,
\label{Eq:iCVI_0}
\end{equation}
in such a case, when a new sample $\bm{x}$ is presented and encoded by cluster~$i$, then its new compactness becomes: 
\begin{equation}
CP_i^{new} = \sum \limits_{
\substack{j=1 \\ \bm{x}_j\in \omega_i}}^{n_i^{new}} \| \bm{x}_j - \bm{v}_i^{new} \|^2,
\label{Eq:iCVI_1}
\end{equation}
\noindent where 
\begin{equation}
n_i^{new} = n_i^{old}+1,
\label{Eq:iCVI_1b}
\end{equation}
\begin{equation}
\bm{v}_i^{new} = \bm{v}_i^{old} + (\bm{x} - \bm{v}_i^{old})/n_i^{new},
\label{Eq:iCVI_1c}
\end{equation}
\noindent and
\begin{equation}
N^{new} = N^{old} + 1.
\label{Eq:nSamples}
\end{equation}

The compactness in Eq.~(\ref{Eq:iCVI_1}) can be updated incrementally as~\cite{Moshtaghi2018,Moshtaghi2018b}:
\setlength{\arraycolsep}{0.0em}
\begin{eqnarray}
CP_i^{new}&{}={}&CP_i^{old} + \| \bm{z}_i \|^2 + n_i^{old} \| \Delta \bm{v}_i \|^2 + 2\Delta \bm{v}_i^T\bm{g}_i^{old},
\label{Eq:iCVI_2}
\end{eqnarray}
\setlength{\arraycolsep}{5pt}
\noindent where
\begin{equation}
\bm{g}_i^{new} = \bm{g}_i^{old} + \bm{z}_i + n_i^{old} \Delta \bm{v}_i,
\label{Eq:iCVI_3}
\end{equation}
\begin{equation}
\bm{g}_i= \sum\limits_{j=1}^{n_i} \left( \bm{x}_j - \bm{v}_i \right),
\label{Eq:iCVI_4}
\end{equation}
\begin{equation}
\bm{z}_i=\bm{x} - \bm{v}_i^{new},
\label{Eq:iCVI_5}
\end{equation}
\begin{equation}
\Delta \bm{v}_i = \bm{v}_i^{old} - \bm{v}_i^{new}.
\label{Eq:iCVI_6}
\end{equation}

The compactness $CP$ and vector $\bm{g}$ are initialized as $0$ and~$\Vec{\bm{0}}$ (since $\bm{v} = \bm{x}$), respectively. Note that, at each iteration, the variable $\bm{g}$ is updated after $CP$.
Using such incremental formulation, the following iCVIs were derived in~\cite{Moshtaghi2018, Moshtaghi2018b} (their hard partition counterparts are shown here)
\subsubsection{incremental Xie-Beni (iXB)}
\begin{equation}
XB^{new} = \frac{1}{N^{new}} \times \frac{\sum \limits_{i=1}^{k^{new}} CP_i^{new}}{\min\limits_{i \neq j} \left( \| \bm{v}_i^{new} - \bm{v}_j^{new} \|^2 \right) } ,
\label{Eq:valind_xb}
\end{equation}
\subsubsection{incremental Davies-Bouldin (iDB - based on~\cite{Araki1993})}
\begin{equation}
DB^{new} = \frac{1}{k^{new}} \sum_{i=1}^{k^{new}} \max_{j, j \neq i}\left( \frac{\frac{CP_i^{new}}{n_i^{new}} + \frac{CP_j^{new}}{n_j^{new}}}{\| \bm{v}_i^{new} - \bm{v}_j^{new} \|^2} \right).
\label{Eq:valind_db}
\end{equation}

If a new cluster emerges, then $k^{new}= k^{old}+1$; otherwise its previous value is maintained. Note that only one prototype~$\bm{v}$ is updated after each input presentation.

\subsection{Adaptive Resonance Theory (ART)}

For this study's experiments, adaptive resonance theory (ART)~\cite{Carpenter1987} has been implemented. It is a fast and stable online clustering method with automatic category recognition encompassing a rich history with many implementations well-suited to iCVI computation ~\cite{Carpenter1987, Carpenter1987b, Carpenter1988, Carpenter1991, Carpenter1990c, xu2011, leonardo2018b, leonardo2018c, Chen1999, wunsch2009, Bartfai1994, williamson1996, anagnostopoulos2000, anagnostopoulos2001, vigdor2007, seiffertt2006, huang2014, Isawa2008, leonardo2017}. The following ART models were used in these experiments.

\subsubsection{Fuzzy ART~\cite{Carpenter1991}}
This model implements fuzzy logic~\cite{Zadeh1965} to bound data within hyper-boxes. For a normalized data set $\bm{X}=\{\bm{x}_i\}_{i=1}^N$ $( 0 \leq x_{i,j} \leq 1~,~ j=\{1,...,d\})$, the fuzzy ART algorithm, with parameters $(\alpha,\beta,\rho)$, is defined by:
\begin{equation}
\bm{I} = (\bm{x}_i,1-\bm{x}_i),
\label{Eq:FA_CC}
\end{equation}
\begin{equation}
{T}_j =  \frac{\|\min_{}(\bm{I},\bm{w}_j)\|_1}{\alpha + \|\bm{w}_j\|_1},
\label{Eq:FA_activation}
\end{equation}
\begin{equation}
\|\min_{}(\bm{I},\bm{w}_j)\|_1\geq \rho\|\bm{I}\|_1,
\label{Eq:FA_res}
\end{equation}
\begin{equation}
\bm{w}_j^{new} = \bm{w}_j^{old}(1-\beta)+\beta\min_{}(\bm{I},\bm{w}_j^{old}).
\label{Eq:FA_update}
\end{equation}

Equation~(\ref{Eq:FA_CC}) is the complement coding function, which concatenates sample $\bm{x}$ and its complement to form an input vector~$\bm{I}$ with dimension~$2d$. Equation~(\ref{Eq:FA_activation}) is the activation function for each category $j$, where $\| \cdot \|_1$ is the $L_1$ norm, $min(\cdot)$ is performed component-wise, and~$\alpha$ is a tie breaking constant. Each category is checked for validity against Eq.~(\ref{Eq:FA_res})'s vigilance parameter $\rho$ in a descending order of activation. If no valid category is found during training, then a new category is initialized using $\bm{I}$ as the new weight vector~$\bm{w}$. Otherwise, the winning category is updated according to Eq.~(\ref{Eq:FA_update}) using learning rate $\beta$. In this study, when fuzzy ART is set to evaluation mode (learning is disabled), if no valid category is found during search, then the winning category defaults to the highest activated one. 

\subsubsection{Fuzzy self-consistent modular ART (SMART)~\cite{Bartfai1994}} 
This model is a hierarchical clustering technique based on the ARTMAP architecture~\cite{Carpenter1991}. In an ARTMAP network, two ART modules, A- and B-side, are supplied with separate but dependent data streams. Both ART modules can cluster according to local topology and parameters while an inter-ART module enforces a surjective mapping of the A-side to the B-side, effectively learning the functional map of the A-side to the B-side categories. 

To build a fuzzy SMART module, it is only necessary to stream the same sample to both the A- and B-sides of a fuzzy ARTMAP module, \ie~use fuzzy ARTMAP in an auto-associative mode. If all else is equal in the A and B modules' parameters, fuzzy SMART will begin to form a two-level self-consistent cluster hierarchy when  $\rho_A > \rho_B$. This hierarchy will be required to extend the iCVI study to prototype-based CVIs such as the Conn\_Index. For such CVIs, the A-side categories act as cluster prototypes while the B-side provides the actual data partition.

\section{Extensions of iCVIs}  \label{Sec:method}

To compute the CVIs mentioned in Section~\ref{Sec:theory_CVI} incrementally, employing one of the following approaches is sufficient:
\begin{enumerate}
\item The recursive computation of compactness developed in~\cite{Moshtaghi2018, Moshtaghi2018b} (CVIs: CH, I/PBM, and SIL).
\item The incremental computation of probabilities, means and covariance matrices (CVIs: rCIP and NI). Naturally, if the clustering algorithm of choice already models the clusters using a priori probabilities, means and covariance matrices (such as Gaussian ART~\cite{williamson1996} and Bayesian ART~\cite{vigdor2007}), then, similarly to PS, these CVIs can be readily computed.
\item The incremental building of a multi-prototype representation of clusters in a self-consistent two-level hierarchy while tracking the density-based connections between neighboring prototypes (CVI: Conn\_index). Specifically, increment and/or expand the CADJ and CONN matrices as clusters grow and/or are dynamically created.     
\end{enumerate}

In the following iCVIs' extensions (iCH, iI/iPBM, iSIL, irCIP, iNI, and iConn\_index), if a new cluster is formed after sample $\bm{x}$ is presented, then the number of clusters is \mbox{$k^{new} = k^{old} + 1$},  the number of samples encoded by this cluster is $n^{new}_{k^{new}} = 1$, the clusters' prototype is set to $\bm{v}^{new}_{k^{new}} = \bm{x}$, the initial compactness is $CP^{new}_{k^{new}} = 0$, and vector $\bm{g}^{new}_{k^{new}} = \Vec{\bm{0}}$ (unless otherwise noted). Naturally, clusters that do not encode the presented sample remain with constant parameter values for the duration of that input presentation. Also note that, when necessary, the Euclidean norm is replaced with the squared Euclidean norm (\ie~$\norm{\cdot}^2$) to allow for the computation of compactness $CP$ (as per~\cite{Moshtaghi2018, Moshtaghi2018b}). Finally, for iCVIs that require the computation of pairwise (dis)similarity between prototypes, the (dis)similarity matrix is kept in memory, where only the rows and columns corresponding to the prototype that is adapted are modified. 

\subsection{Incremental Calinski-Harabasz index (iCH)}
\label{Sec:iCH}

The iCH computation is defined as:
\begin{equation}
CH^{new} = \frac{\sum \limits_{i=1}^{k^{new}} SEP_i^{new}}{\sum \limits_{i=1}^{k^{new}} CP_i^{new}} \times \frac{N^{new}-k^{new}}{k^{new}-1},
\label{Eq:iCH}
\end{equation}
where
\begin{equation}
SEP_i^{new} = n_i^{new} \| \bm{v}_i^{new} - \bm{\mu}_{data}^{new} \|^2.
\label{Eq:iCH2}
\end{equation}
Note that the variables $\{n_1,...,n_k\}$, $\{\bm{v}_1,...,\bm{v}_k\}$, $\{CP_1,...,CP_k\}$, $\{\bm{g}_1,...,\bm{g}_k\}$, $\bm{\mu}_{data}$, $k$, $N$, and $\{SEP_1,...,SEP_k\}$ are all kept in memory. These are updated using Eqs.~(\ref{Eq:iCVI_1b}) to~(\ref{Eq:iCVI_3}), except for $SEP$, which is adapted using Eq.~(\ref{Eq:iCH2}). The data mean $\bm{\mu}_{data}$ is updated similarly to the prototypes $\bm{v}$ (\ie~Eq.~(\ref{Eq:iCVI_1c})).

\subsection{Incremental I index (iI)}
\label{Sec:iPBM}

The iI computation is defined as:
\begin{equation}
I^{new} = \left[ \frac{\max\limits_{i \neq j}\left( \| \bm{v}_i^{new} - \bm{v}_j^{new} \|^2 \right)}{\sum \limits_{i=1}^{k} CP_i^{new}} \times \frac{CP_0^{new}}{k^{new}} \right]^p,
\label{Eq:iPBM}
\end{equation}
\noindent where $CP_0$ and $\sum \limits_{i=1}^{k} CP_i^{new}$ correspond to $E_1$ and $E_k$, respectively. These are updated according to Eqs.~(\ref{Eq:iCVI_1b}) to~(\ref{Eq:iCVI_3}) along with the remaining compactness variables. Only the pairwise distances with respect to the updated prototype at any given iteration need to be recomputed. 

\subsection{Incremental Silhouette index (iSIL)}
\label{Sec:iSIL}

The SIL index is inherently batch (offline), since it requires the entire data set to be computed (the silhouette coefficients are averaged across all data samples in Eq.~(\ref{Eq:valind_sil1})). To remove such a requirement and enable incremental updates, a hard version of the centroid-based SIL variant introduced in~\cite{Rawashdeh2012} is employed here as well as the squared Euclidean norm (\ie~\mbox{$\| \cdot \| ^2$}): this is done in order to employ the recurrent formulation of the compactness in Eq.~(\ref{Eq:iCVI_2}). Consider the matrix $\bm{S}_{k \times k}$, where $k$ prototypes~$\bm{v}_i$ are used to compute the centroid-based SIL (instead of the $N$ samples $\bm{x}_i$ - which, by definition, are discarded after each presentation in online mode). Define each entry $s_{i,j} = D(\bm{v}_i,\omega_j)$ (dissimilarity of $\bm{v}_i$ to cluster $\omega_j$)  of $\bm{S}_{k \times k}$ as:
\begin{equation}
s_{i,j} = \frac{1}{n_j} \sum\limits_{\substack{l=1 \\ \bm{x}_l \in \omega_j}}^{n_j} \| \bm{x}_l - \bm{v}_i \|^2 = \frac{1}{n_j}CP(\bm{v}_i, \omega_j), 
\label{Eq:Smat_inc1}
\end{equation}
\noindent where $i=\{1,...,k\}$ and $j=\{1,...,k\}$. The silhouette coefficients can be obtained from the entries of~$\bm{S}_{k \times k}$ as:
\begin{equation}
sc_i = \frac{ \min\limits_{l,l \neq J}(s_{i,l}) - s_{i,J}}{\max\left[s_{i,J} , \min\limits_{l,l \neq J}(s_{i,l}) \right]}, \bm{v}_i \in \omega_J.
\label{Eq:cSIL1}
\end{equation}
\noindent where $a_i=s_{i,J}$ and $b_i=\min\limits_{l,l \neq J}(s_{i,l})$. 

At first, when examining Eq.~(\ref{Eq:Smat_inc1}), one might be tempted to store a $k \times k$ matrix of compactness entries along with their accompanying $k^2$ vectors $\bm{g}$ (one for each entry) to enable incremental updates of each element of matrix of $\bm{S}_{k \times k}$; this approach, however, may lead to unnecessarily large memory requirements. A more careful exam shows that it is sufficient to simply redefine $CP$ and $\bm{g}$ for each cluster~$i$ ($i=\{1,...,k\}$) as:
\begin{equation}
CP_i = \sum \limits_{
\substack{j=1 \\ \bm{x}_j\in \omega_i}}^{n_i} \| \bm{x}_j - \Vec{\bm{0}} \|^2 = \sum \limits_{
\substack{j=1 \\ \bm{x}_j\in \omega_i}}^{n_i} \| \bm{x}_j \|^2,
\label{Eq:Smat_inc2}
\end{equation}
\begin{equation}
\bm{g}_i = \sum \limits_{
\substack{j=1 \\ \bm{x}_j\in \omega_i}}^{n_i} \left( \bm{x}_j - \Vec{\bm{0}} \right)  = \sum \limits_{
\substack{j=1 \\ \bm{x}_j\in \omega_i}}^{n_i} \bm{x}_j,
\label{Eq:Smat_inc3}
\end{equation}
\noindent which is equivalent to fixing $\bm{v}=\Vec{\bm{0}}$. Therefore, their incremental update equations become (as opposed to Eqs.~(\ref{Eq:iCVI_2}) and~(\ref{Eq:iCVI_3})):
\begin{equation}
CP_i^{new} = CP_i^{old} + \| \bm{x} \|^2,
\label{Eq:Smat_inc4}
\end{equation}
\begin{equation}
\bm{g}_i^{new} = \bm{g}_i^{old} + \bm{x}.
\label{Eq:Smat_inc5}
\end{equation}

Using this trick, when a sample~$\bm{x}$ is assigned to cluster~$\omega_J$, then the update equations for each entry $s_{i,j}$ of $\bm{S}_{k \times k}$ are given by Eq.~(\ref{Eq:Smat_inc6}). Note that the numerators of the expressions in Eq.~(\ref{Eq:Smat_inc6}) update the compactness ``as if'' the prototype has changed from $\Vec{\bm{0}}$ to $\bm{v}^{new}$  at every iteration ($\Delta \bm{v}
= - \bm{v}^{new}$). The remaining variables such as $n$, $N$, and $\bm{v}$ are updated as previously described. This allows $\{CP_1,...,CP_k\}$ and $\{\bm{g}_1,...,\bm{g}_k\}$ to continue being stored similarly to the previous iCVIs, instead of a $k \times k$ matrix of compactness and the associated $k^2$ vectors $\bm{g}$.
\begin{equation}
s_{i,j}^{new} = \begin{cases}
\frac{1}{n_j^{new}} \left( CP_j^{old} + \| \bm{z}_i \|^2 + n_j^{old} \| \bm{v}_i^{old} \| ^2 - 2\bm{v}_i^{old~^T}\bm{g}_j^{old} \right)&,~(i \neq J,j=J) \\ 
\frac{1}{n_j^{old}} \left( CP_j^{old} + n_j^{old} \| \bm{v}_i^{new} \| ^2 - 2\bm{v}_i^{new~^T}\bm{g}_j^{old} \right) &,~(i = J,j \neq J) \\ 
\frac{1}{n_j^{new}} \left( CP_j^{old} + \| \bm{z}_j \|^2 + n_j^{old} \| \bm{v}_j^{new} \| ^2 - 2\bm{v}_j^{new~^T}\bm{g}_j^{old} \right) &,~(i=J,j=J) \\ 
s_{i,j}^{old} &,~(i \neq J,j \neq J) \\  
\end{cases}
\label{Eq:Smat_inc6}
\end{equation}
 
In the case where a new cluster $\omega_{k+1}$ is created following the presentation of sample $\bm{x}$, then a new column and a new row are appended to the matrix $\bm{S}_{k \times k}$. Unlike the other iCVIs, the compactness $CP_{k+1}$ and vector $\bm{g}_{k+1}$ of this cluster are initialized as $\| \bm{x} \|^2$ and $\bm{x}$, respectively. Then, the entries of $\bm{S}_{k \times k}$ are updated using Eq.~(\ref{Eq:Smat_inc7}). 
\begin{equation}
s_{i,j}^{new} = \begin{cases}
CP_{k+1} + \| \bm{v}_i^{old} \| ^2 - 2\bm{v}_i^{old~^T}\bm{g}_{k+1} &,~(i \neq k+1,j=k+1) \\ 
\frac{1}{n_j^{old}} \left( CP_j^{old} + n_j^{old} \| \bm{v}_i^{new} \| ^2 - 2\bm{v}_i^{new~^T}\bm{g}_j^{old} \right) &,~(i = k+1,j \neq k+1) \\ 
0 &,~(i=k+1,j=k+1) \\ 
s_{i,j}^{old} &,~(i \neq k+1,j \neq k+1) \\  
\end{cases}
\label{Eq:Smat_inc7}
\end{equation}

Following the incremental updates of the entries of $\bm{S}_{k \times k}$ (Eq.~(\ref{Eq:Smat_inc6}) or~(\ref{Eq:Smat_inc7})), the silhouette coefficients ($sc_i$) are computed (Eq.~(\ref{Eq:cSIL1})), and the iSIL is updated as:
\begin{equation}
SIL^{new} = \frac{1}{k^{new}}\sum_{i=1}^{k^{new}} sc_i^{new}.
\label{Eq:icSIL}
\end{equation}

\subsection{Incremental Negentropy Increment (iNI)}
\label{Sec:iNI}

The iNI computation is defined as:
\begin{equation}
NI^{new} = \sum\limits_{i=1}^{k} p_i^{new}\ln \left(
\frac{\sqrt{| \bm{\Sigma}_i^{new} |}}{p_i^{new}} \right)
- \frac{1}{2} \ln |\bm{\Sigma}_{data}|
\label{Eq:iNI}
\end{equation}

\noindent where \mbox{$p_i^{new} = n_i^{new}/N^{new}$}, and $\bm{\Sigma}_i^{new}$ is computed using the following recursive formula~\cite{duda2000}:
\setlength{\arraycolsep}{0.0em}
\begin{eqnarray}
\bm{\Sigma}^{new}&{}={}&\frac{n^{new}-2}{n^{new}-1}\left(\bm{\Sigma}^{old} - \delta I \right) + \frac{1}{n^{new}}\left(\bm{x} - {v}^{old} \right)\left(\bm{x} - {v}^{old} \right)^T + \delta I
\label{Eq:iSigma}
\end{eqnarray}
\setlength{\arraycolsep}{5pt}

This work's authors set $\delta = 10^{-\frac{\epsilon}{d}}$ to avoid numerical errors, where $\epsilon$ is a user-defined parameter. If a new cluster is created, then $\bm{\Sigma} = \delta I$ and $|\bm{\Sigma}|=10^{-\epsilon}$.

\subsection{Incremental representative Cross Information Potential~~~(irCIP) and cross-entropy (irH)}
\label{Sec:irCIP}

Section~\ref{Sec:results} will show that using the representative cross-entropy rH for computing the CEF makes it easier to observe the behavior of the incremental clustering process (this corroborates a previous study in which rH was deemed more informative than rCIP for multivariate data visualization~\cite{leonardo2018a}):
\begin{equation}
rH(\omega_i, \omega_j) = - \ln \left[rCIP(\omega_i, \omega_j)\right],
\label{Eq:valind_rCIP4}
\end{equation}
\begin{equation}
CEF = \sum\limits_{i=1}^{k-1} \sum\limits_{j=i+1}^k rH(\omega_i, \omega_j).
\label{Eq:valind_rH_CEF}
\end{equation}

Note that, as opposed to the rCIP-based CEF, larger values of rH-based CEF indicate better clustering solutions (maximization). Concretely, since the CEF only measures separation, then, like iNI, it is only necessary to update the means and the covariance matrices online in order to construct the incremental CEF (iCEF). This is also done using Eqs.~(\ref{Eq:iCVI_1c}) and~(\ref{Eq:iSigma}), respectively. The iCEFs, based on rCIP and rH, are hereafter referred to as irCIP and irH, respectively.

\subsection{Incremental Conn\_Index (iConn\_Index)}
\label{Sec:iConn}

The Conn\_Index is another inherently batch CVI, as each element $(i,j)$ of the CADJ matrix requires the count of the samples in the data set with the first and second closest prototypes, $\bm{v}_i$ and $\bm{v}_j$ respectively. Naturally, when clustering data online, $\bm{v}_i$ and $\bm{v}_j$ may change for previously presented samples as prototypes are continuously modified or created. However, for the purpose of building and incrementing CADJ and CONN matrices online (with only one element changing per sample presentation), it is assumed that the trends exhibited over time by the iConn\_Index does not differ dramatically from its offline counterpart. Batch calculation can be eliminated entirely by keeping the values of Eqs.~(\ref{Eq:valind_conn4}) and~(\ref{Eq:valind_conn7}) in memory and updating only the entries corresponding to the winning prototype $\bm{v}_i$. 

In this study, the self-consistent hierarchy and multi-prototype cluster representation required by the iConn\_Index was generated using fuzzy SMART, whose modules~A and~B are used for prototype and cluster definition, respectively. Fuzzy SMART's module A was modified in such a way that it forcefully creates two prototypes from the first two samples of every emerging cluster in module B. By enforcing this dynamic, each cluster always possesses at least two prototypes for the computation of the iConn\_Index. This strategy addresses two problems: first, it allows CADJ to be created from the second sample seen and onward; second, it prevents some cases in which well-separated clusters are strongly connected simply because one of them does not have another prototype to assume the role of the second winner. The second winning prototype for a sample $\bm{v}_j$ is the winning A-side category when the first winning prototype $\bm{v}_i$ has been removed from the A-side category set.

The iConn\_Index demands certain boundary conditions. In the case of exactly one prototype and one category, such as the case for the very first sample presentation, the CADJ matrix cannot be incremented, and the iConn\_Index will default to~0~\cite{tasdemir2011}. This paper presents a remedy for this whereby a count of samples is kept separate from the CADJ matrix (instance counting~\cite{Carpenter1998}). Upon creation of the second prototype $\bm{v}_2$ in fuzzy SMART's module~A, the CADJ matrix will be incremented for the first time at element $(2,1)$. At this point, the element $(1,2)$ will be set to the number of samples seen so far belonging to $\bm{v}_1$. This situation is encountered in the very first sample presentation to fuzzy SMART.

Note that, in the case of a single category, $Inter\_Conn$, given by Eq.~(\ref{Eq:valind_conn5}), defaults to 1~\cite{tasdemir2011}. In the case of a category with a single prototype, the $Intra\_Conn$ for that category, given by Eq.~(\ref{Eq:valind_conn4}), also defaults to a value of 1~\cite{tasdemir2011}. Finally, instead of the original constraint \mbox{$CADJ(i,j)>0$} imposed by Eq.~(\ref{Eq:valind_conn8}), this paper's iConn\_Index implementation uses \mbox{$CONN(i,j)>0$}, as this makes its behavior smoother and more consistent in this application domain.

\section{Numerical experiments setup} \label{Sec:setup}

The numerical experiments were carried out using the MATLAB software environment. The Cluster Validity Analysis Platform Toolbox~\cite{cvap} was used to compute the Adjusted Rand Index ($ARI$)~\cite{hubert1985} to evaluate the partitions detected by the fuzzy ART-based clustering algorithms. Two synthetic data sets were used: (1) \textit{R15}~\cite{veenman2002,shape}, consisting of 800 samples and 15 clusters in two dimensions and (2) \textit{D4}, which is an in-house artificially generated data set with 2000 samples and 4 clusters also in two dimensions. For comparison purposes, hard clustering versions of iDB, iXB and PS CVIs were used in the experiments. Finally, it should be noted that this study does not employ multi-prototype representations for the irCIP and irH (\ie~$M_i=M_j=1, \forall i,j$ in Eq.~(\ref{Eq:valind_rCIP2})) since each of the clusters from the data sets used in these experiments can be modeled using single Gaussian distributions.

All fuzzy ART and SMART dynamics were performed with normalized and complement coded input, whereas the CVI computations were performed using the normalized data. To emulate scenarios in which there is a natural order of presentation, the samples were presented to fuzzy ART/SMART in a cluster-by-cluster fashion where samples within a given cluster were randomized. Finally, in these experiments, $\epsilon=12$ in Eq.~(\ref{Eq:iSigma}) for the incremental computation of the covariance matrices used by irCIP, irH and iNI. The source code of the CVIs/iCVIs, fuzzy ART/SMART, and experiments is provided at the Applied Computational Intelligence Laboratory public GitLab repository\footnote[1]{https://github.com/ACIL-Group/iCVIs}.

\section{A comparative study} \label{Sec:results}

This section discusses the behavior of the iCVIs in three general cases when assessing the quality of the partitions detected by fuzzy ART-based systems in real-time: (1) high-quality partitions, (2) under-partitions, and (3) over-partitions. It should be emphasized that this analysis is not focused on evaluating the performance or capabilities of the chosen clustering algorithms, but instead the purpose of this study is to observe the behavior of the iCVIs in these different scenarios to gain insight on their applicability. Moreover, in each of these scenarios, the iCVIs' dynamics are investigated in two sub-cases: (a) the creation of a new cluster and (b) the presentation of samples within a given cluster. 

The following discussion is relative to the data sets used in the experiments and their respective order of cluster and sample presentation (Fig.~\ref{Fig:all_orders}). This is not an exhaustive study of all possible permutations of clusters and samples, as each of them may trigger different global behaviors of the iCVIs. Nonetheless, it can be assumed that some behaviors are typical, which allows the inference of some particular problems that may arise during incremental unsupervised learning.

Similar to~\cite{Moshtaghi2018, Moshtaghi2018b, Keller2018, Ibrahim2018, Ibrahim2018b}, a natural ordering, \ie~meaningful temporal information is assumed. The \textit{R15} data set was used to illustrate the behavior of the iCVIs in cases (1) and (2), which are depicted in Figs.~\ref{Fig:R15HQ} and~\ref{Fig:R15UP}, respectively. Alternately, the \textit{D4} data set was used to illustrate the behavior of the iCVIs in cases (1) and (3), which are depicted in Figs.~\ref{Fig:D4HQ} and~\ref{Fig:D4OP}, respectively. For both data sets, case (1) is used as a reference to which their respective cases (2) and (3) are compared. Moreover, Figs.~\ref{Fig:R15HQ} to~\ref{Fig:D4OP} depict the iCVIs immediately following the creation of the second cluster. 

\newcommand{\sorder}{0.27}
\begin{figure}[!b]
\centerline{
\subfloat[]{\includegraphics[width=\sorder\columnwidth]{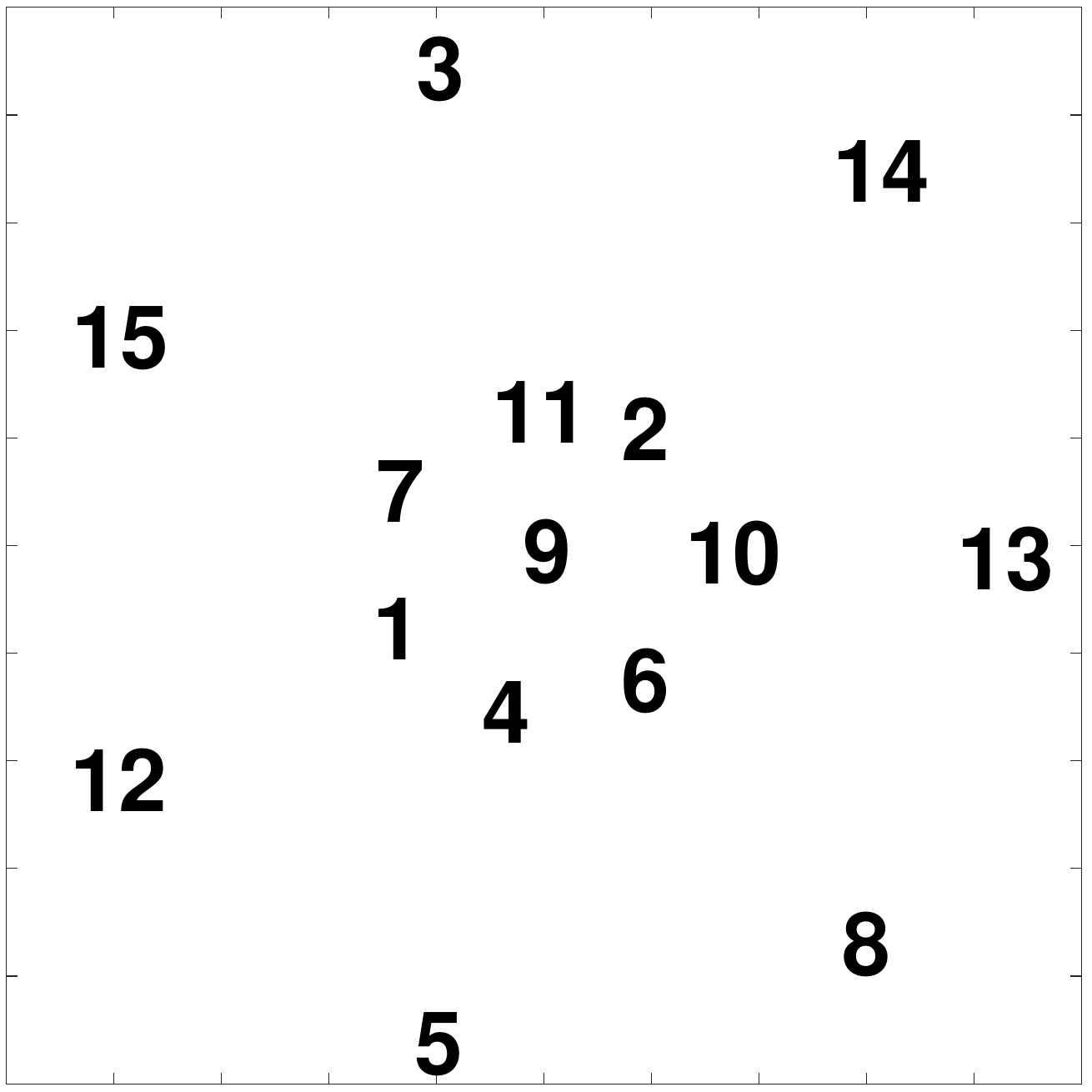}
\label{Fig:order1}}
\hfil
\subfloat[]{\includegraphics[width=\sorder\columnwidth]{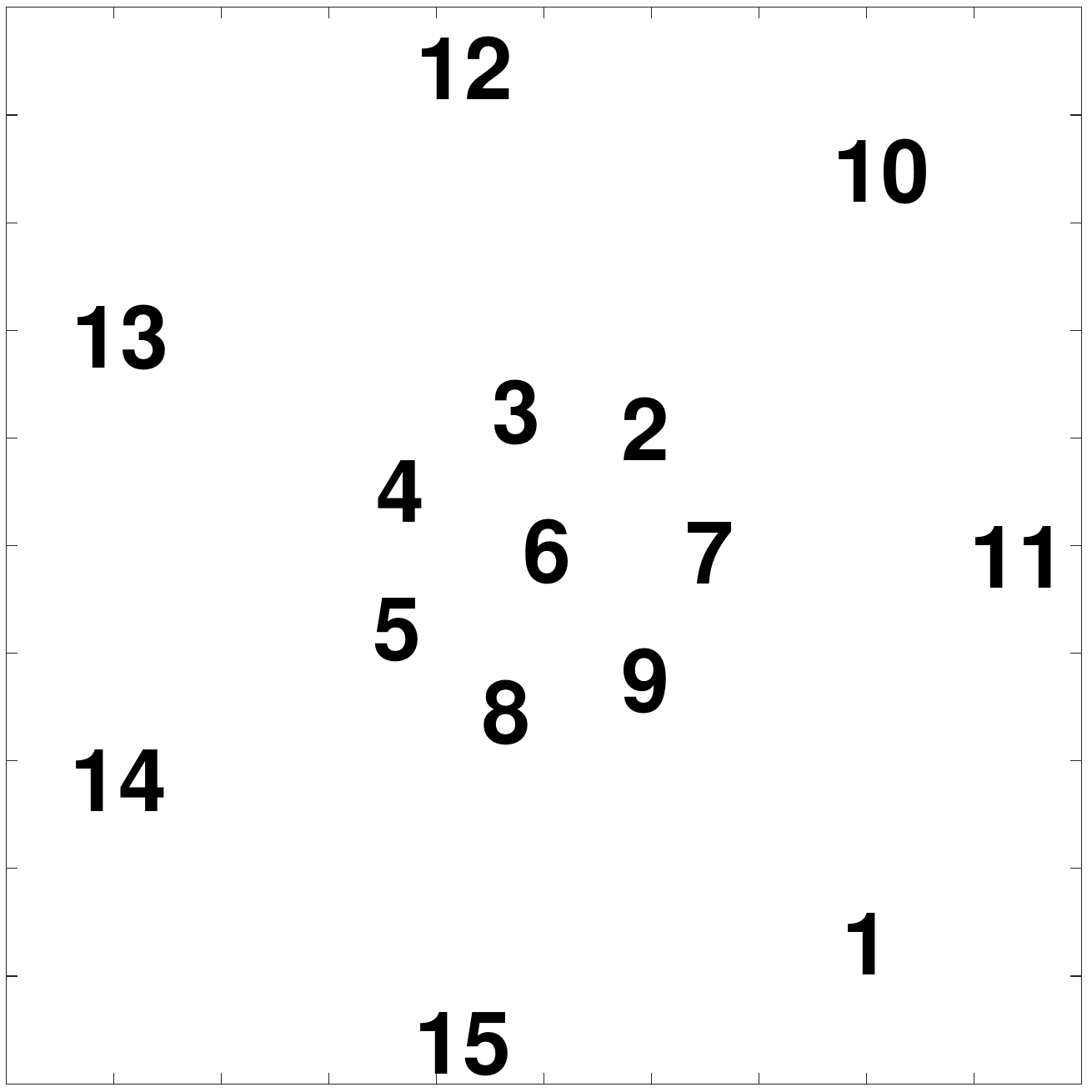}
\label{Fig:order2}}
\hfil
\subfloat[]{\includegraphics[width=\sorder\columnwidth]{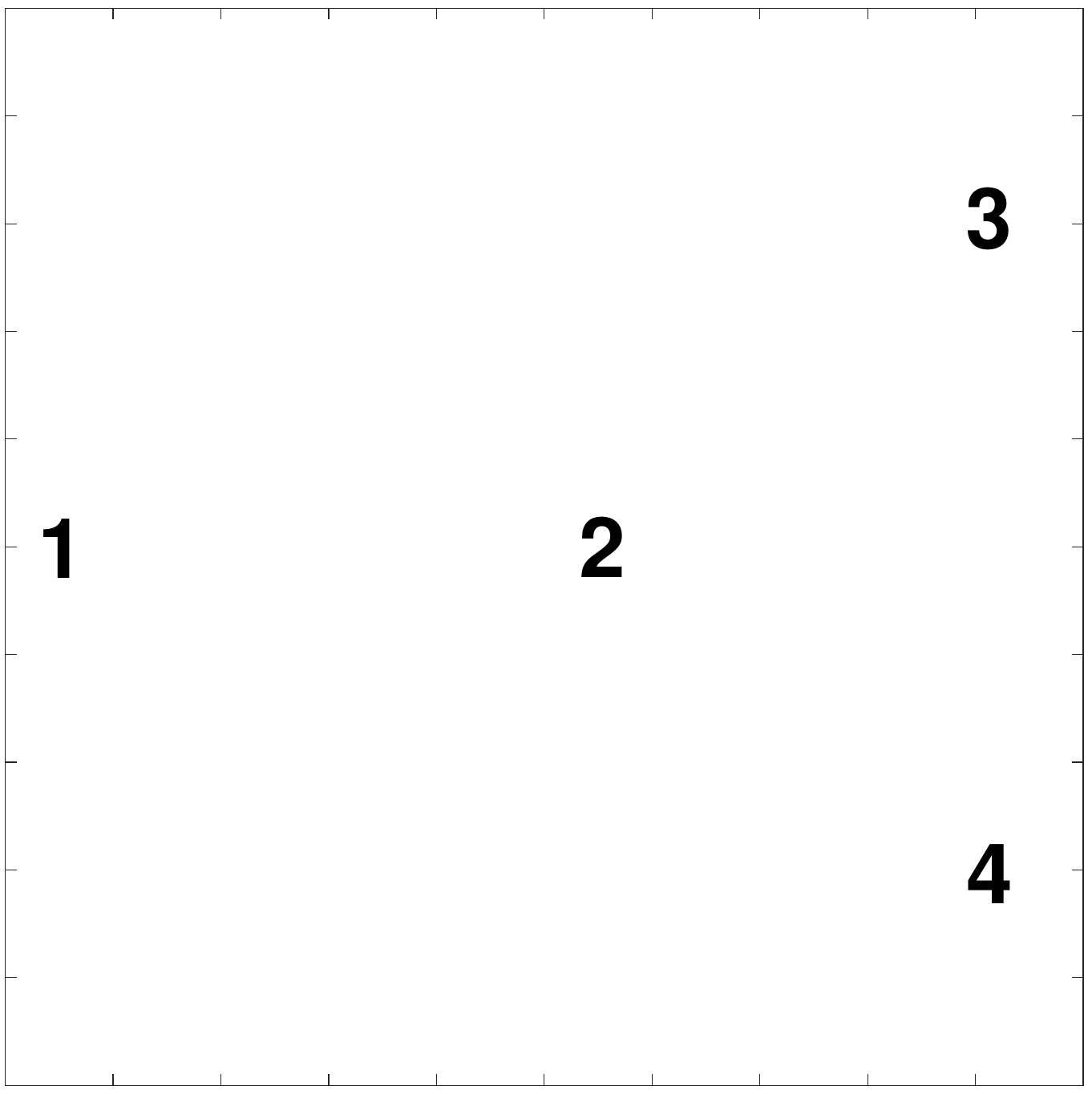}
\label{Fig:order3}}
}
\caption{Presentation order of the classes for the experiments carried out in (a) Fig.~\ref{Fig:R15HQ}, (b) Fig.~\ref{Fig:R15UP}, and (c) Figs.~\ref{Fig:D4HQ} and~\ref{Fig:D4OP}.}
\label{Fig:all_orders}
\end{figure}

\subsection{Correct estimation and underestimation of the number of clusters}

Consider the high-quality partition of the \textit{R15} data set shown in Fig.~\ref{Fig:partition1}, which was obtained when presenting samples in the cluster-by-cluster ordering depicted in Fig.~\ref{Fig:order1}. This study shows, in general (and as expected from previous studies on iDB and iXB~\cite{Moshtaghi2018, Moshtaghi2018b, Keller2018, Ibrahim2018, Ibrahim2018b}), that the drastic changes in most iCVI values follow the emergence of new clusters. The exceptions are the iXB and irCIP, which appear much less informative than the other iCVIs used in this particular experiment, as they show no clearly defined tendencies and seem insensitive to the well-separated clusters numbered 12 to 15 in Fig.~\ref{Fig:partition1}.

During the presentation of samples within a given cluster, many different behaviors can be observed. Typically, iCH either improves or has small fluctuations; iSIL and iDB either worsen or have small fluctuations; iI/iPBM and iNI either worsen or improve; iConn\_Index and PS improve; and irH consistently undergoes small fluctuations. Again, irCIP and iXB do not appear to be particularly useful compared to the other iCVIs since no apparent trends were found over the iterations. If an iCVI displays more than one trend, these usually do not occur prominently and simultaneously (\ie~during the presentation of samples from the same cluster). Note that these are important characteristics, since they will help in identifying the under-partition cases.

Now consider the case of underestimating the number of clusters, as shown in Fig.~\ref{Fig:partition2}. The latter was obtained when presenting samples in the cluster-by-cluster ordering depicted in Fig.~\ref{Fig:order2}. This research notes that most iCVIs consistently worsen while the algorithm incorrectly agglomerates samples from different clusters (clusters numbered 2 to 9 in Fig.~\ref{Fig:order2}) into a single cluster (cluster numbered 2 in Fig.~\ref{Fig:partition2}), except for the iConn\_Index (which actually improves due to the strong connectivity among prototypes) and irCIP (which remains constant). Moreover, when incorrectly merging clusters 10 and 11 in Fig.~\ref{Fig:order2} into a single cluster labeled 3 in Fig.~\ref{Fig:partition2}, the performances of all iCVIs are accompanied by a drastic change typically toward worse values (except for PS, which only undergoes a slight slope change), while the number of clusters remains constant. 

The behavior previously described can also be observed for clusters labeled 4 and 1 in Fig.~\ref{Fig:partition2}. Drastic (iSIL, irCIP, irH, iNI, iDB, iXB, and iConn\_Index) or more subtle (iCH, iI/iPBM) changes entailing worsening trends take place in the behavior of all CVIs in Fig.~\ref{Fig:R15UP} when these samples are classified to the same cluster - again, with the exception of PS, which still improves, but with a different inclination. These clearly indicate that the clustering algorithm is mistakenly encoding the samples under the same cluster umbrella.

At this stage, it is important to be cautious because even when a high-quality partition is retrieved (Fig.~\ref{Fig:R15HQ}), some iCVIs (such as iSIL, iConn\_Index, and iDB), can both improve and worsen when fuzzy ART is allocating samples to the same cluster (although this happens less frequently and less drastically). Therefore, it is recommended to observe more than one iCVI to determine if under-partition is taking place.  

\subsection{Correct estimation and overestimation of the number of clusters}

For the sake of clarity, over-partition is illustrated using the \textit{D4} data set, which has a smaller number of clusters. First, the iCVI behaviors regarding the high-quality partition shown in Fig.~\ref{Fig:partition3} are observed as a reference; these were obtained using the cluster sequence depicted in Fig.~\ref{Fig:order3}. The same iCVI trends seem to hold following the emergence of new clusters as well as during the presentation of samples belonging to a given cluster (and again, iXB and irCIP provided the least visually descriptive behavior over time). A notable exception, however, is the iNI, which quickly improves immediately after the creation of a new cluster and then worsens as samples from the same cluster are presented. This supports the fact that the iCVI behaviors are not universal: naturally, they are data- and order-dependent.

Now consider the over-partition problem depicted in Fig.~\ref{Fig:partition4}, which was also obtained using the cluster sequence depicted in Fig.~\ref{Fig:order3}. As expected, a steep descent (or ascent depending on the iCVI) usually occurs when new clusters are created. However, since this trend appears to occur regardless of the partition quality (being inherent to all iCVIs), then it is not sufficient to identify this issue. In this scenario, unless there was additional a priori information (\eg~the cardinality of clusters) to detect a premature partition, these iCVIs were unable to patently identify over-partition solely based on the transitions of their values versus the number of clusters.

Moreover, although there is a natural order for the presentation of clusters (\ie~as a time series), the presentation of samples within each cluster is random. Specifically, when the cluster is over-partitioned, samples are not presented in a subcluster-by-subcluster manner, but instead they are randomly sampled from the different subclusters. This adds another layer of complexity and thus makes this problem even more challenging. Compared to the correct partition in Fig.~\ref{Fig:partition3}, most iCVIs do not exhibit an overall behavior that deviates significantly from the one typically expected when accurately partitioning \textit{D4} (Fig.~\ref{Fig:partition3}), although most of them yield worse cluster quality evaluation values. In reality, in a true unsupervised learning scenario, such reference behavior is unavailable; furthermore, the values of most iCVIs are not bounded, thus making this problem even more challenging to detect.

\newcommand{\szxb}{0.22}
\newcommand{\szC}{0.21}
\newcommand{\szP}{0.24}
\begin{figure}[!hp]
\centerline{
\subfloat[Data partition ]{\includegraphics[width=\szC\columnwidth]{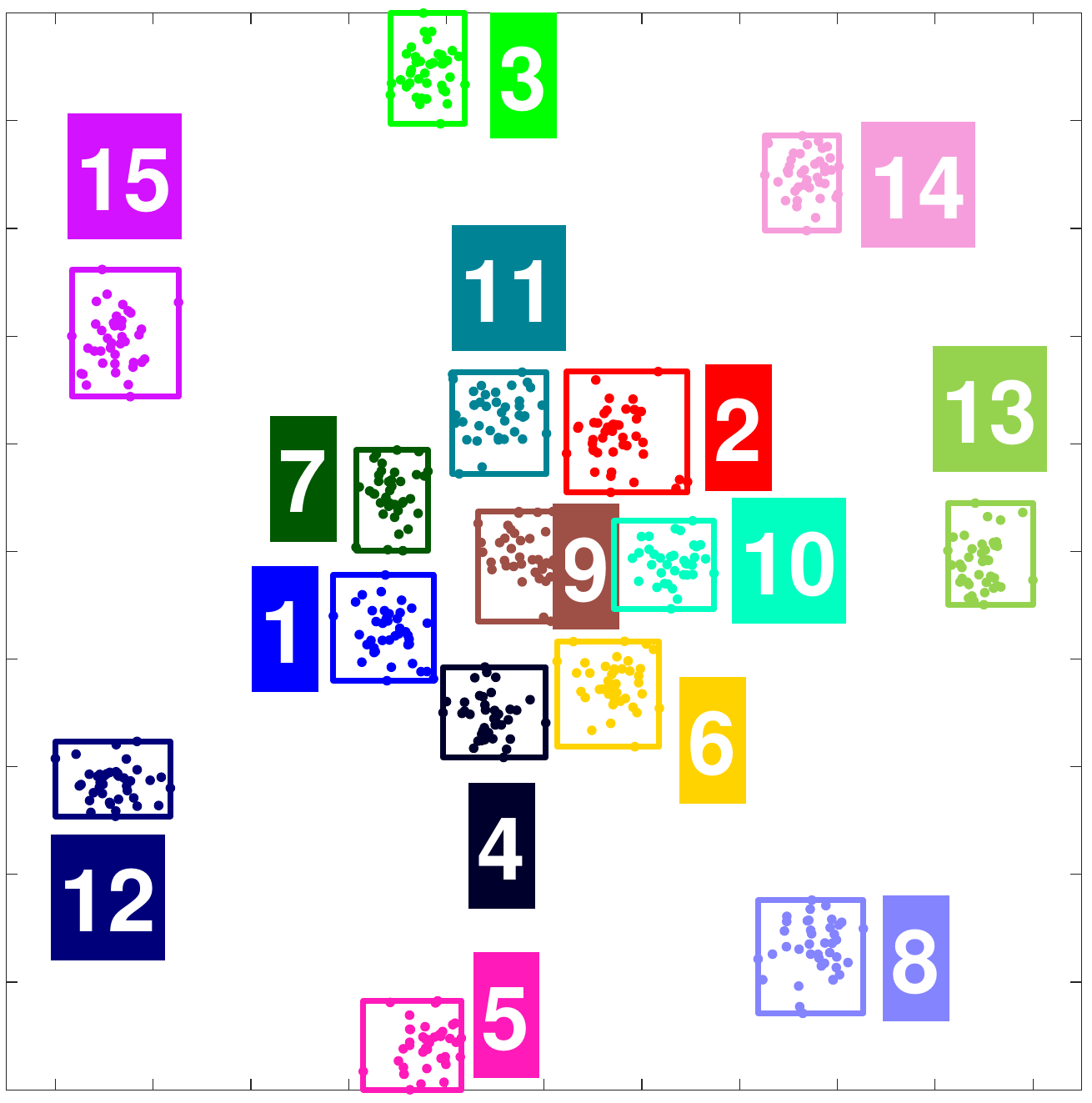}
\label{Fig:partition1}}
\hfil
\subfloat[Fuzzy SMART's module A connectivity]{\includegraphics[width=\szP\columnwidth]{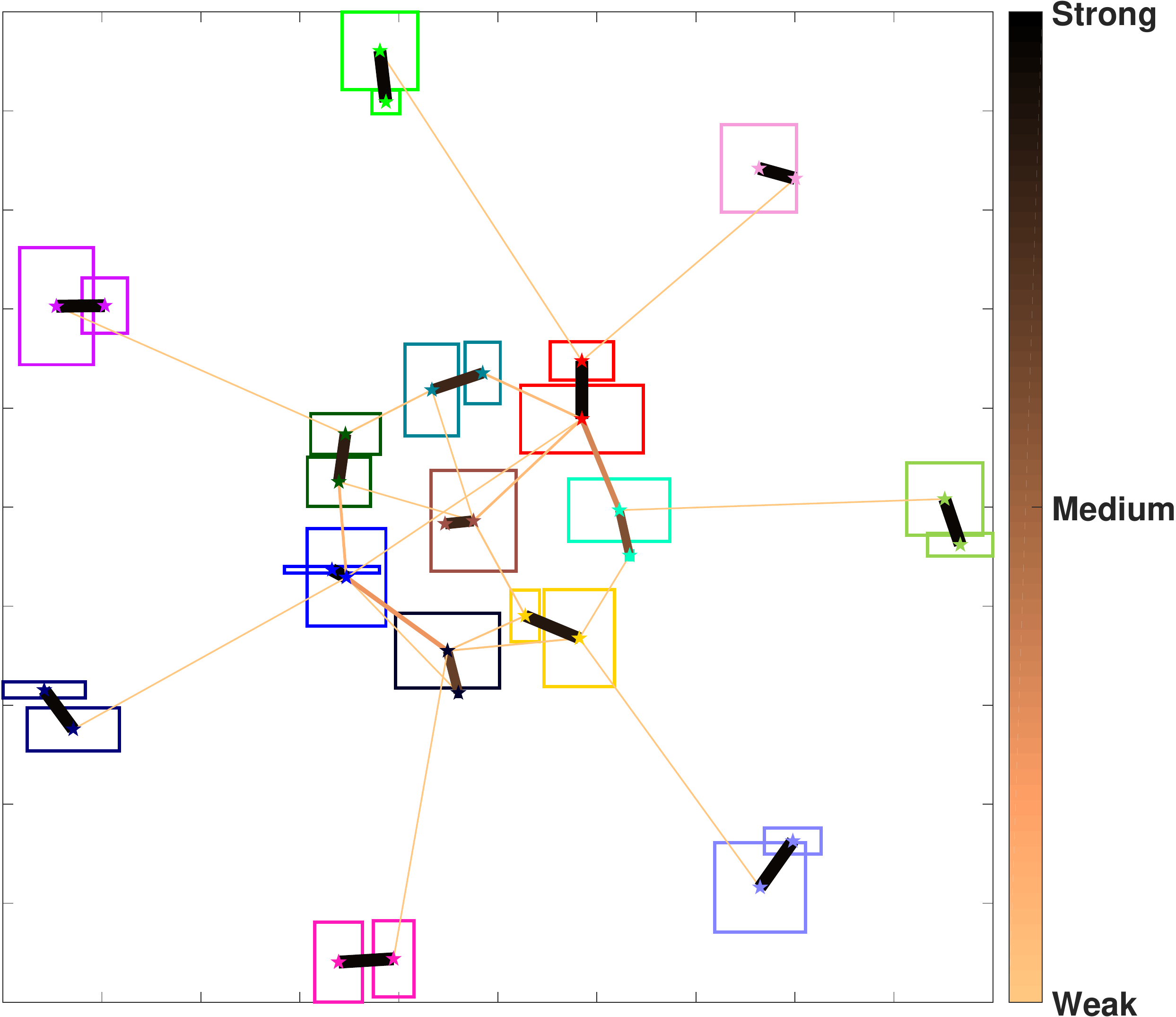}
\label{Fig:SMART_A_1}}
\hfil
\subfloat[iConn\_Index]{\includegraphics[width=\szxb\columnwidth]{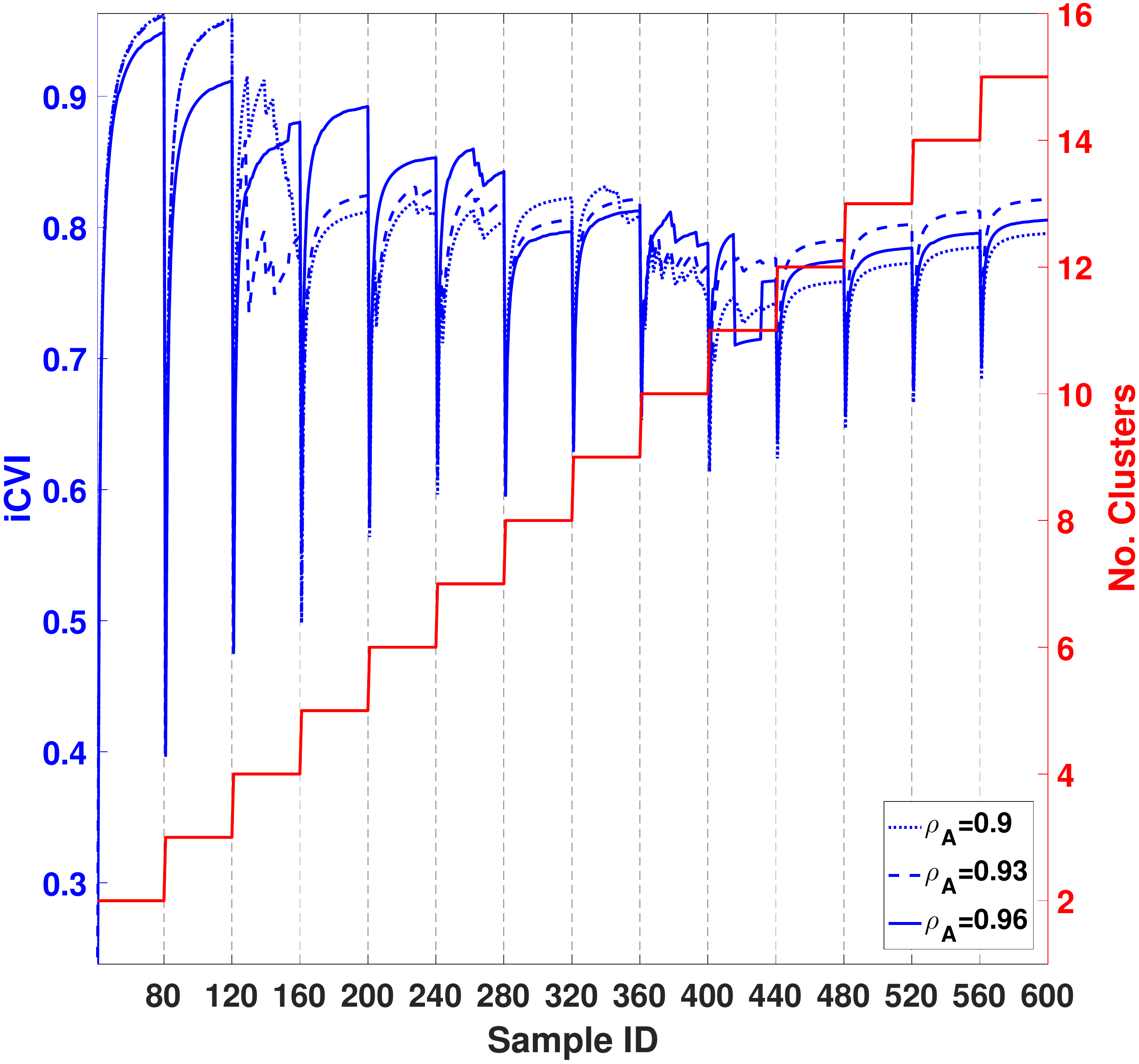}
\label{Fig:iConn1}}
}
\centerline{
\subfloat[iCH]{\includegraphics[width=\szxb\columnwidth]{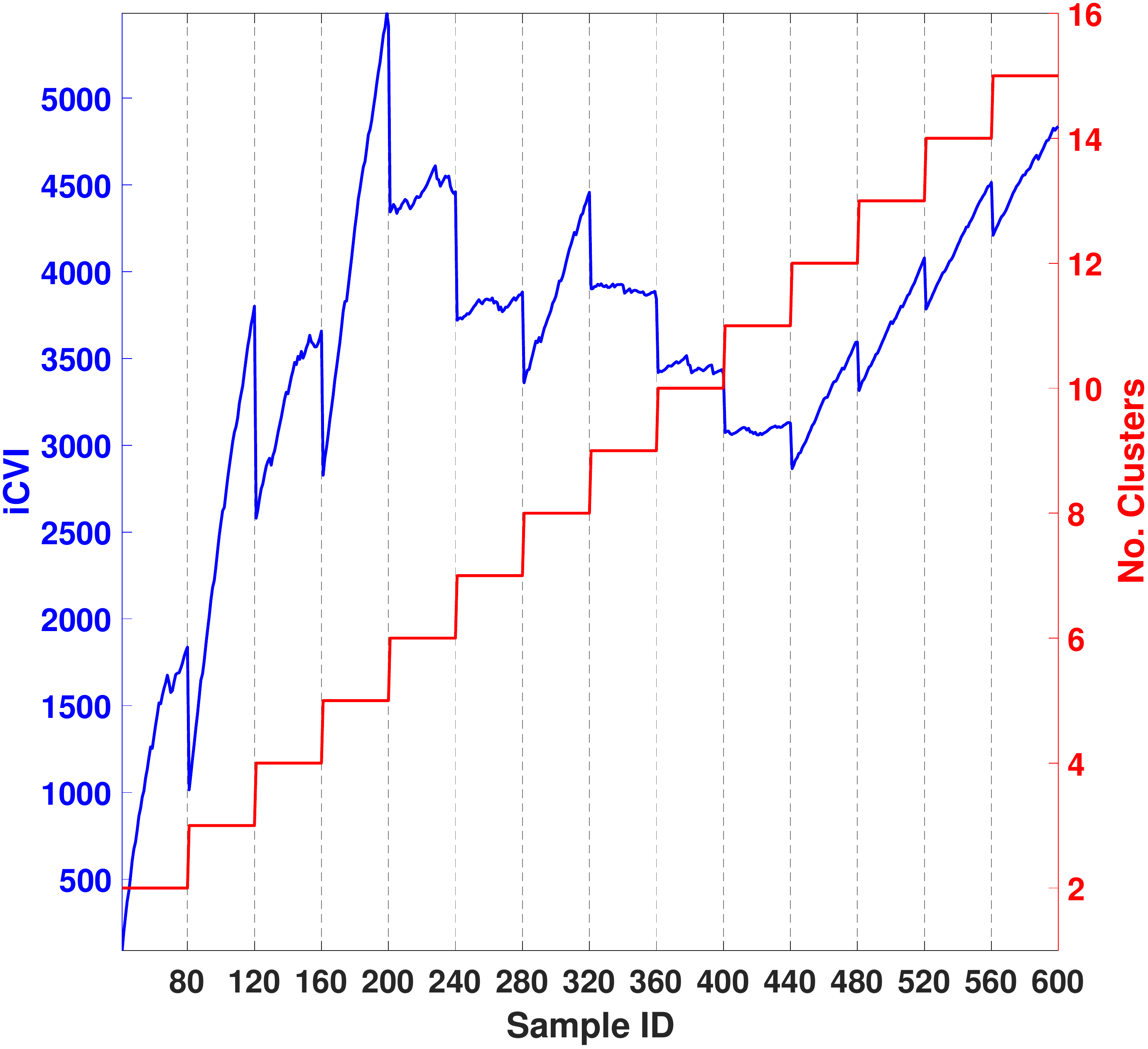}
\label{Fig:iCH1}}
\hfil
\subfloat[iSIL]{\includegraphics[width=\szxb\columnwidth]{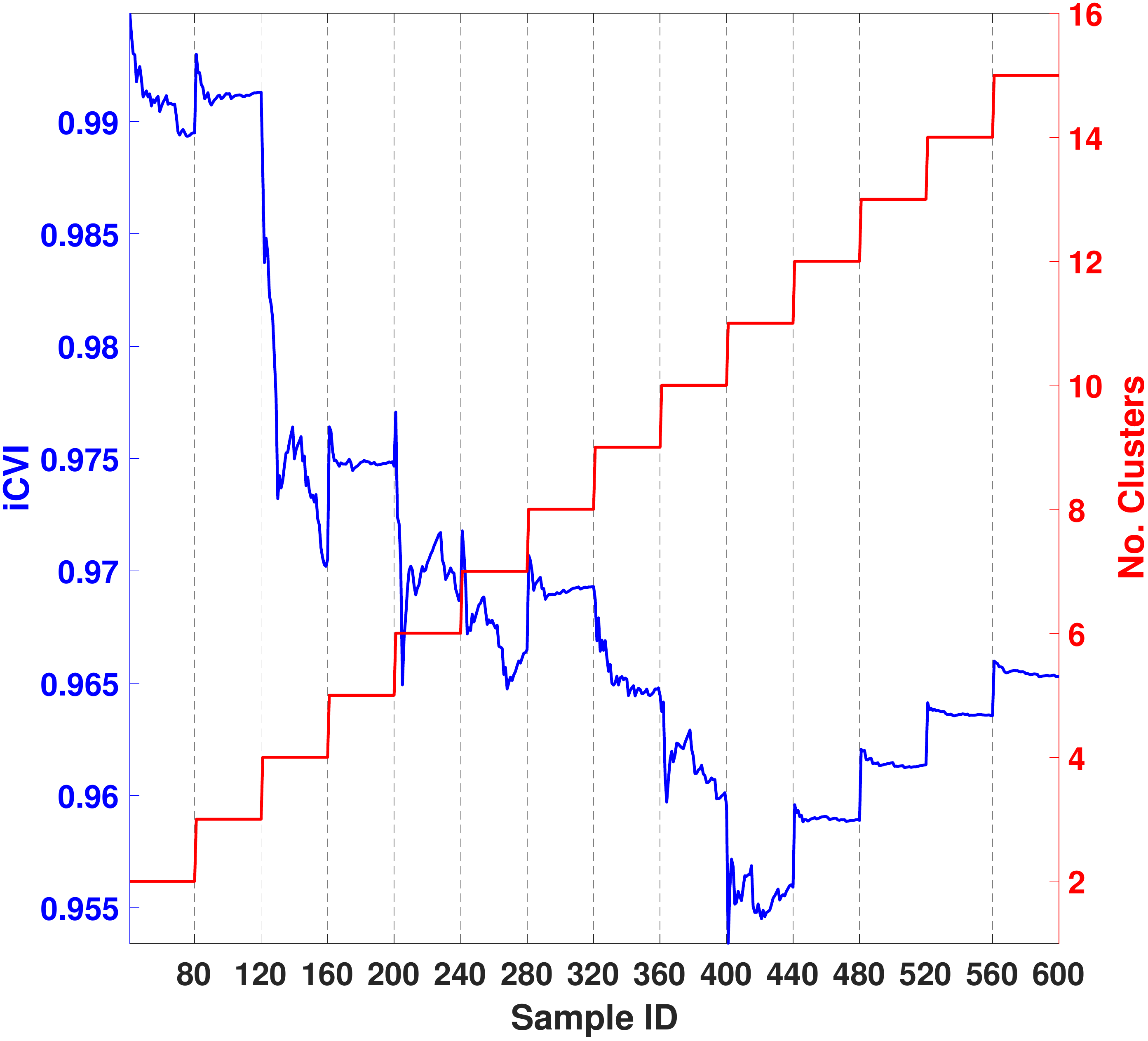}
\label{Fig:iSIL1}}
\hfil
\subfloat[iPBM]{\includegraphics[width=\szxb\columnwidth]{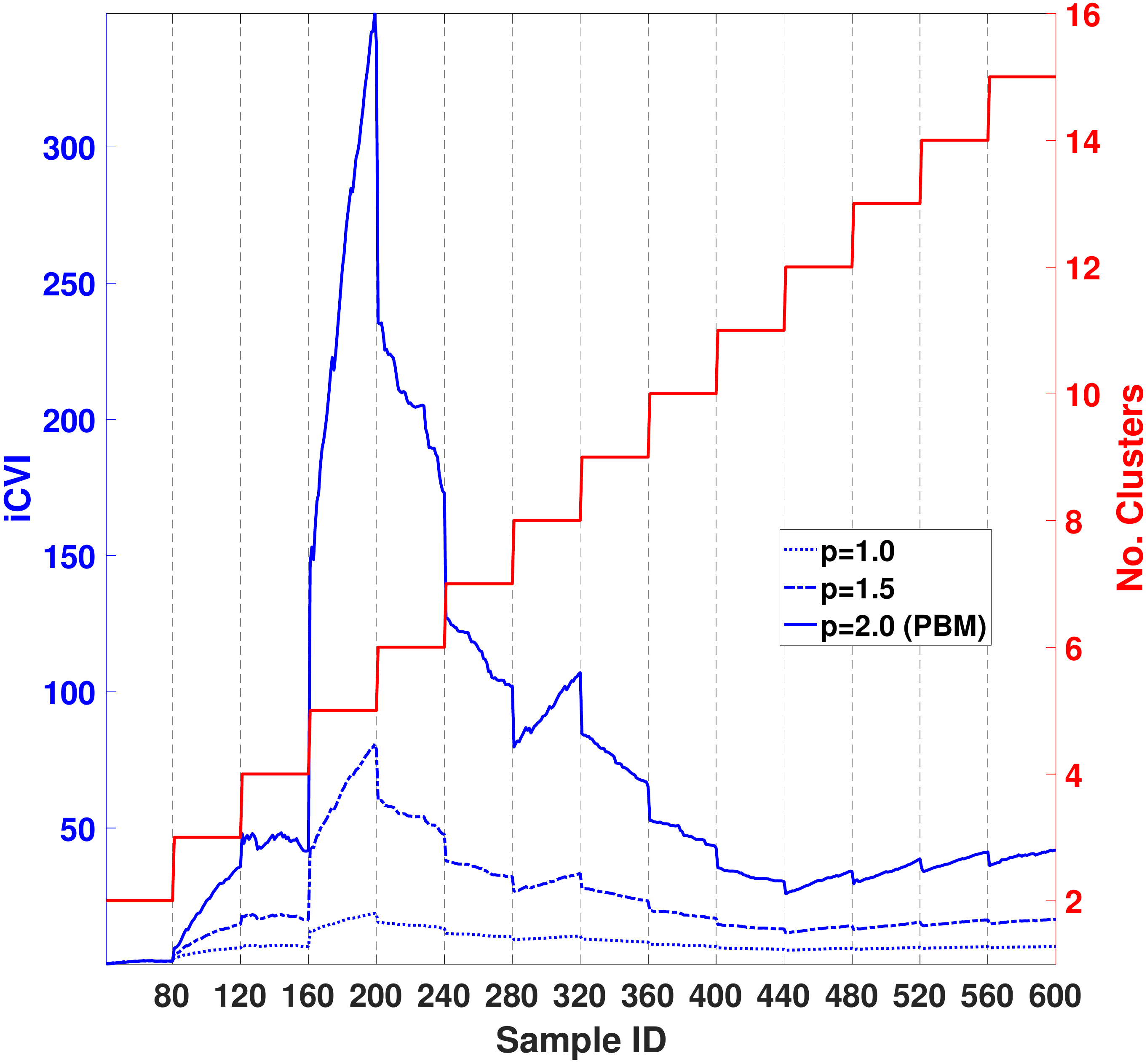}
\label{Fig:iPBM1}}
}
\centerline{
\subfloat[irCIP]{\includegraphics[width=\szxb\columnwidth]{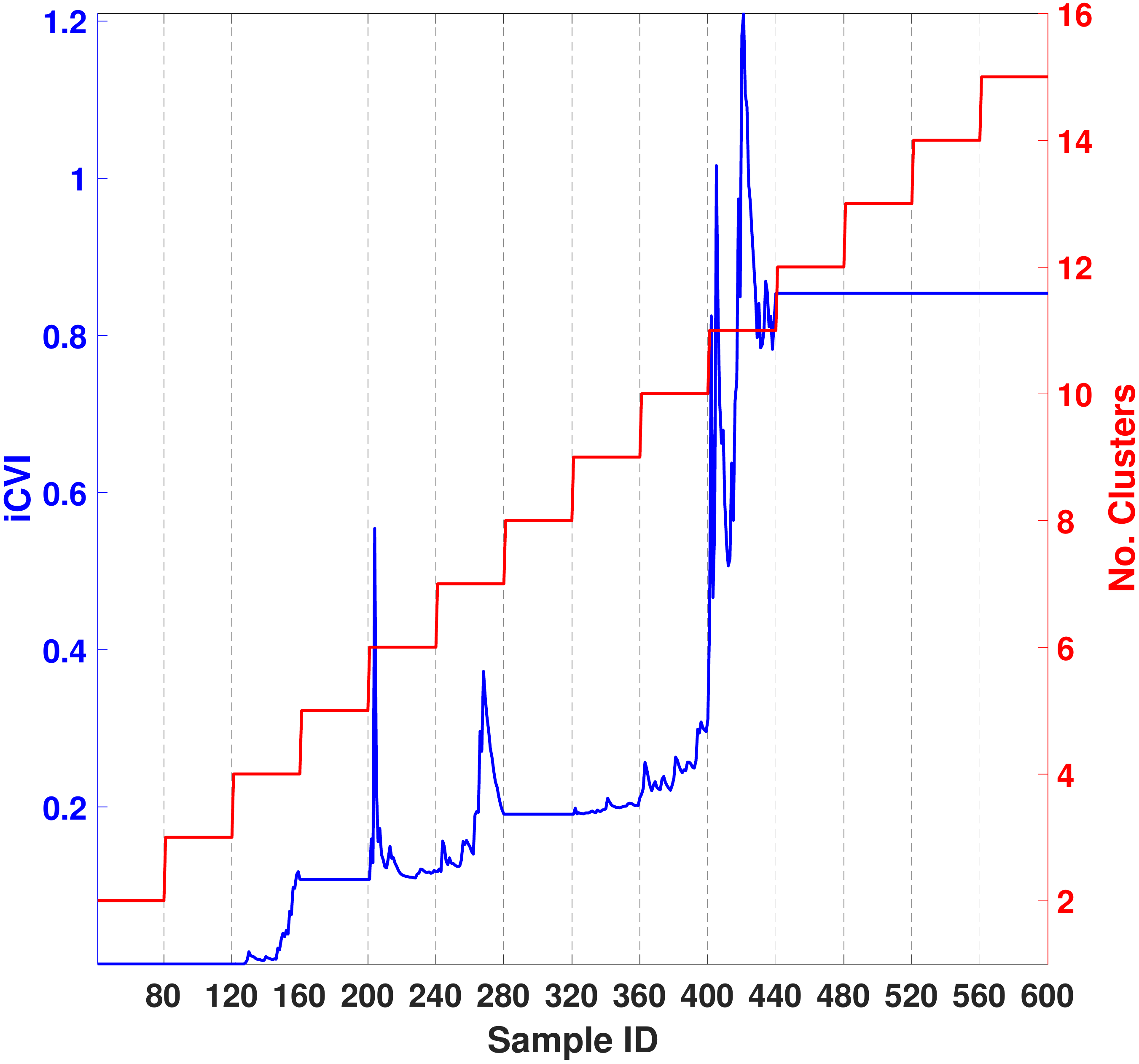}
\label{Fig:irCIP1}}
\hfil
\subfloat[irH]{\includegraphics[width=\szxb\columnwidth]{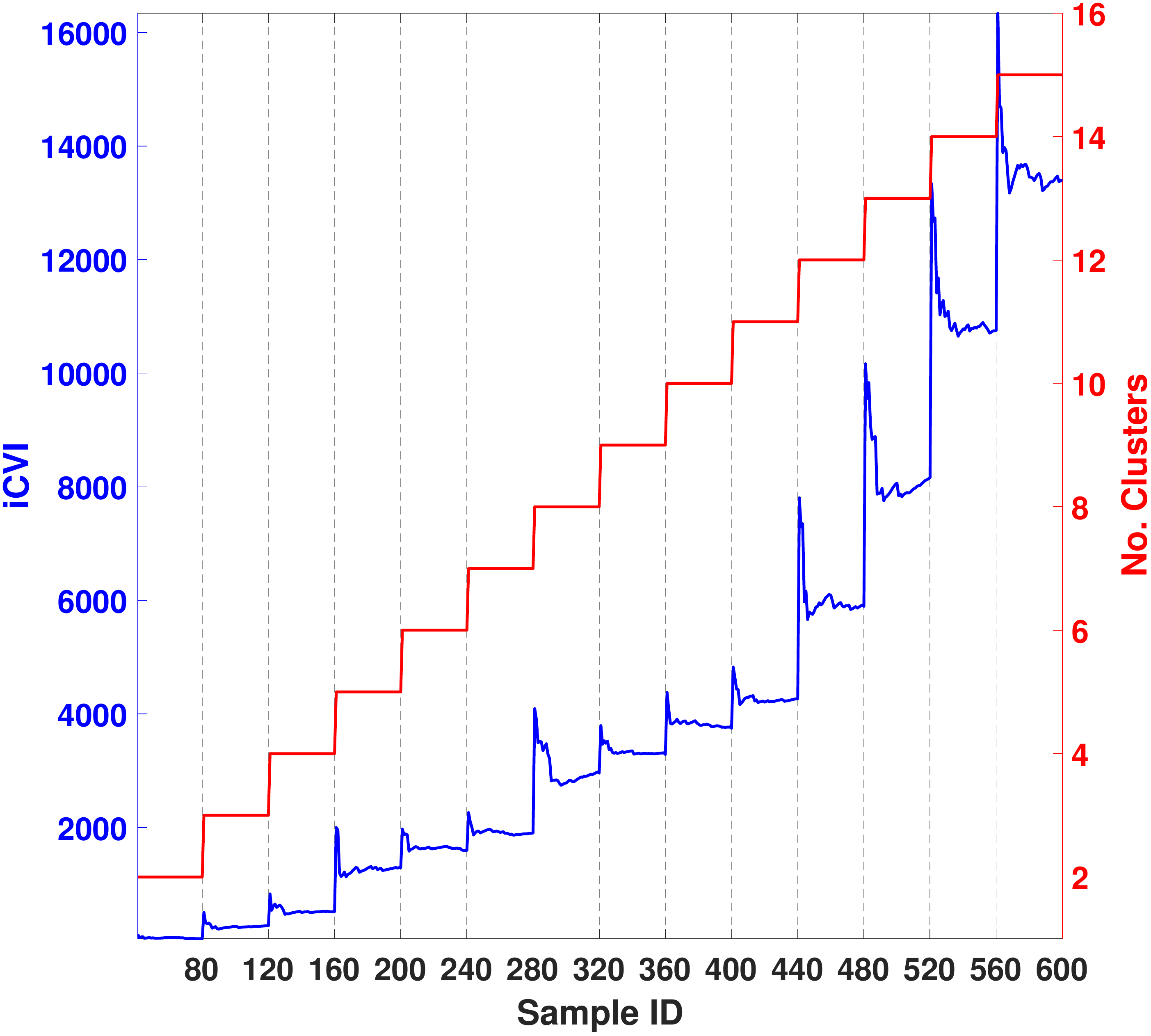}
\label{Fig:irH1}}
\hfil
\subfloat[iNI]{\includegraphics[width=\szxb\columnwidth]{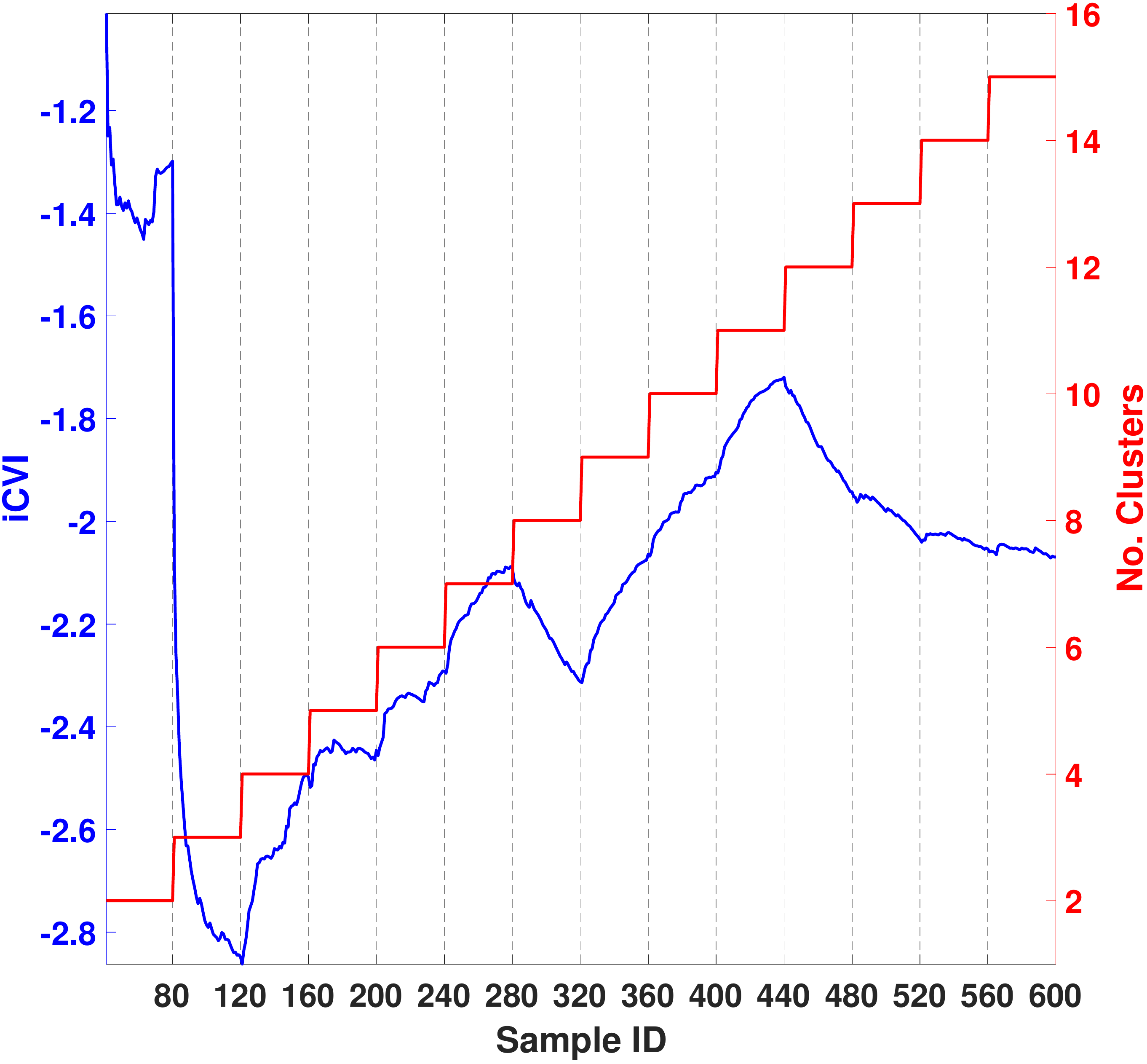}
\label{Fig:iNI1}}
}
\centerline{
\subfloat[iXB]{\includegraphics[width=\szxb\columnwidth]{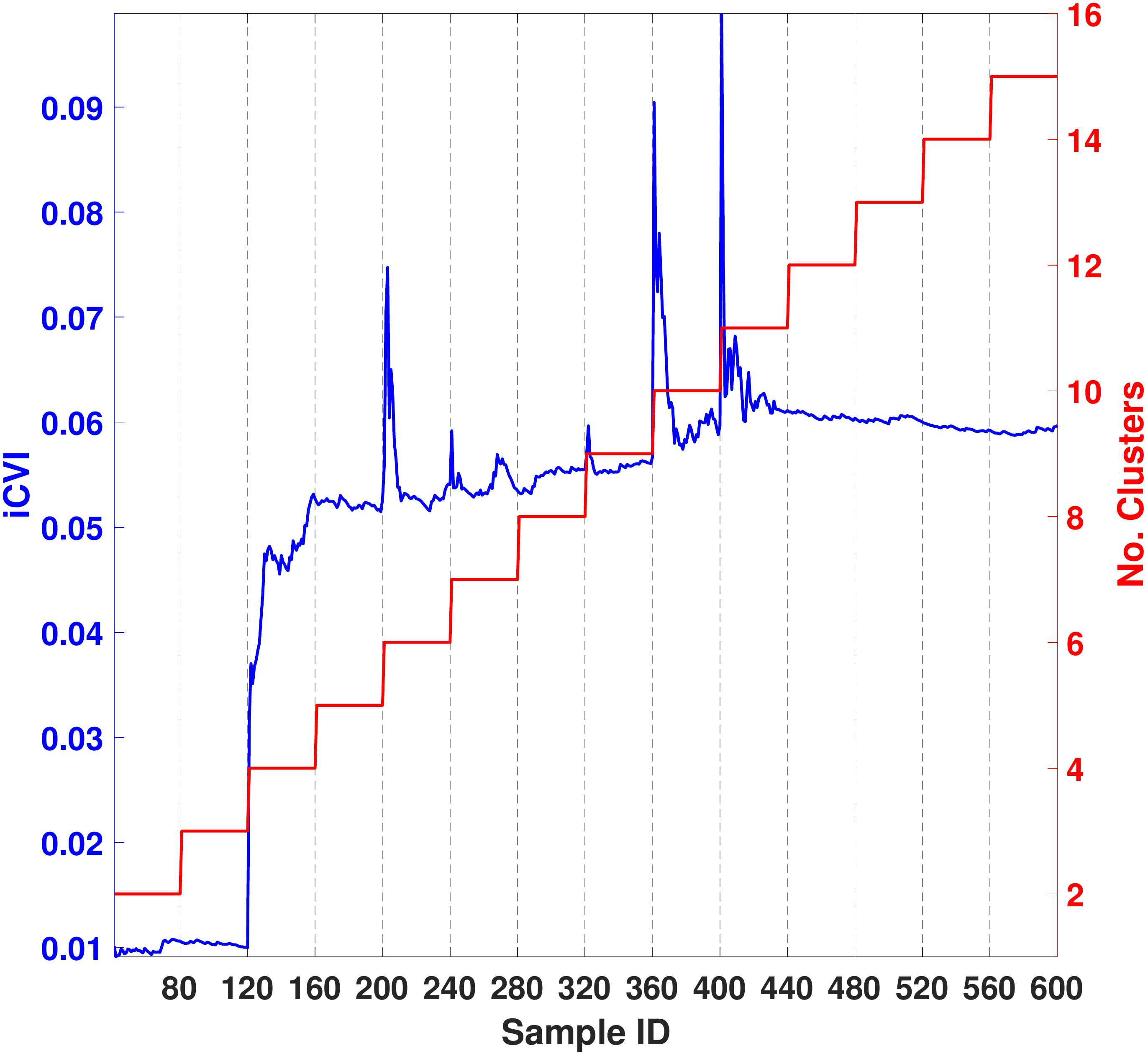}
\label{Fig:iXB1}}
\hfil
\subfloat[iDB]{\includegraphics[width=\szxb\columnwidth]{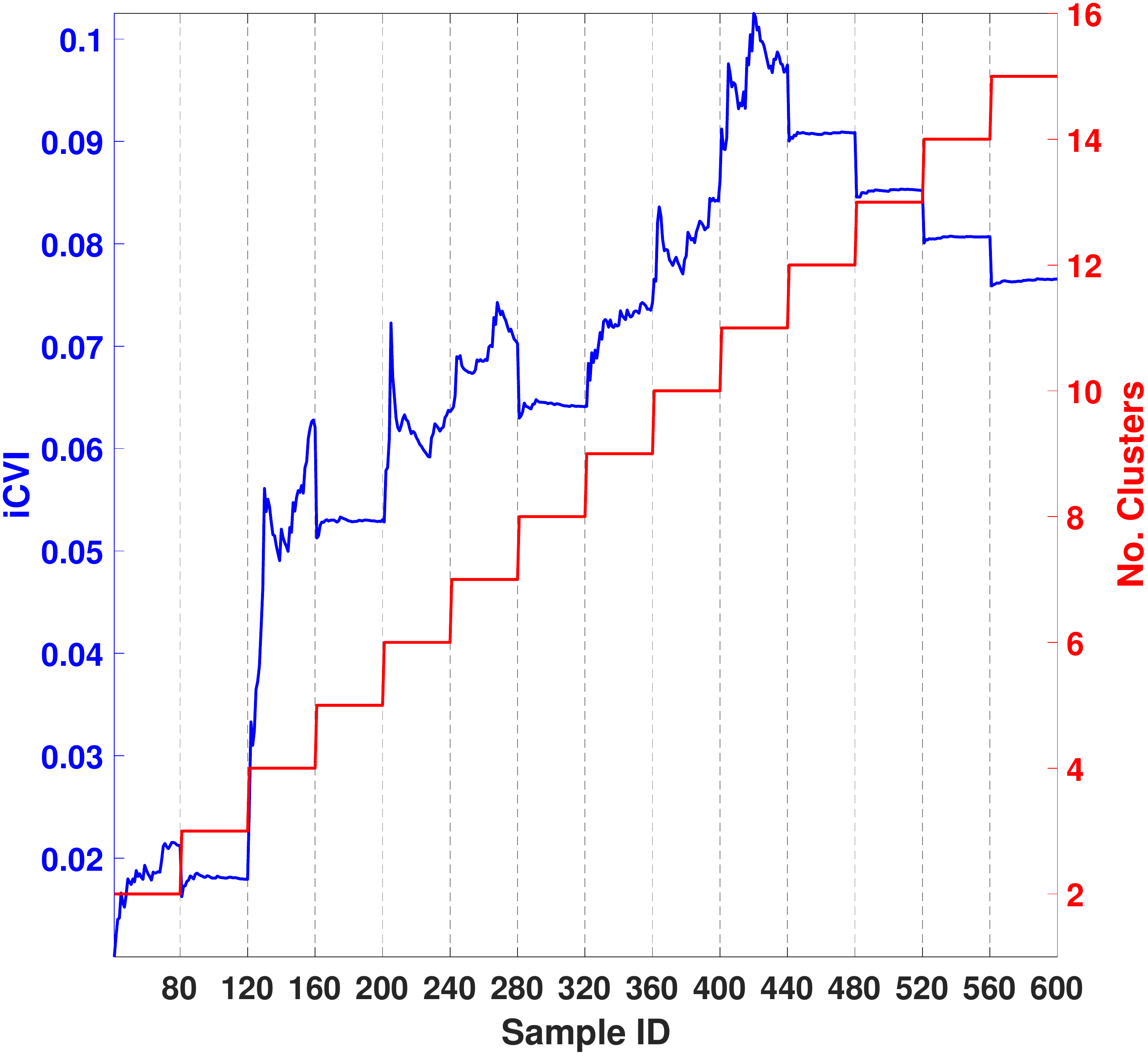}
\label{Fig:iDB1}}
\hfil
\subfloat[PS]{\includegraphics[width=\szxb\columnwidth]{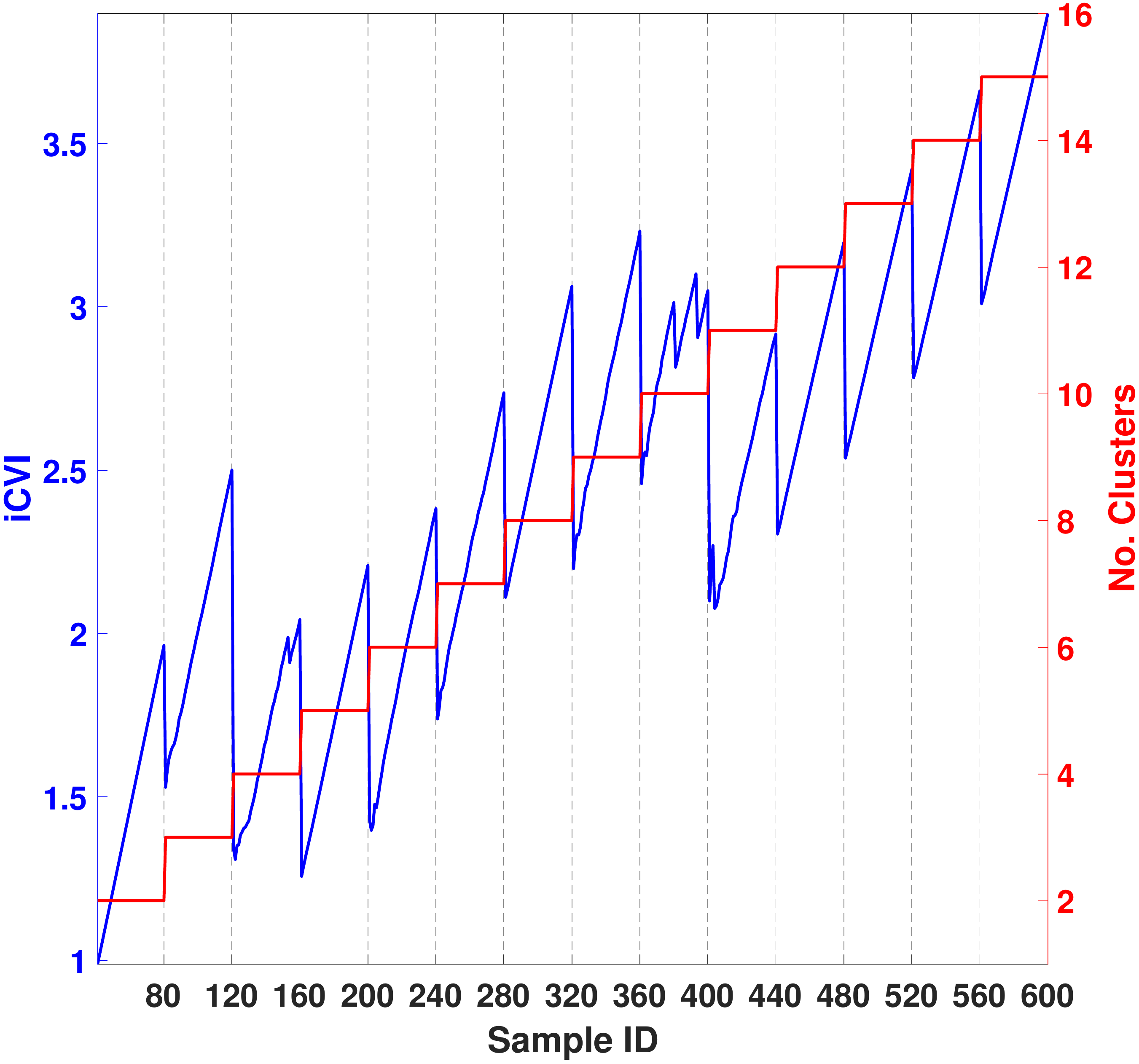}
\label{Fig:PS1}}
}
\caption{(a) A high-quality partition of the data set \textit{R15} by fuzzy ART-based clustering algorithms ($ARI = 0.9821$, $\rho=0.88$). (b) Fuzzy SMART's module A categories ($\rho_A = 0.9$) and CONNvis~\cite{tasdemir2009} (thicker and darker lines indicate stronger connections). (c)-(l) Behavior of the iCVIs (blue curve) for the partition in (a). The number of clusters is tracked by the step-like red curve. The dashed vertical lines represent the limits between two consecutive clusters (ground truth), \ie~samples before a line belong to one cluster whereas samples after it belong to another.}
\label{Fig:R15HQ}
\end{figure}
\begin{figure}[!hp]
\centerline{
\subfloat[Data partition ]{\includegraphics[width=\szC\columnwidth]{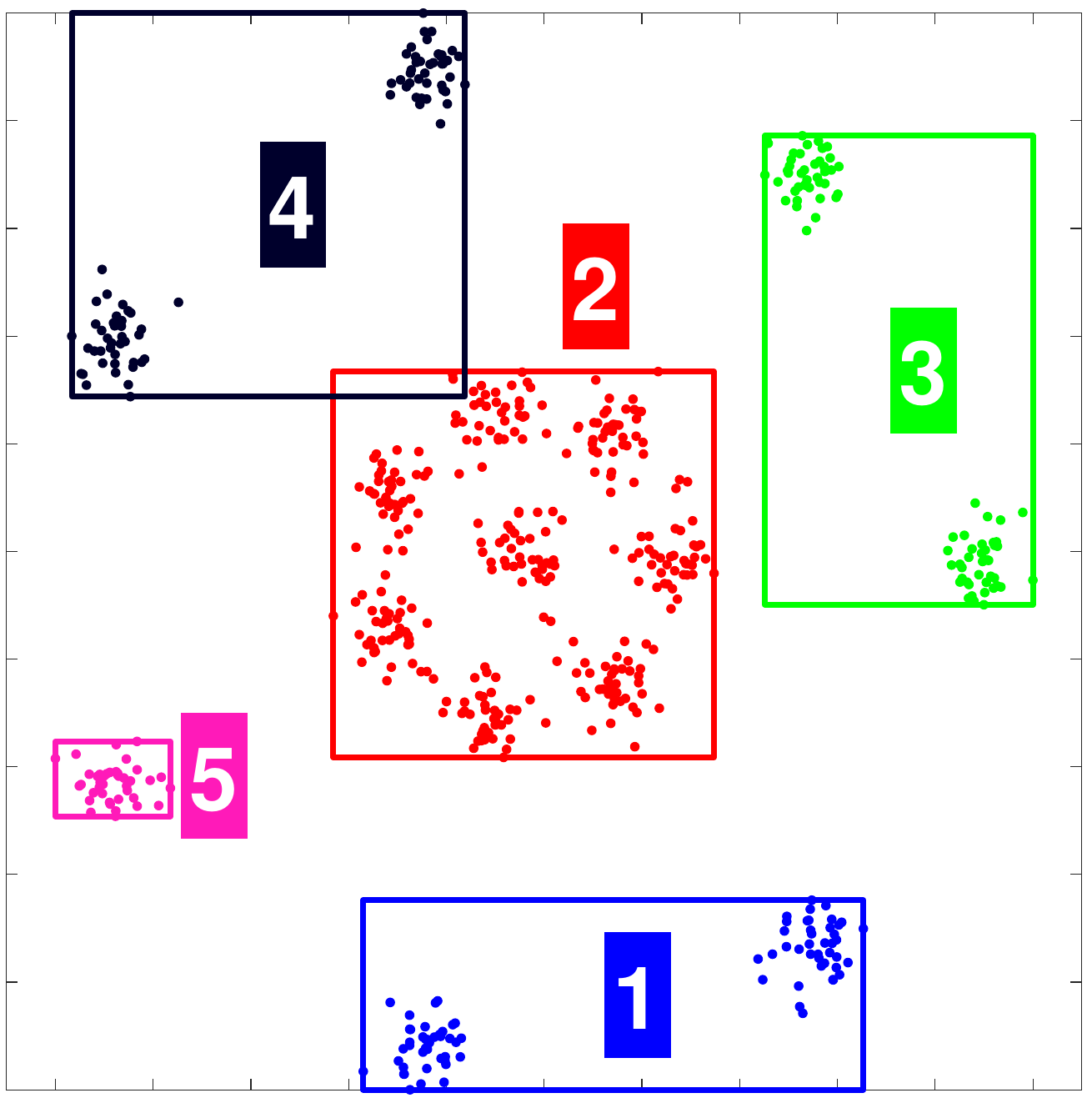}
\label{Fig:partition2}}
\hfil
\subfloat[Fuzzy SMART's module A connectivity]{\includegraphics[width=\szP\columnwidth]{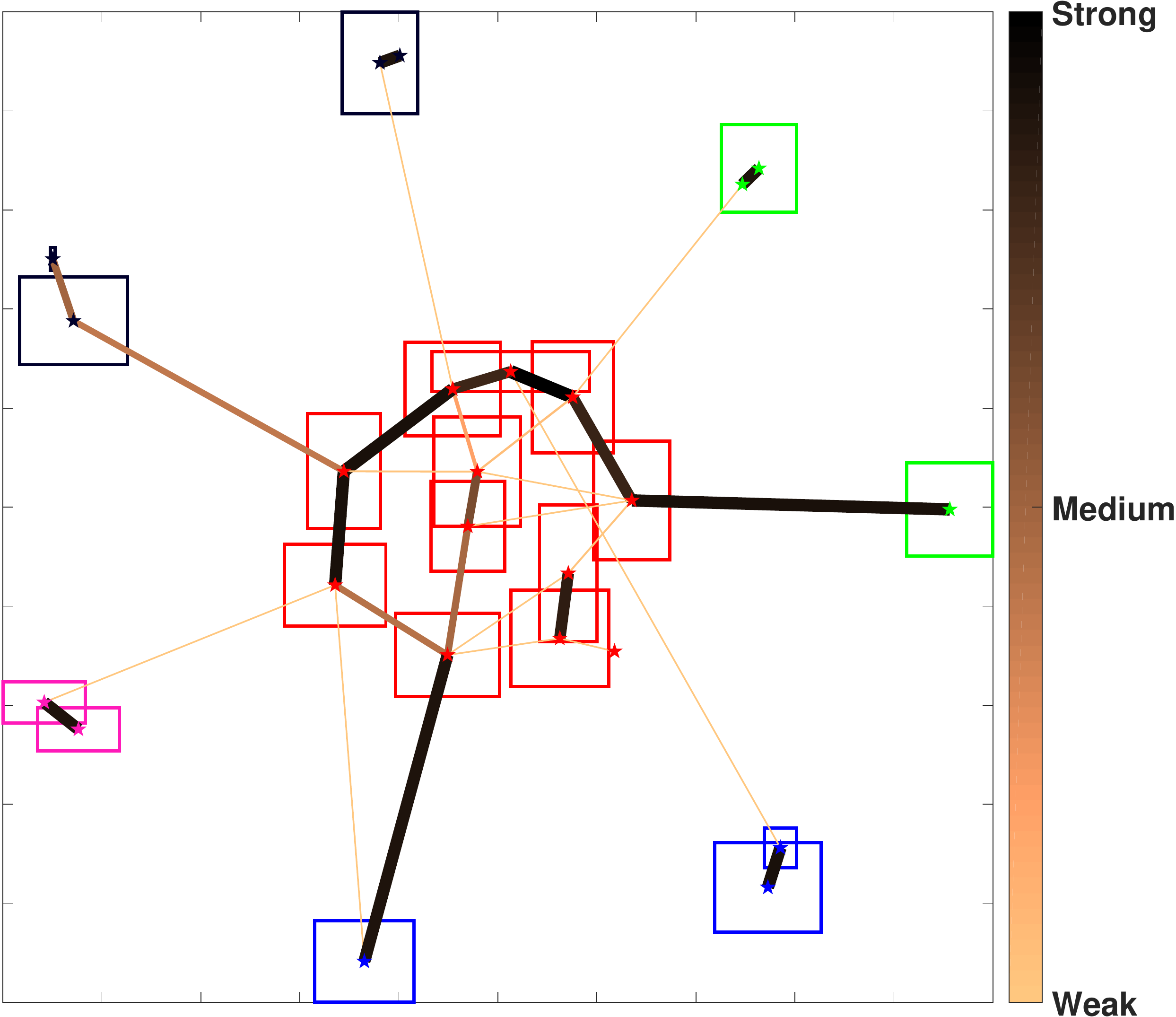}
\label{Fig:SMART_A_2}}
\hfil
\subfloat[iConn\_Index]{\includegraphics[width=\szxb\columnwidth]{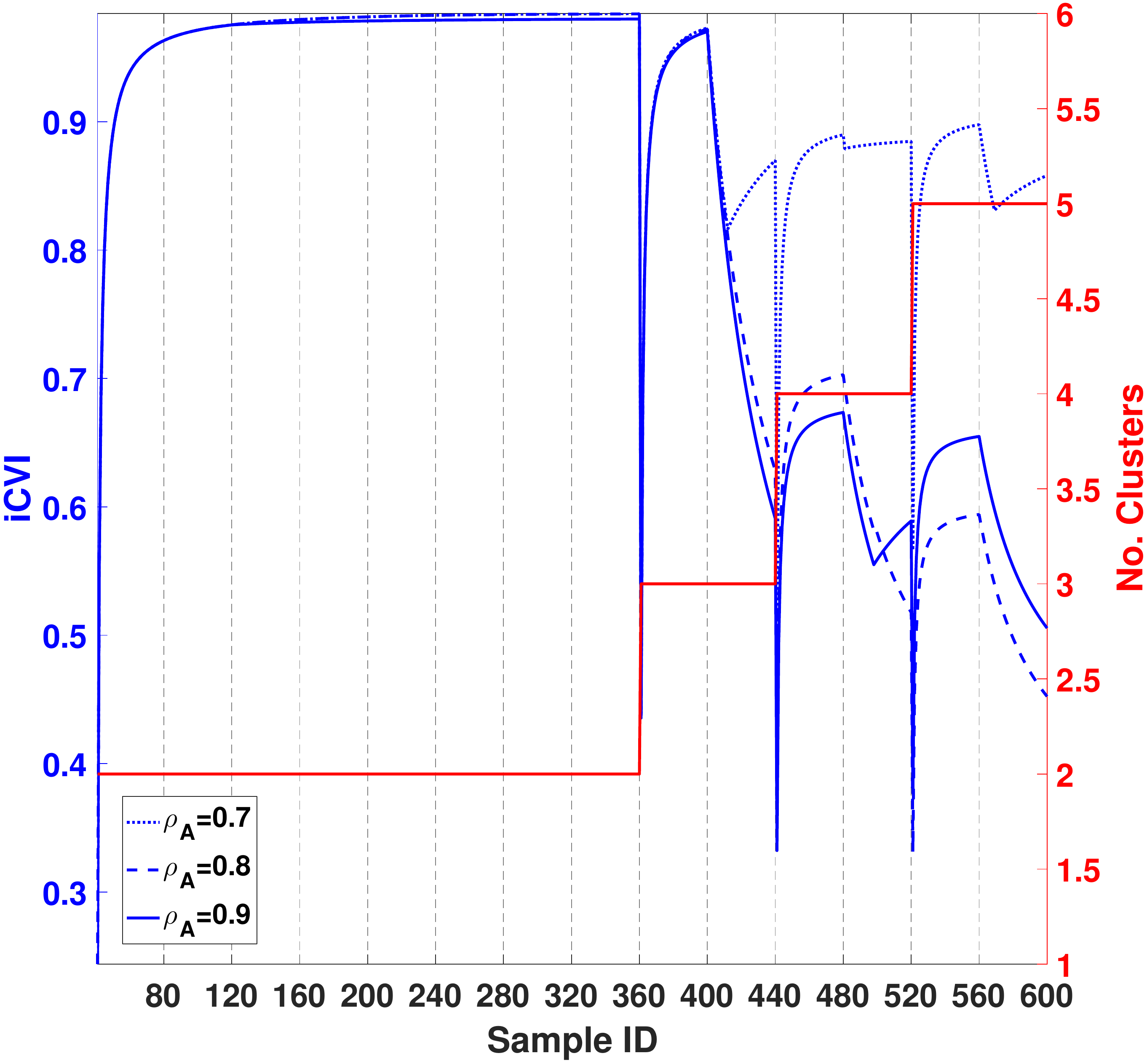}
\label{Fig:iConn2}}
}
\centerline{
\subfloat[iCH]{\includegraphics[width=\szxb\columnwidth]{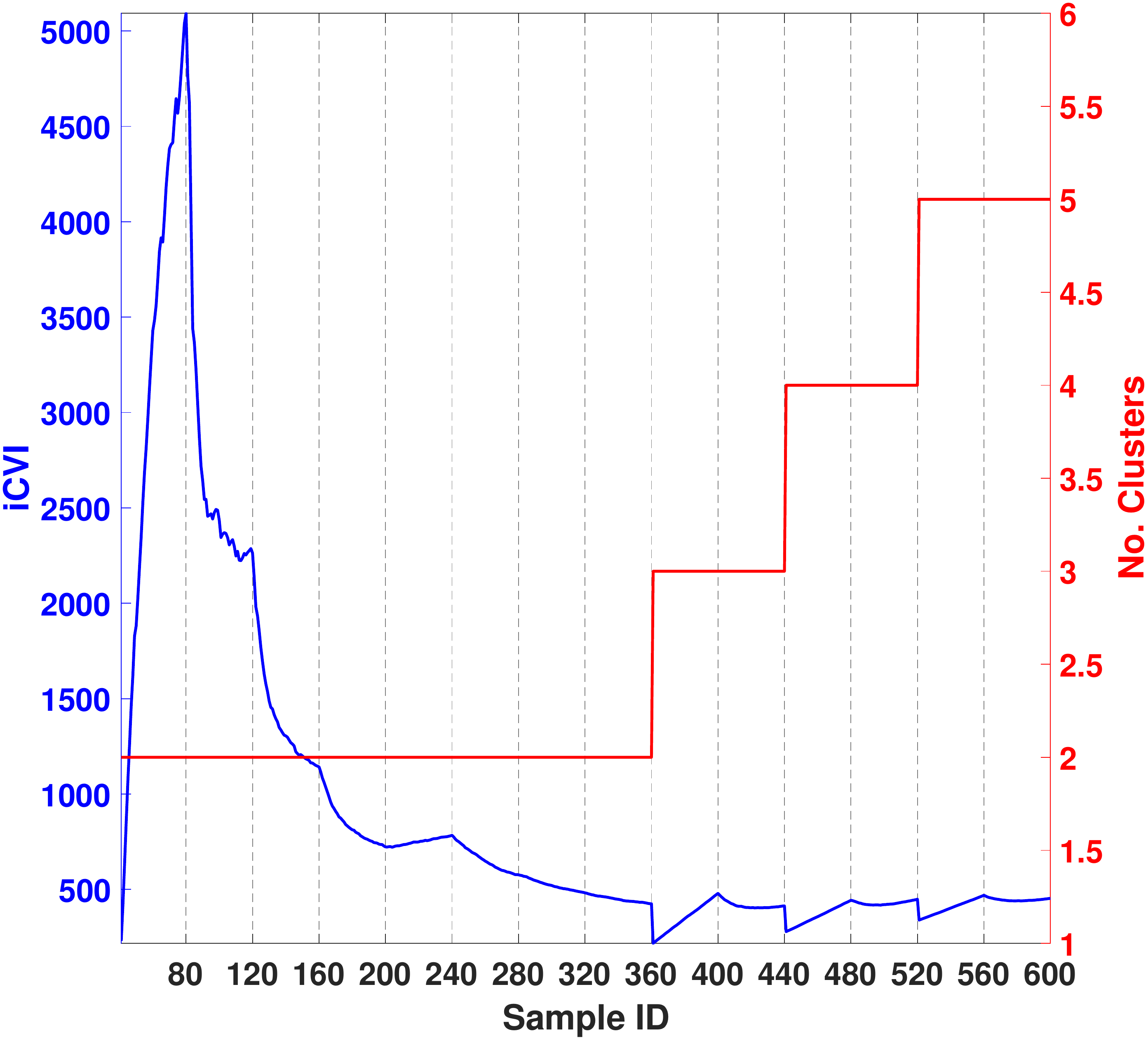}
\label{Fig:iCH2}}
\hfil
\subfloat[iSIL]{\includegraphics[width=\szxb\columnwidth]{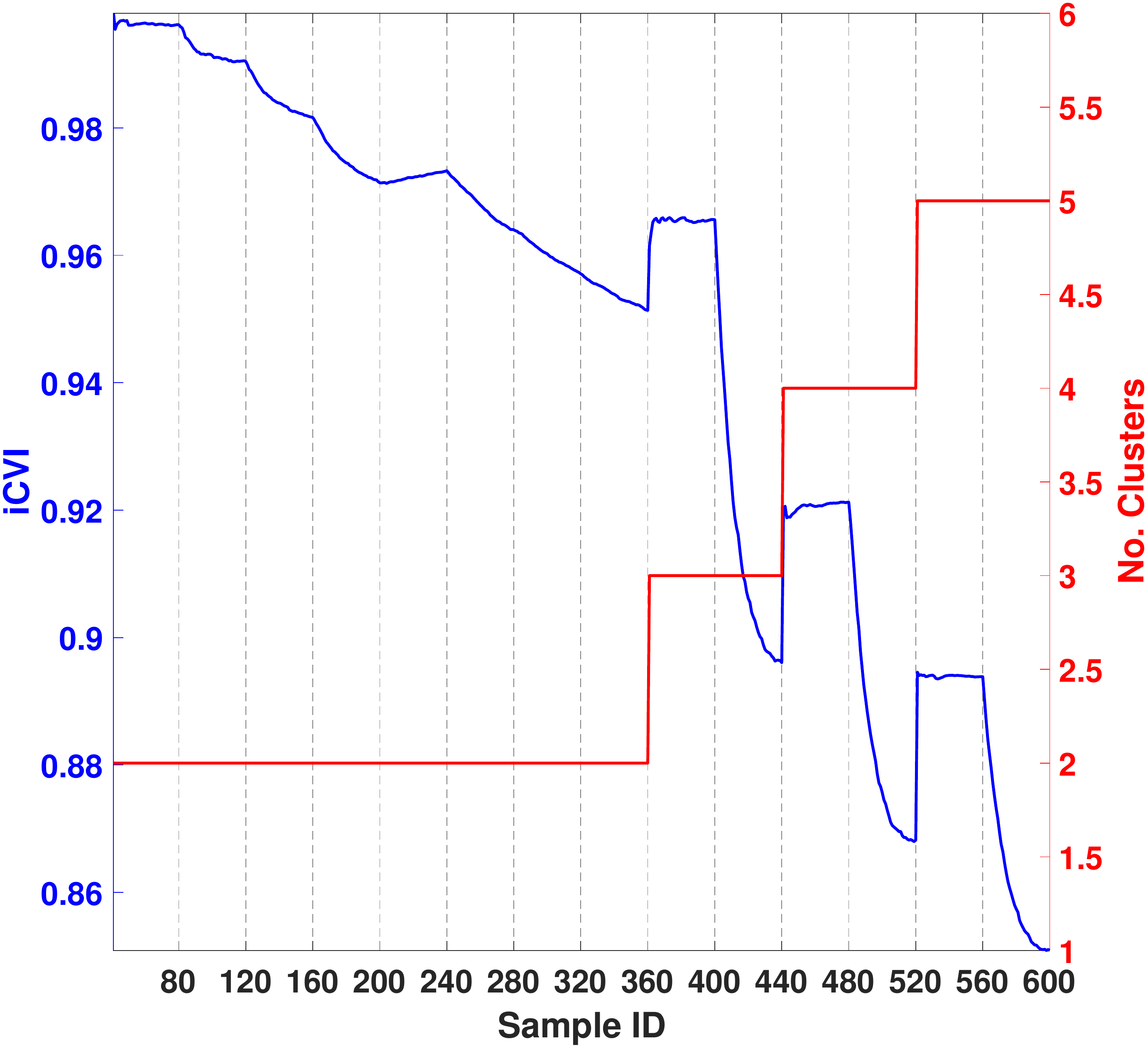}
\label{Fig:iSIL2}}
\hfil
\subfloat[iPBM]{\includegraphics[width=\szxb\columnwidth]{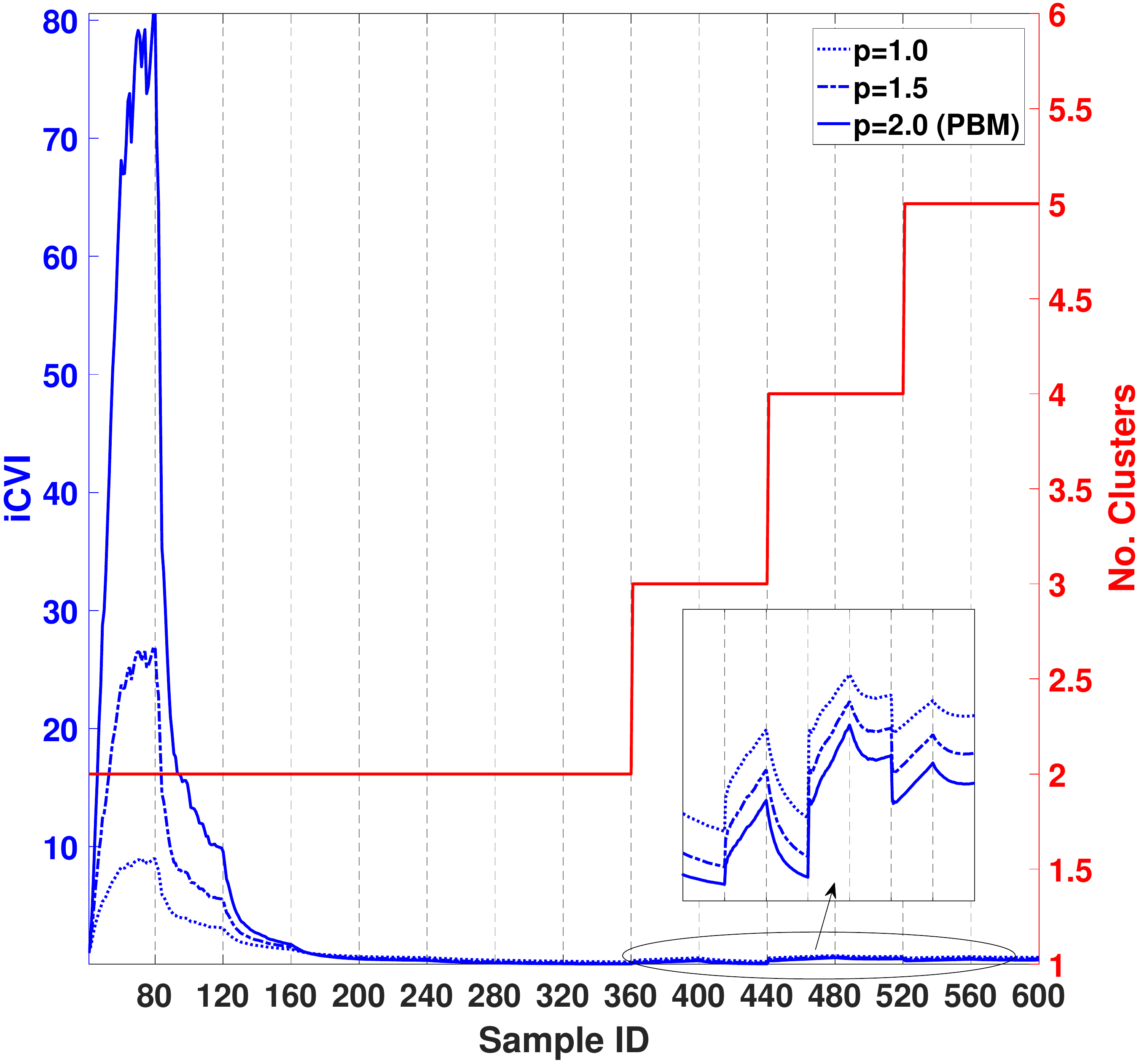}
\label{Fig:iPBM2}}
}
\centerline{
\subfloat[irCIP]{\includegraphics[width=\szxb\columnwidth]{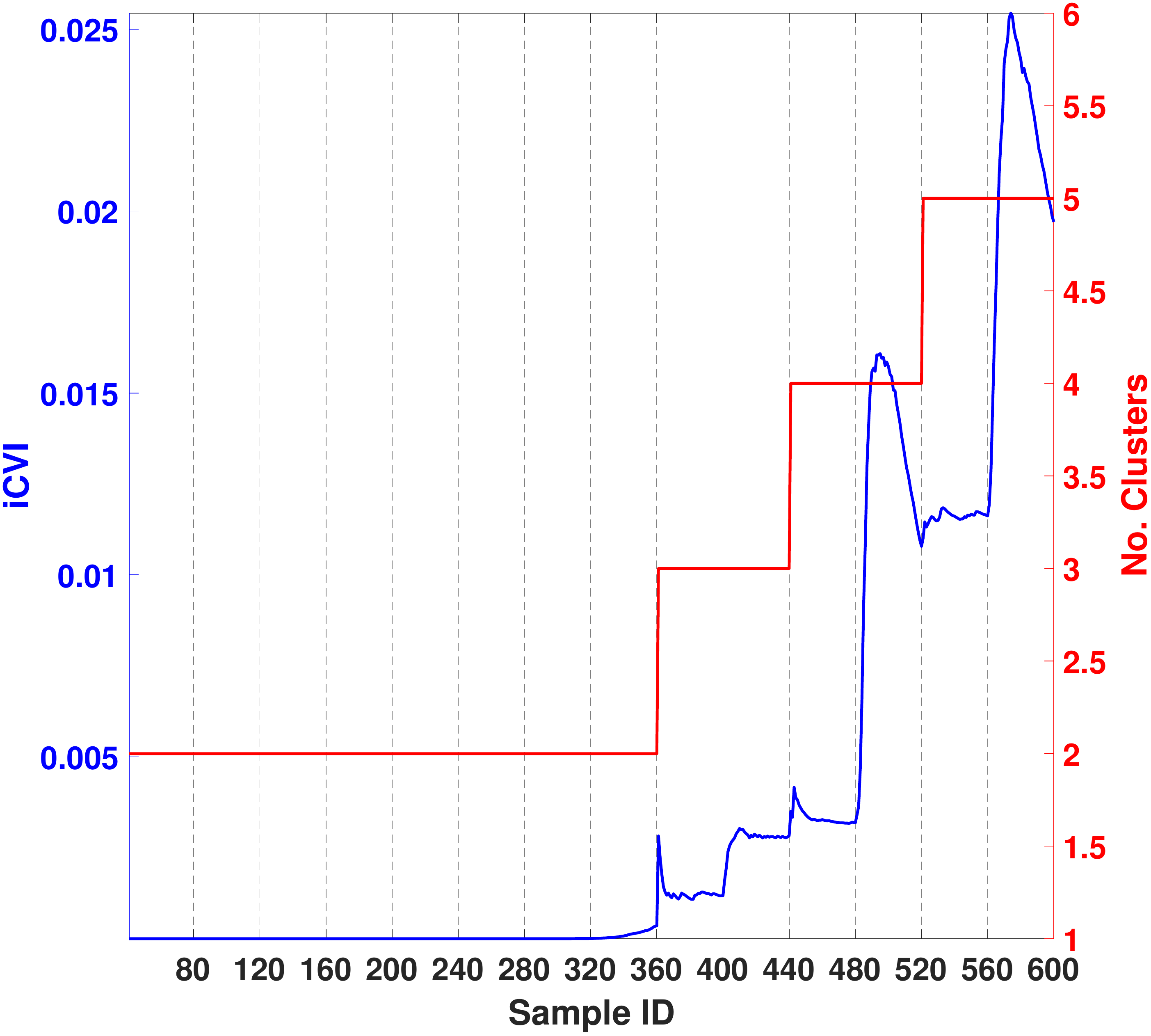}
\label{Fig:irCIP2}}
\hfil
\subfloat[irH]{\includegraphics[width=\szxb\columnwidth]{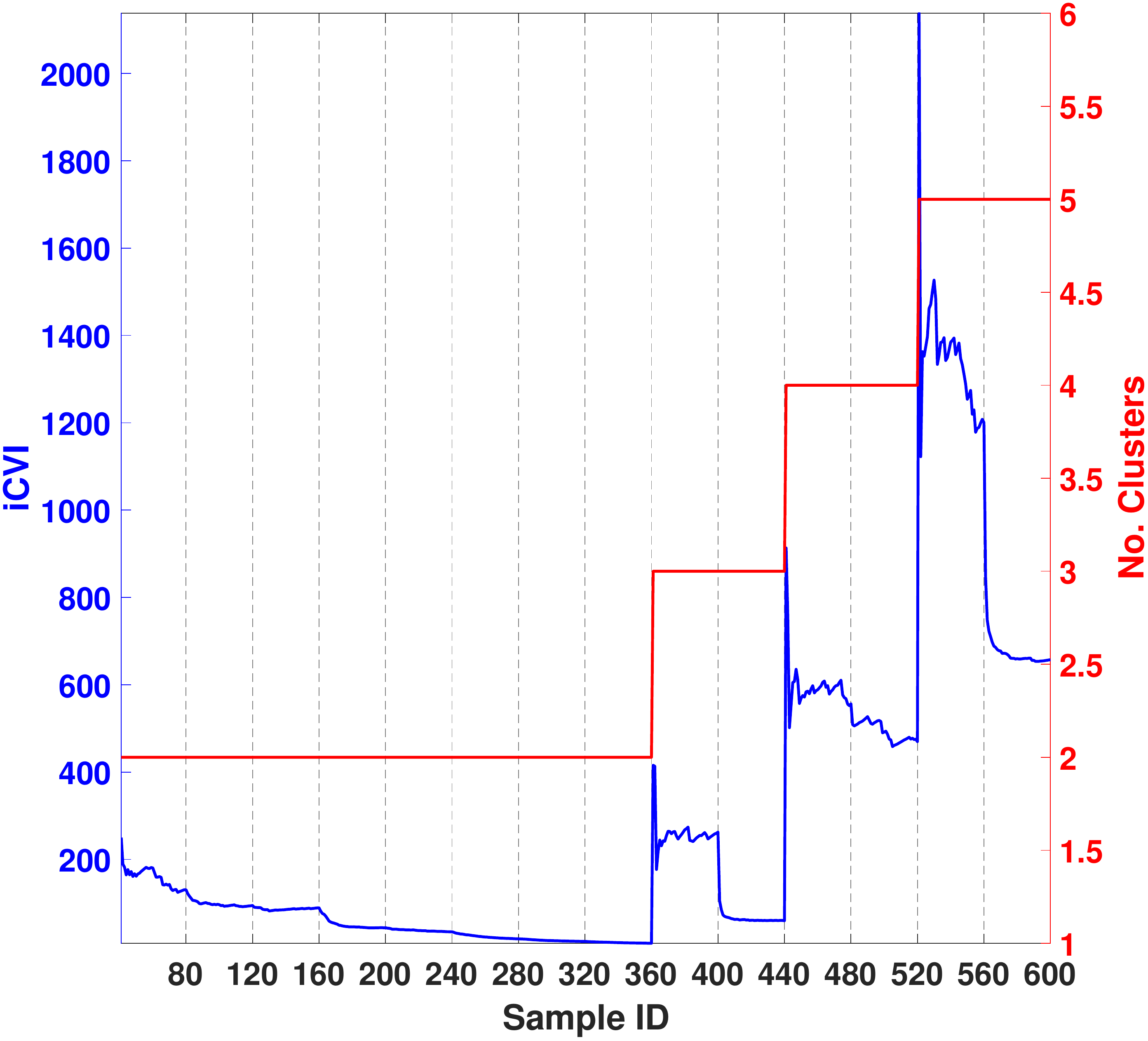}
\label{Fig:irH2}}
\hfil
\subfloat[iNI]{\includegraphics[width=\szxb\columnwidth]{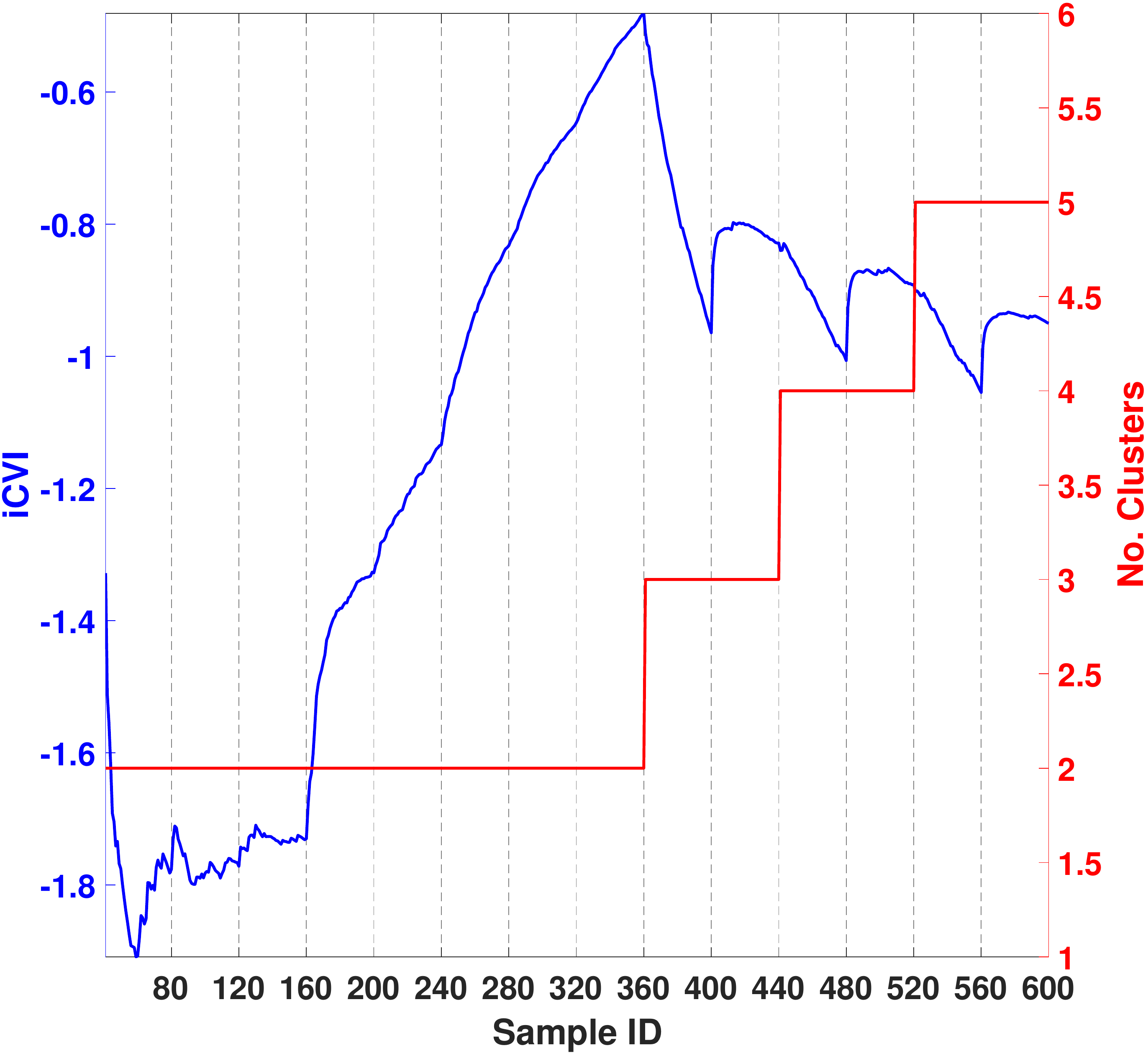}
\label{Fig:iNI2}}
}
\centerline{
\subfloat[iXB]{\includegraphics[width=\szxb\columnwidth]{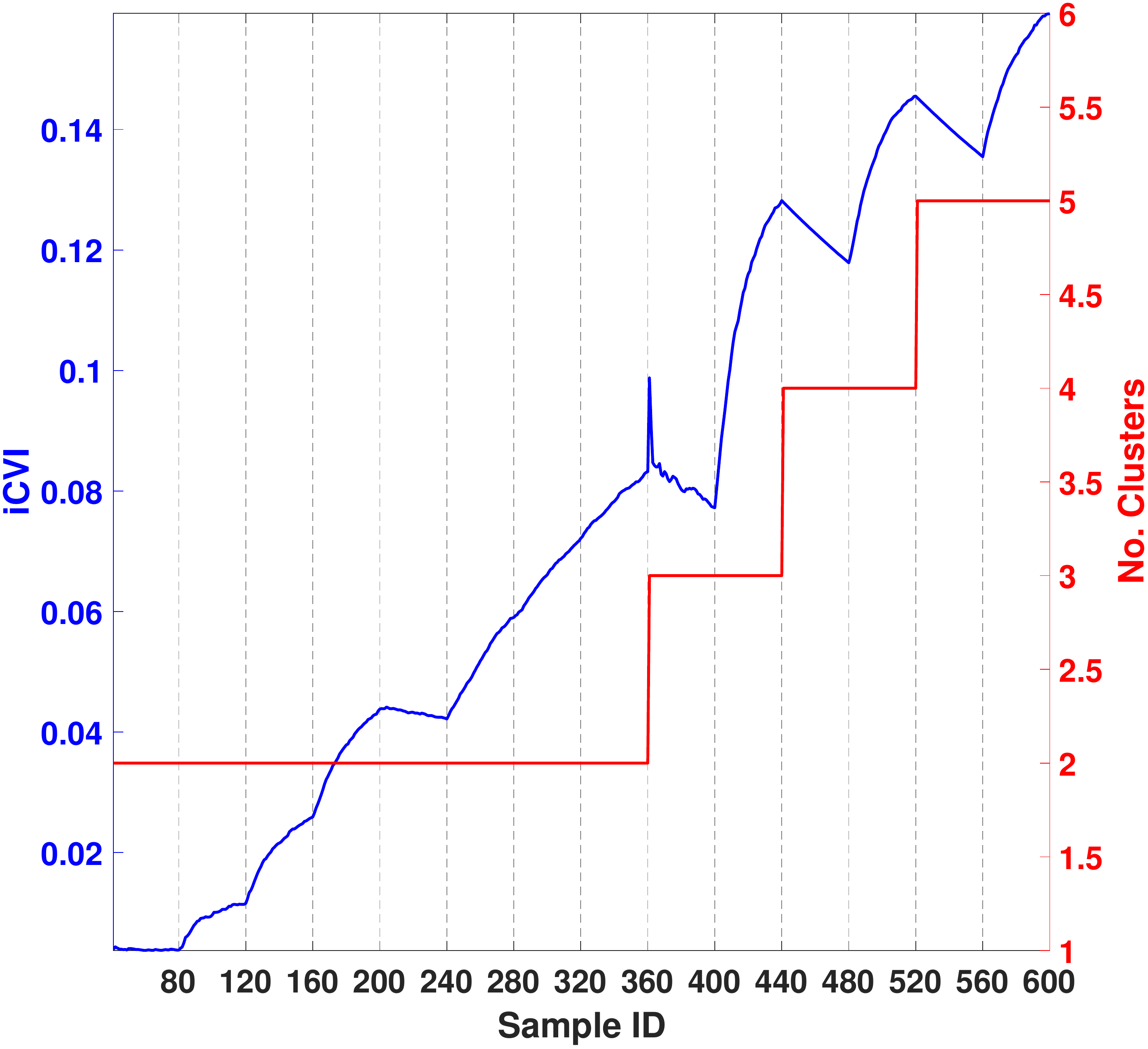}
\label{Fig:iXB2}}
\hfil
\subfloat[iDB]{\includegraphics[width=\szxb\columnwidth]{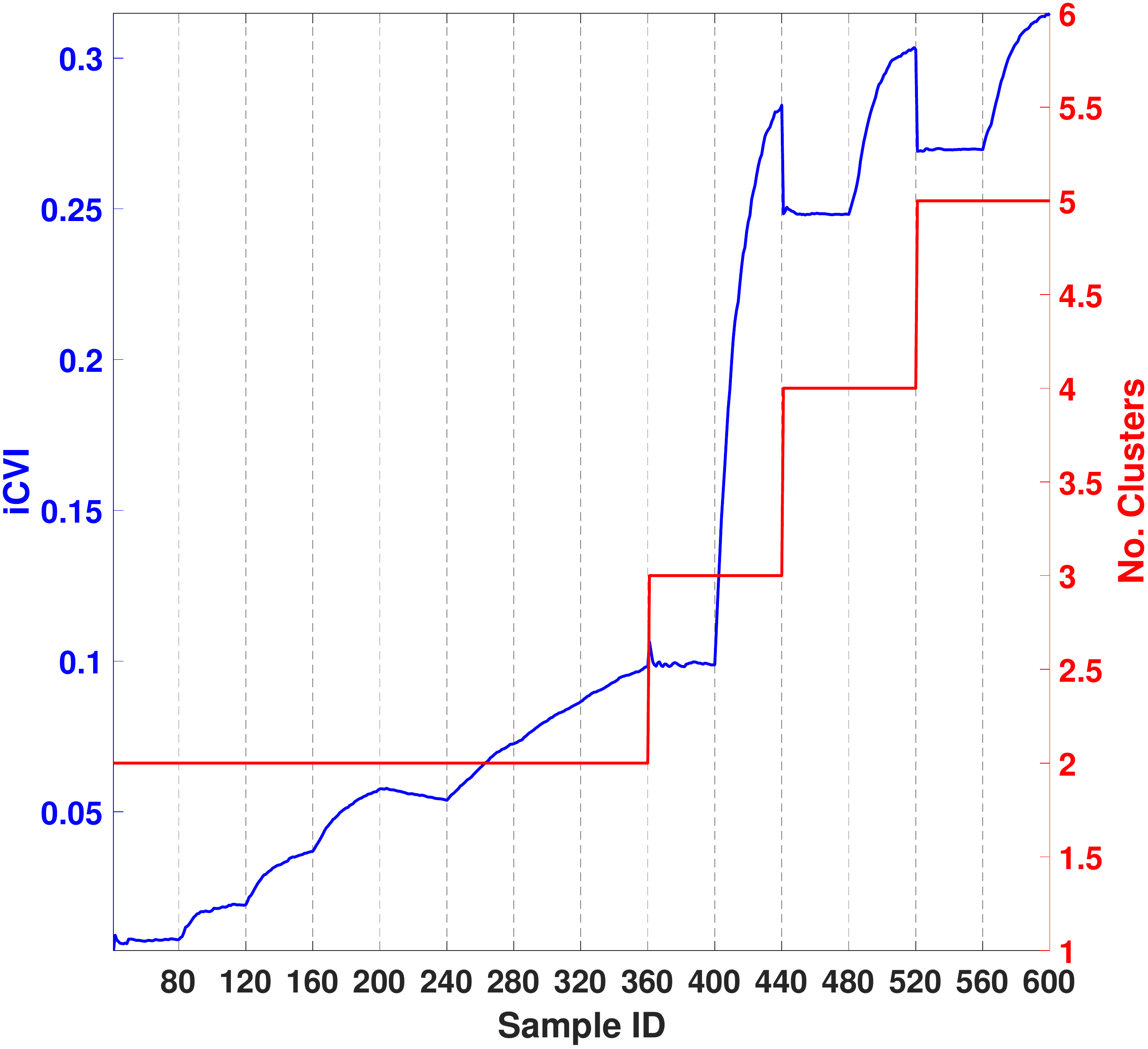}
\label{Fig:iDB2}}
\hfil
\subfloat[PS]{\includegraphics[width=\szxb\columnwidth]{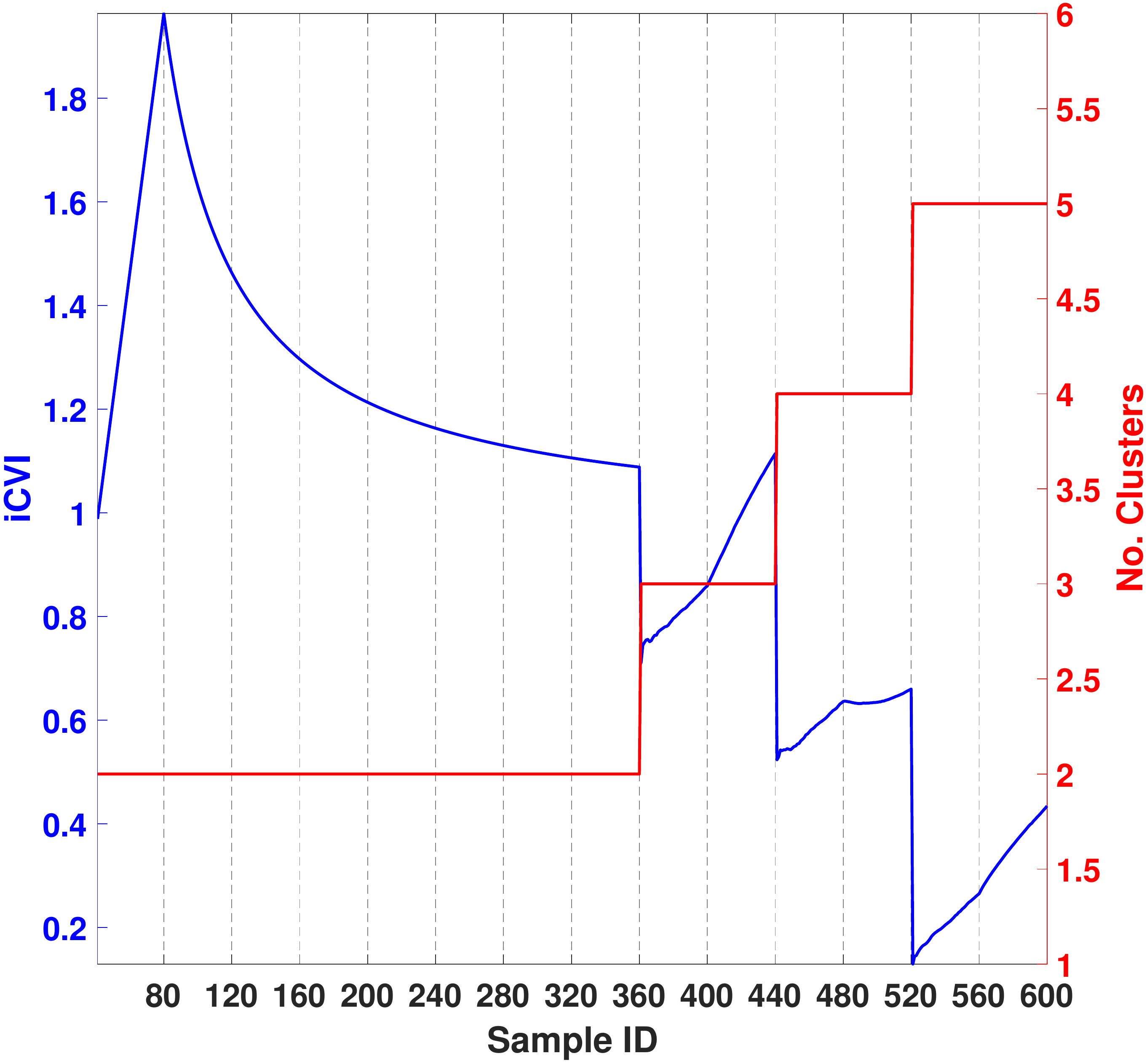}
\label{Fig:PS2}}
}
\caption{(a) An under-partition of the data set \textit{R15} by fuzzy ART-based clustering algorithms ($ARI = 0.2371$, $\rho=0.61$). (b) Fuzzy SMART's module A categories ($\rho_A = 0.9$) and CONNvis~\cite{tasdemir2009} (thicker and darker lines indicate stronger connections). (c)-(l) Behavior of the iCVIs (blue curve) for the partition in (a). The number of clusters is tracked by the step-like red curve. The dashed vertical lines represent the limits between two consecutive clusters (ground truth), \ie~samples before a line belong to one cluster whereas samples after it belong to another.}
\label{Fig:R15UP}
\end{figure}
\begin{figure}[!hp]
\centerline{
\subfloat[Data partition ]{\includegraphics[width=\szC\columnwidth]{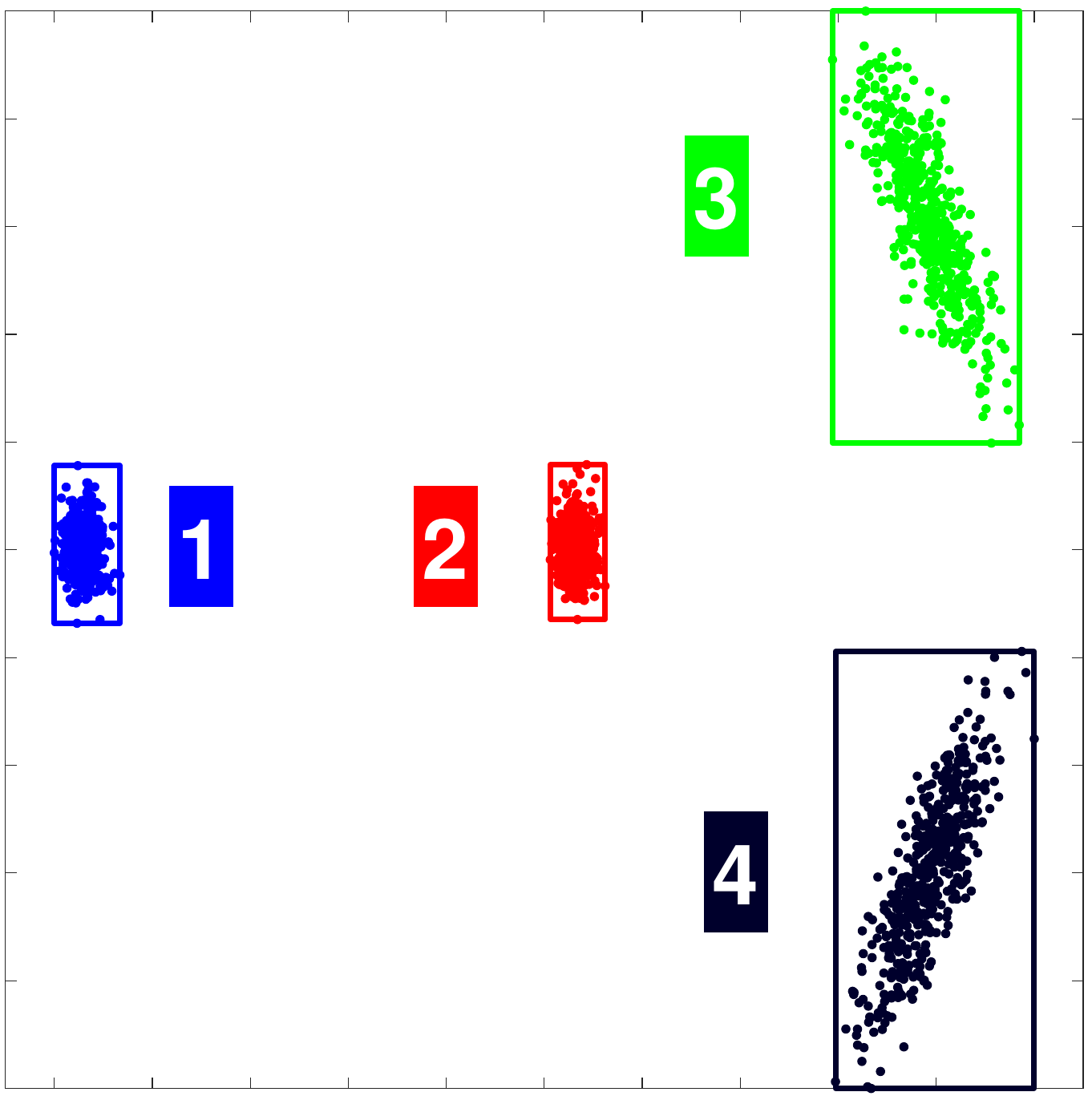}
\label{Fig:partition3}}
\hfil
\subfloat[Fuzzy SMART's module A connectivity]{\includegraphics[width=\szP\columnwidth]{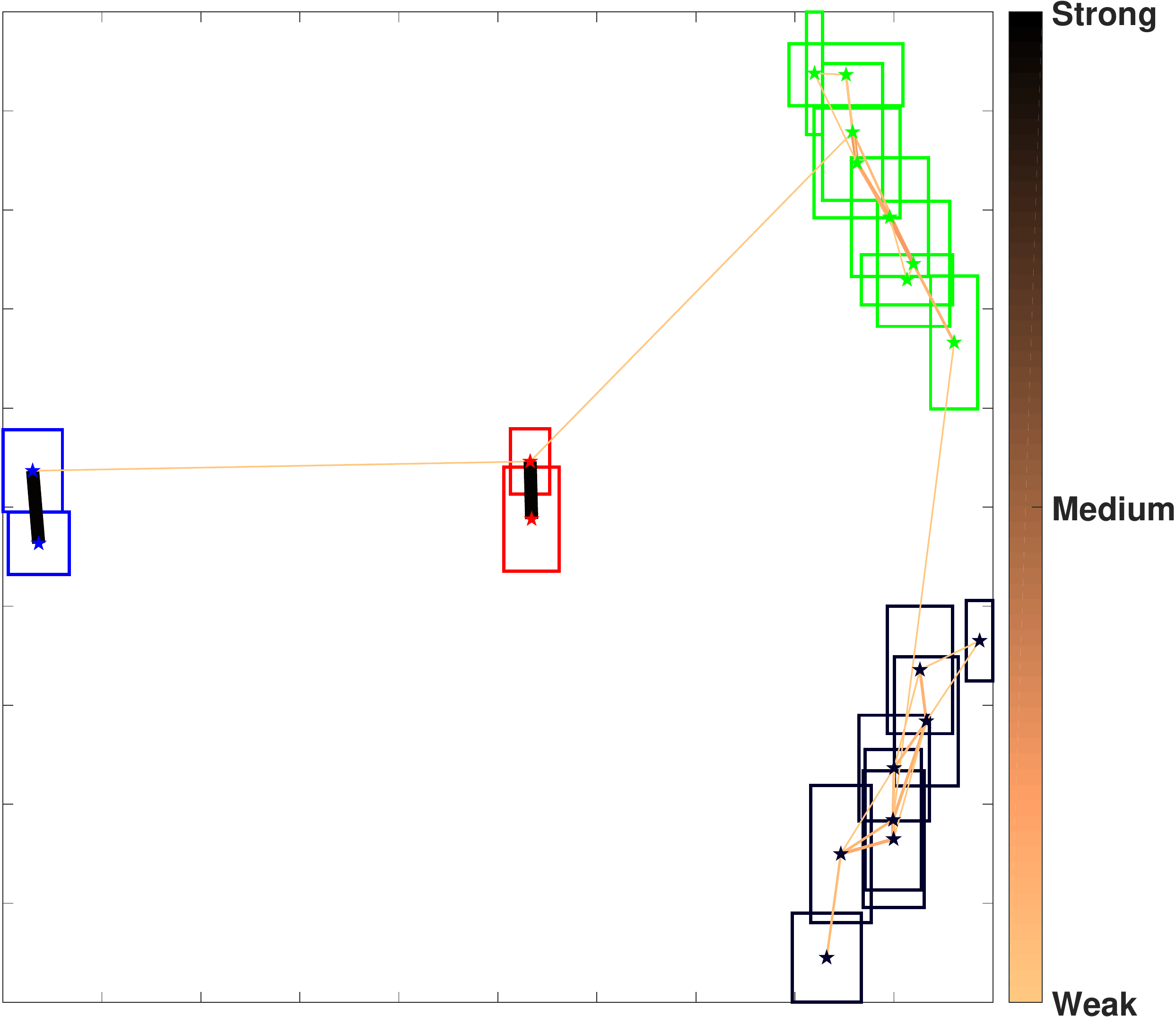}
\label{Fig:SMART_A_3}}
\hfil
\subfloat[iConn\_Index]{\includegraphics[width=\szxb\columnwidth]{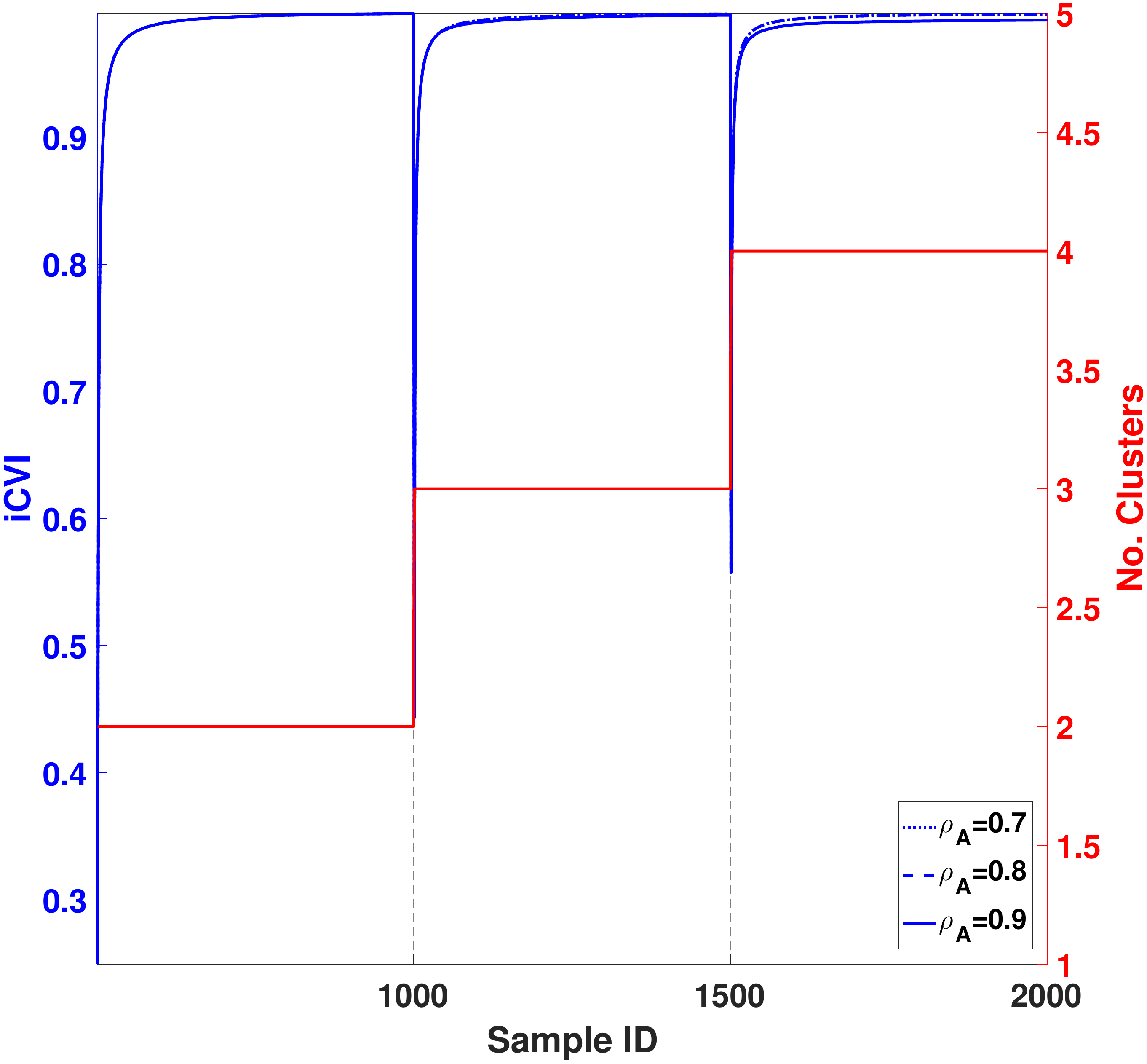}
\label{Fig:iConn3}}
}
\centerline{
\subfloat[iCH]{\includegraphics[width=\szxb\columnwidth]{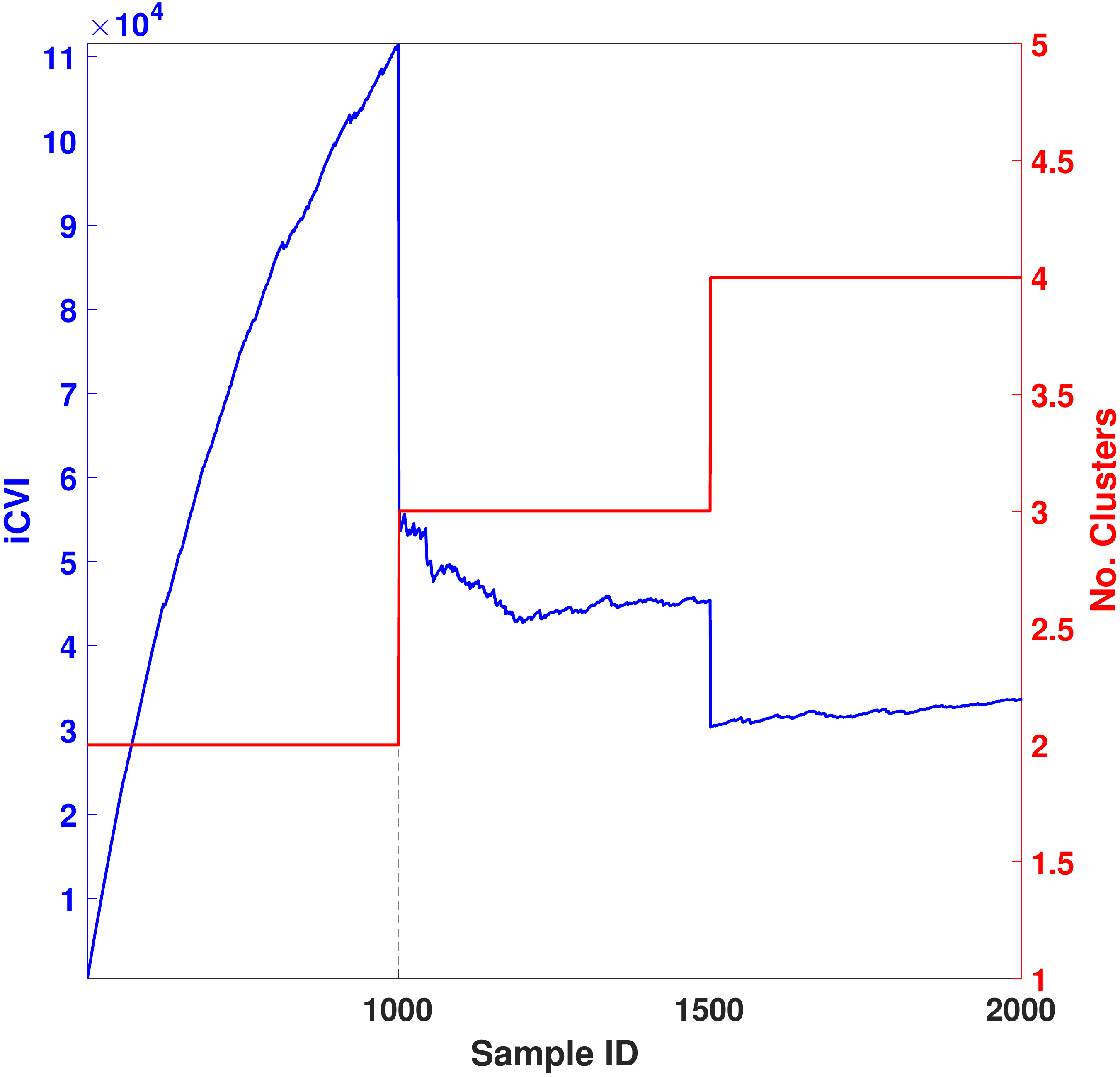}
\label{Fig:iCH3}}
\hfil
\subfloat[iSIL]{\includegraphics[width=\szxb\columnwidth]{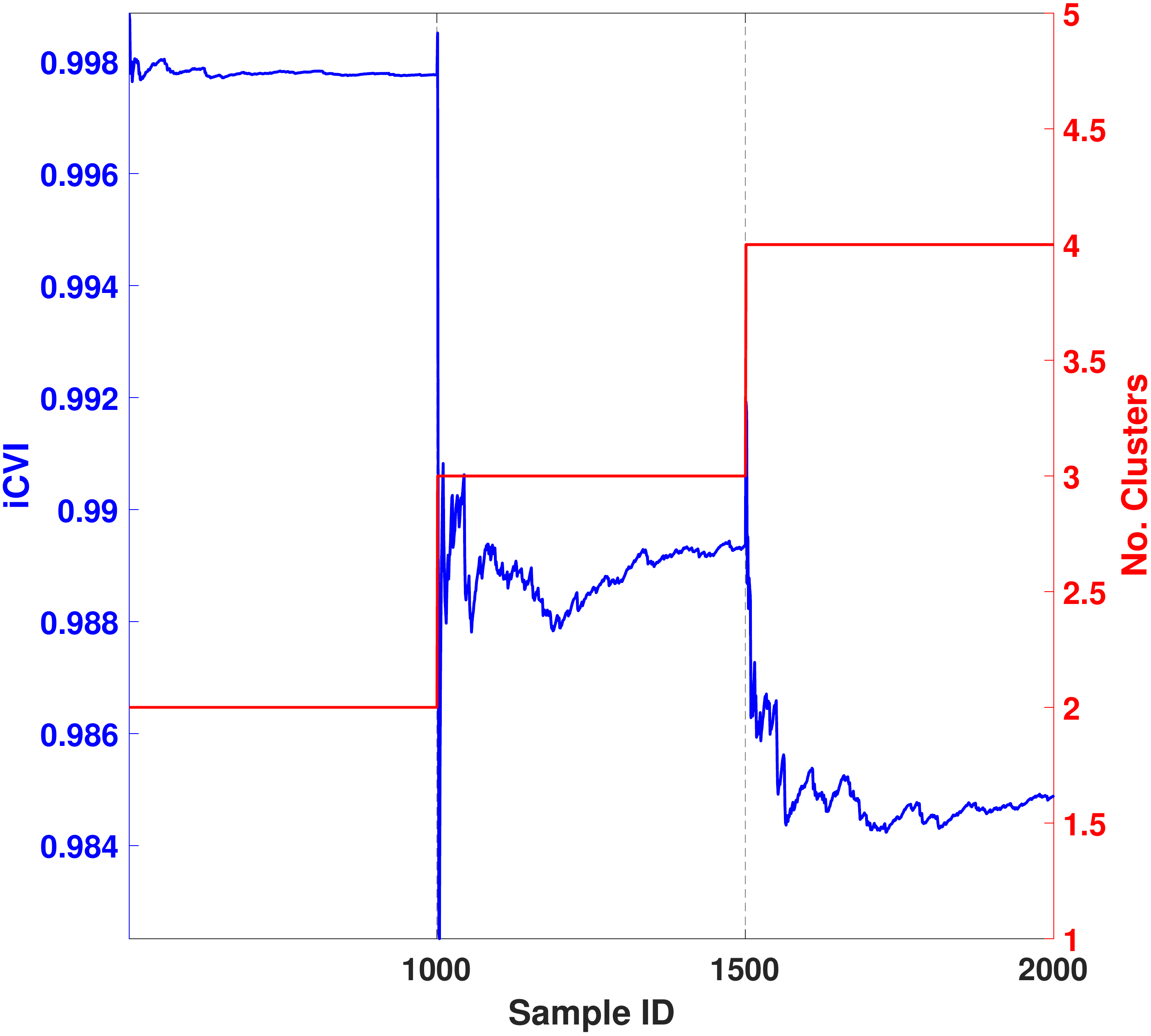}
\label{Fig:iSIL3}}
\hfil
\subfloat[iPBM]{\includegraphics[width=\szxb\columnwidth]{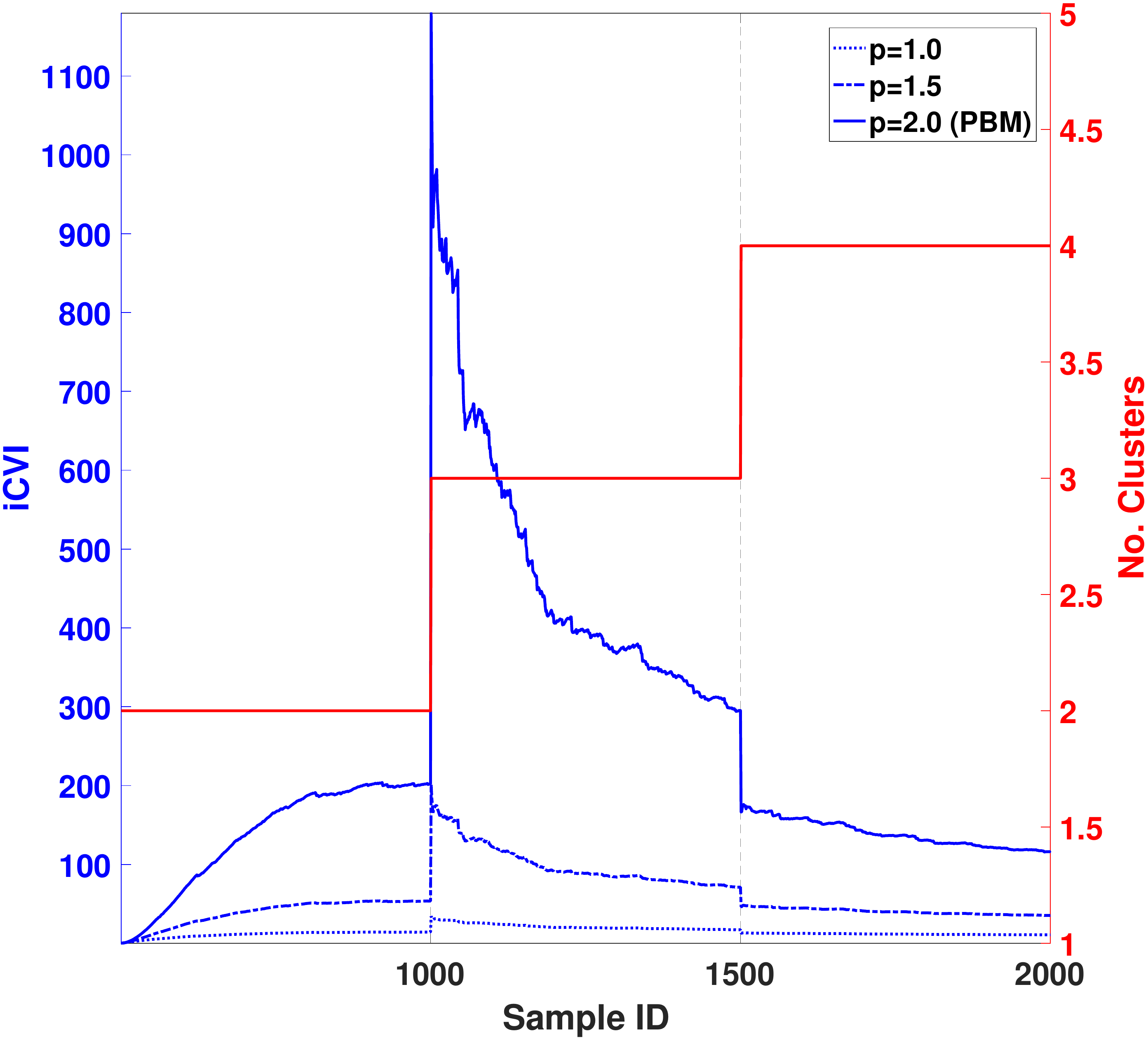}
\label{Fig:iPBM3}}
}
\centerline{
\subfloat[irCIP]{\includegraphics[width=\szxb\columnwidth]{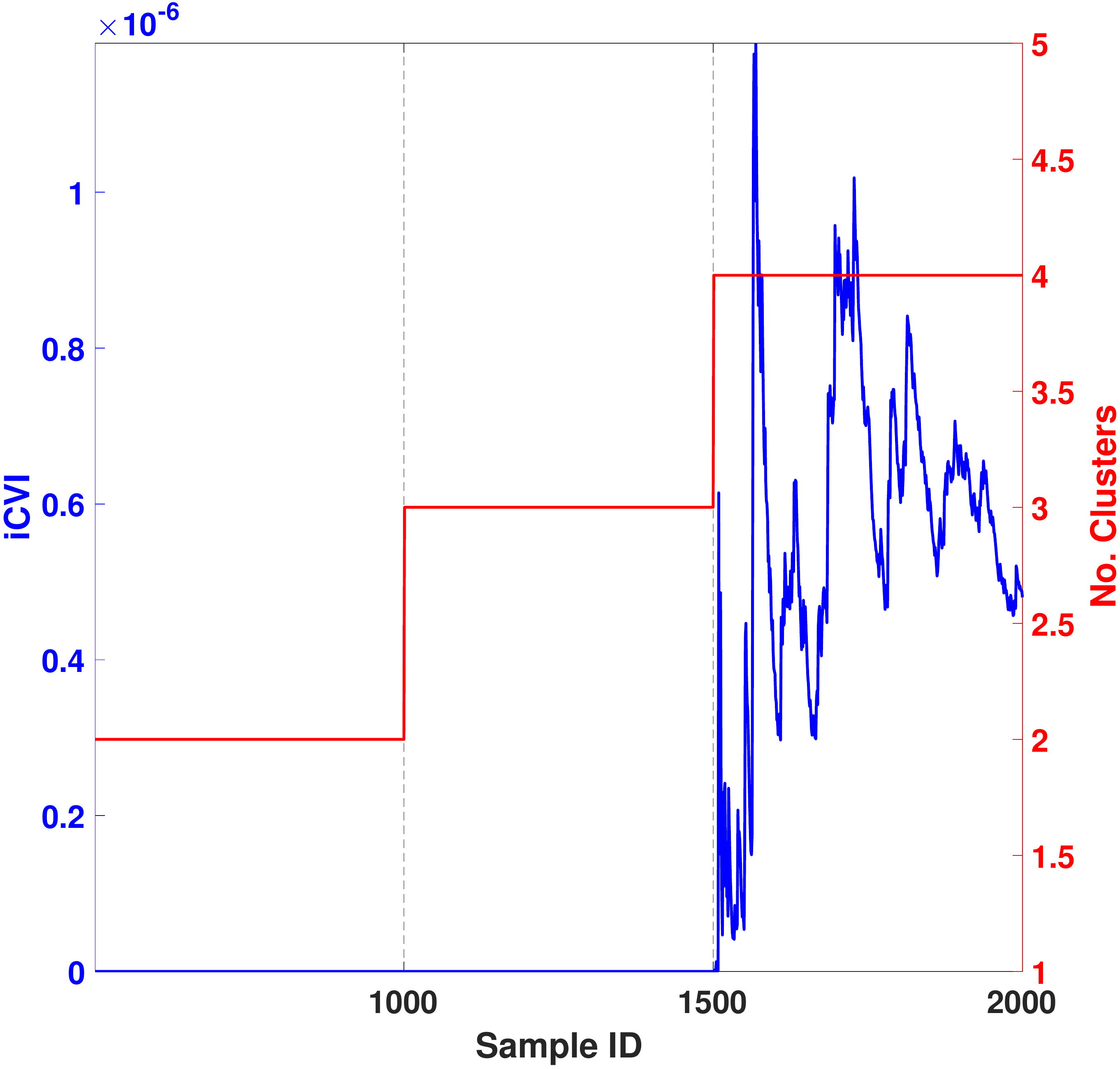}
\label{Fig:irCIP3}}
\hfil
\subfloat[irH]{\includegraphics[width=\szxb\columnwidth]{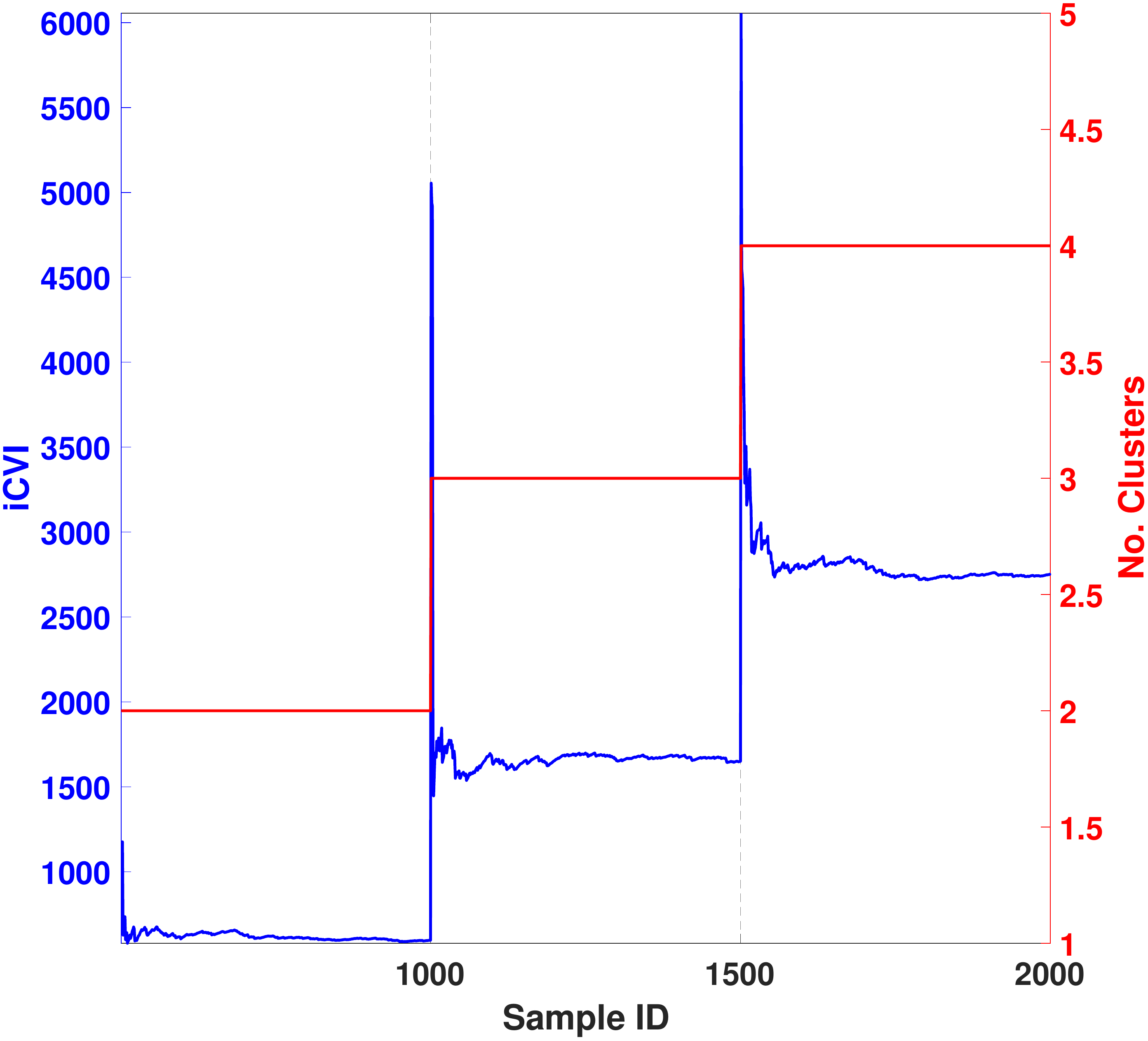}
\label{Fig:irH3}}
\hfil
\subfloat[iNI]{\includegraphics[width=\szxb\columnwidth]{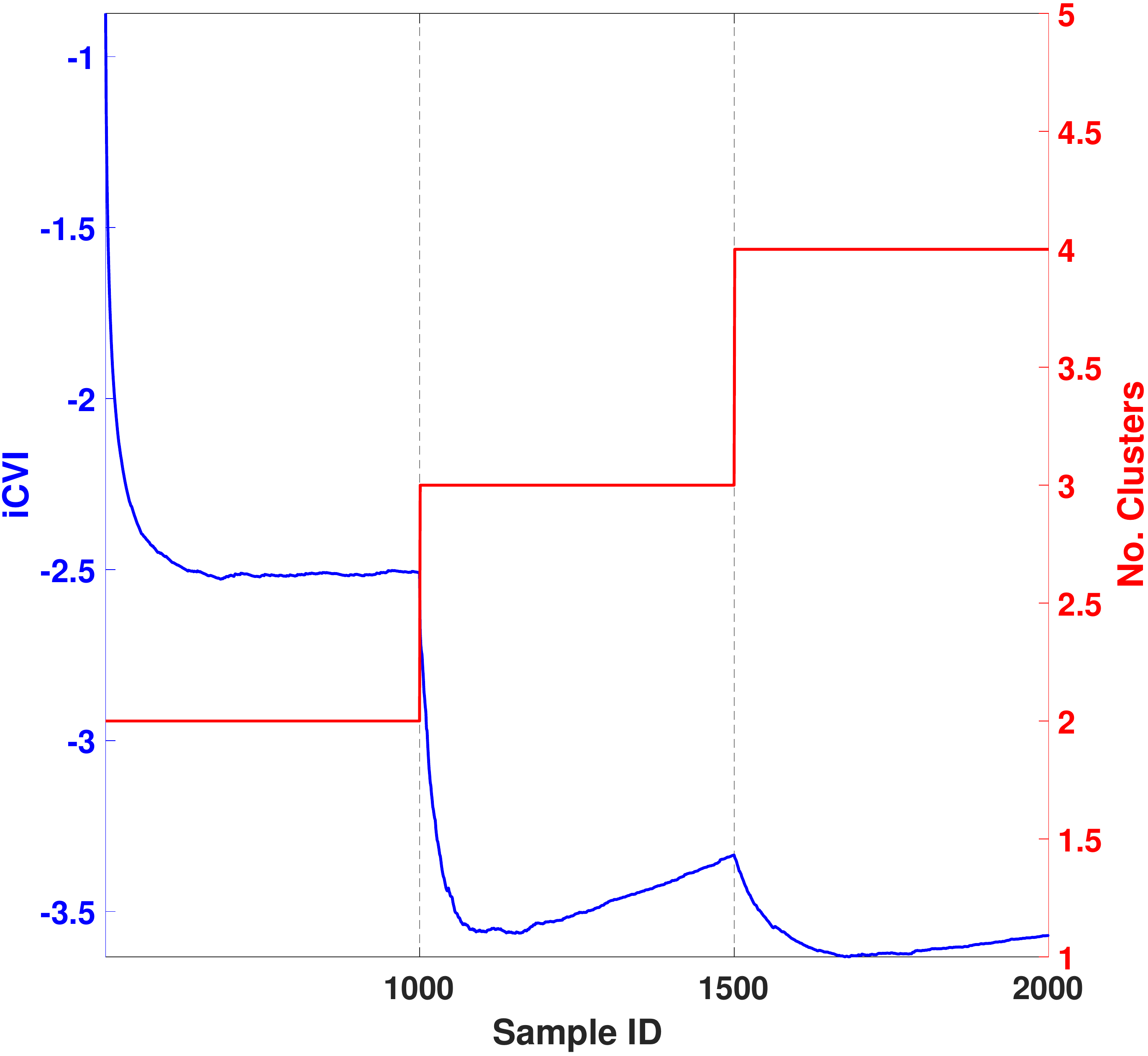}
\label{Fig:iNI3}}
}
\centerline{
\subfloat[iXB]{\includegraphics[width=\szxb\columnwidth]{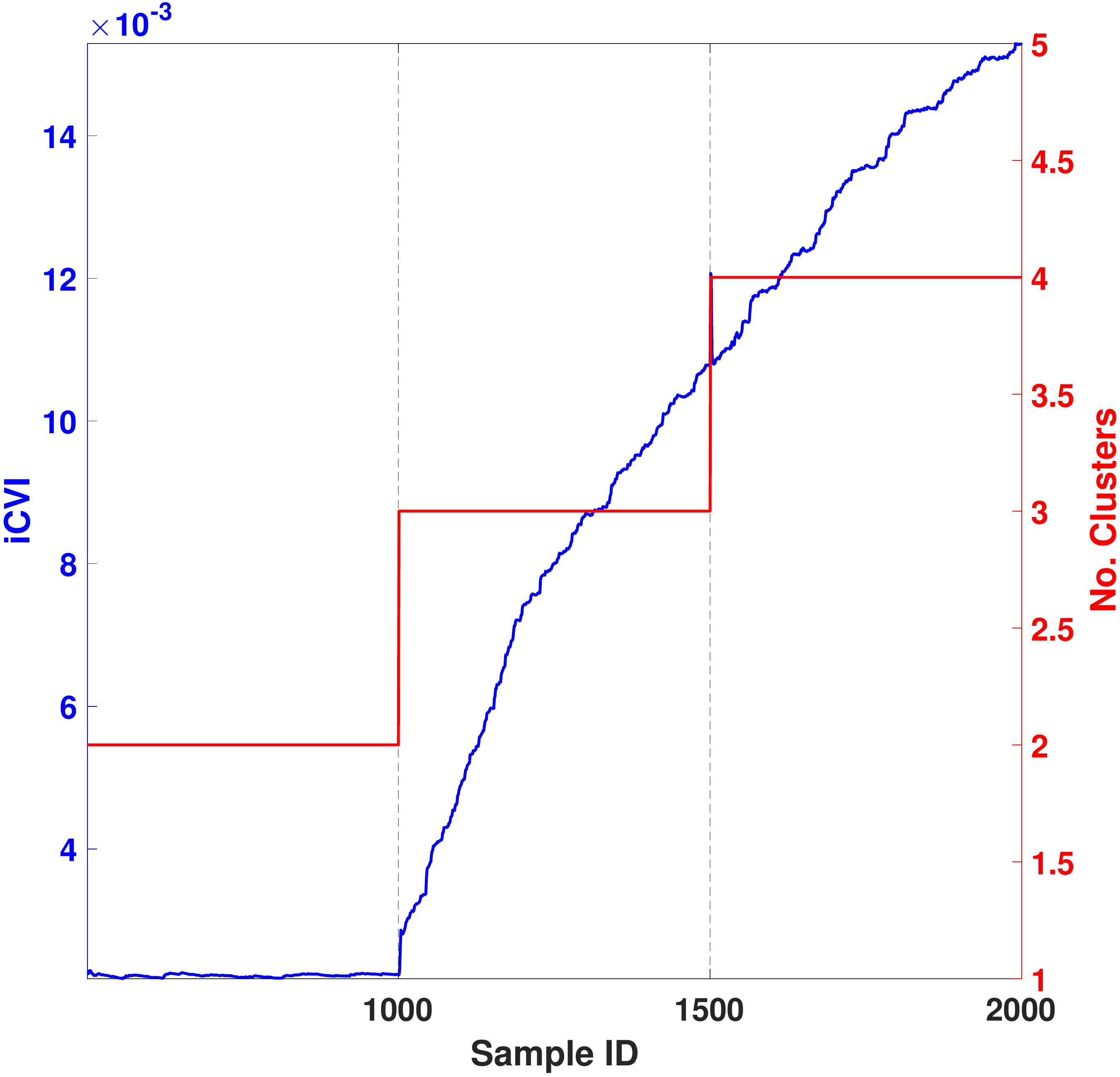}
\label{Fig:iXB3}}
\hfil
\subfloat[iDB]{\includegraphics[width=\szxb\columnwidth]{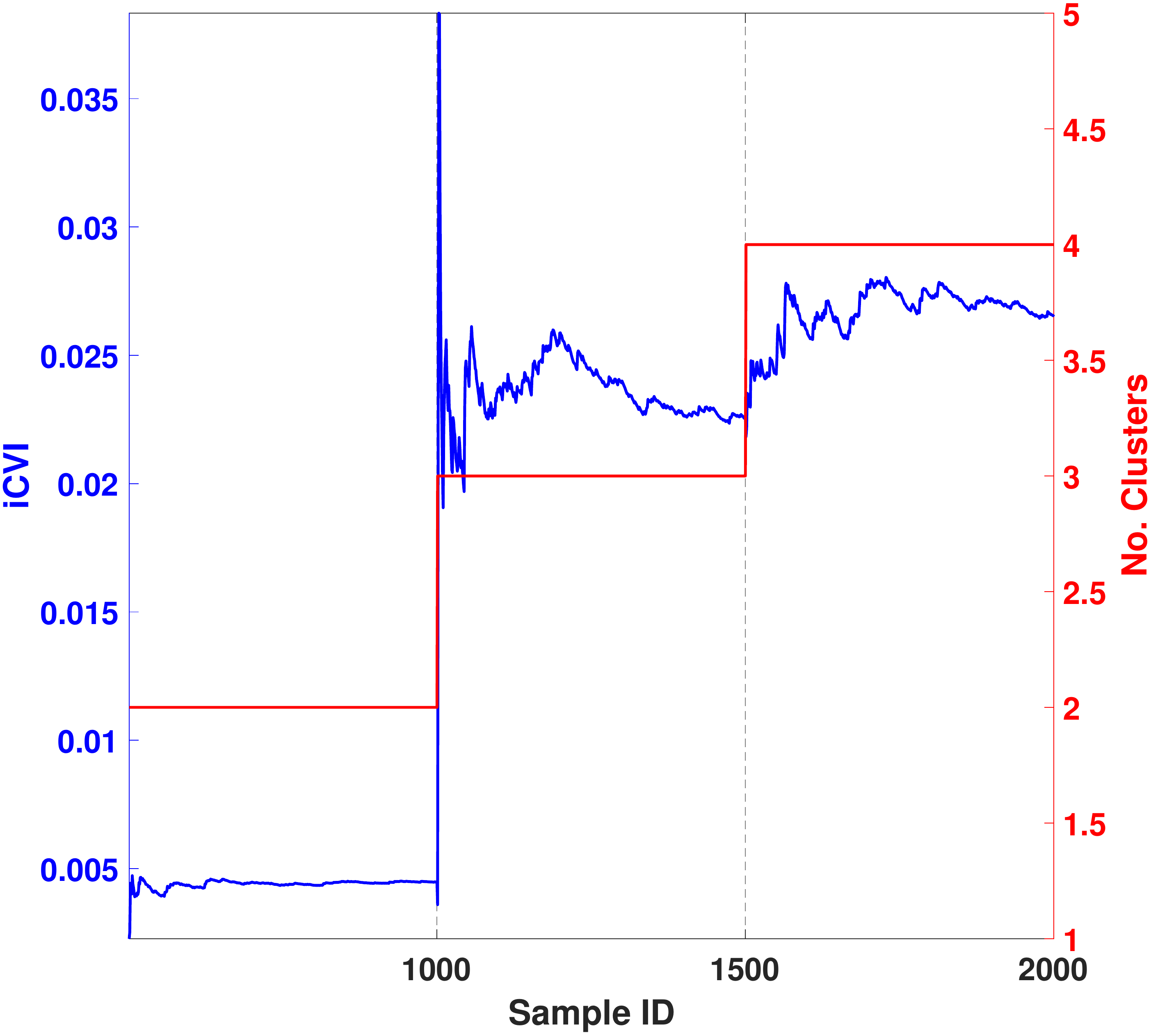}
\label{Fig:iDB3}}
\hfil
\subfloat[PS]{\includegraphics[width=\szxb\columnwidth]{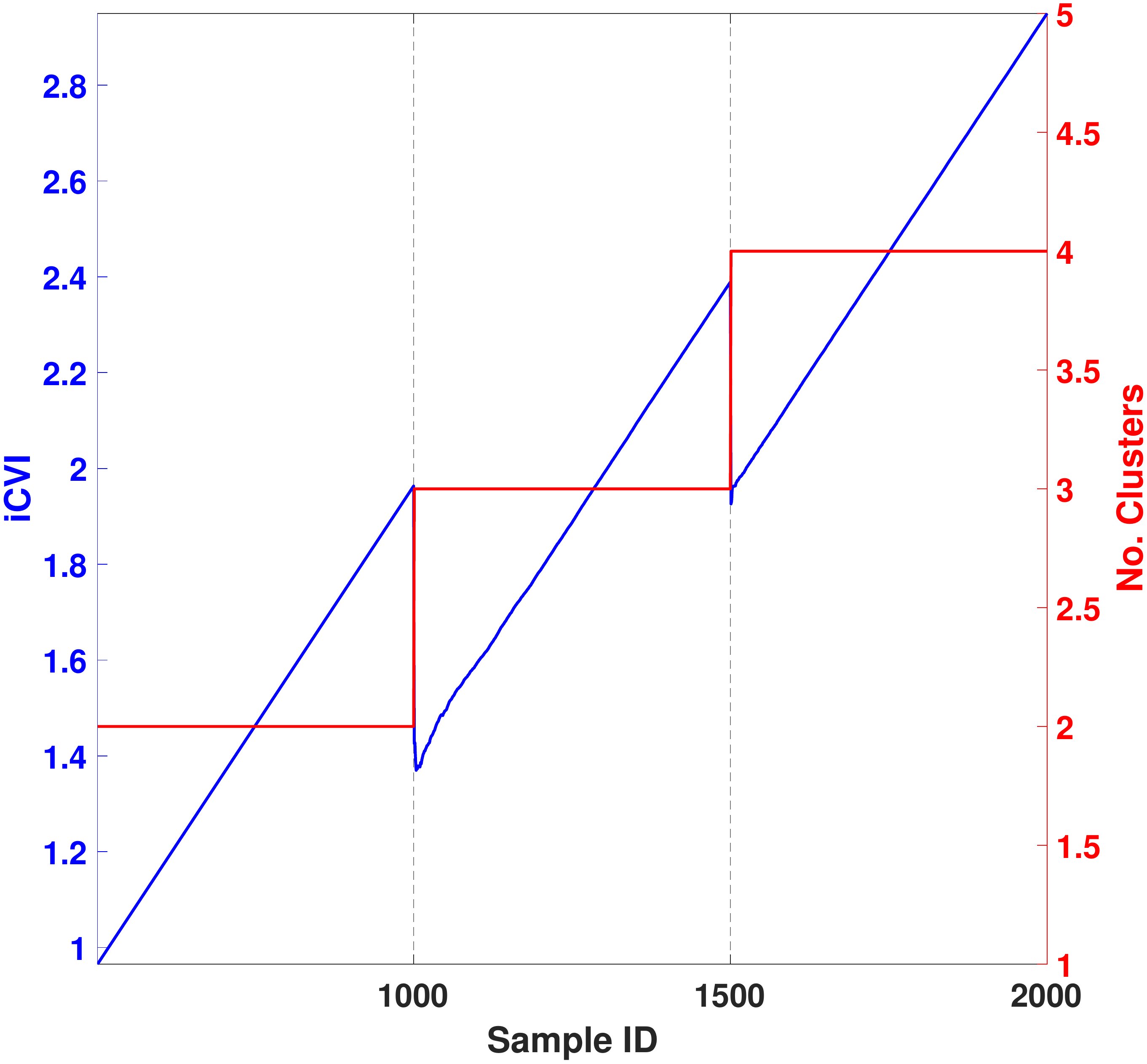}
\label{Fig:PS3}}
}
\caption{(a) A high-quality partitioning of the \textit{D4} data set by fuzzy ART-based clustering algorithms ($ARI = 1.0$, $\rho=0.69$). (b) Fuzzy SMART's module A categories ($\rho_A = 0.9$) and CONNvis~\cite{tasdemir2009} (thicker and darker lines indicate stronger connections). (c)-(l) Behavior of the iCVIs (blue curve) for the partition in (a). The number of clusters is tracked by the step-like red curve. The dashed vertical lines represent the limits between two consecutive clusters (ground truth), \ie~samples before a line belong to one cluster whereas samples after it belong to another.}
\label{Fig:D4HQ}
\end{figure}
\begin{figure}[!hp]
\centerline{
\subfloat[Data partition ]{\includegraphics[width=\szC\columnwidth]{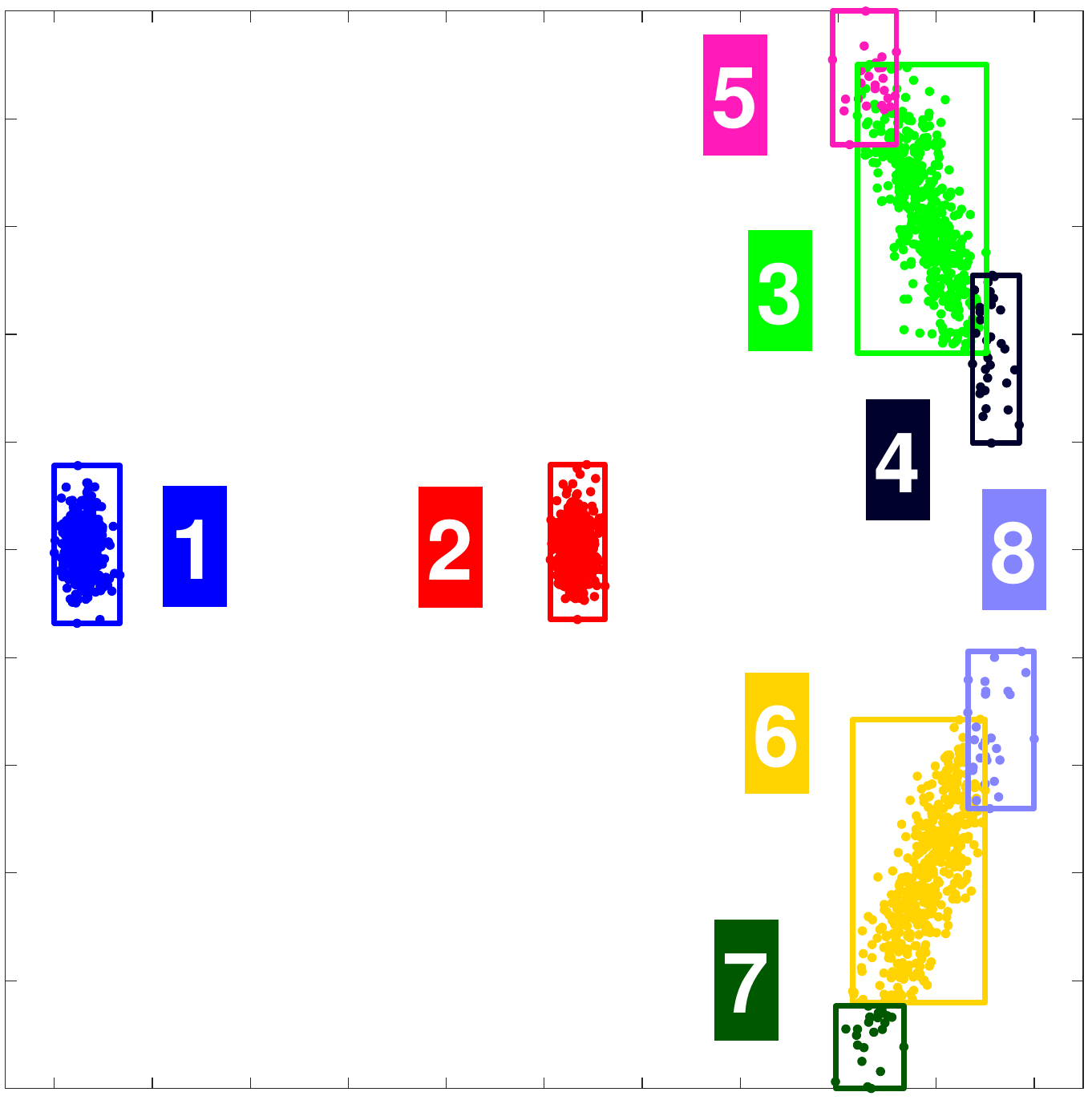}
\label{Fig:partition4}}
\hfil
\subfloat[Fuzzy SMART's module A connectivity]{\includegraphics[width=\szP\columnwidth]{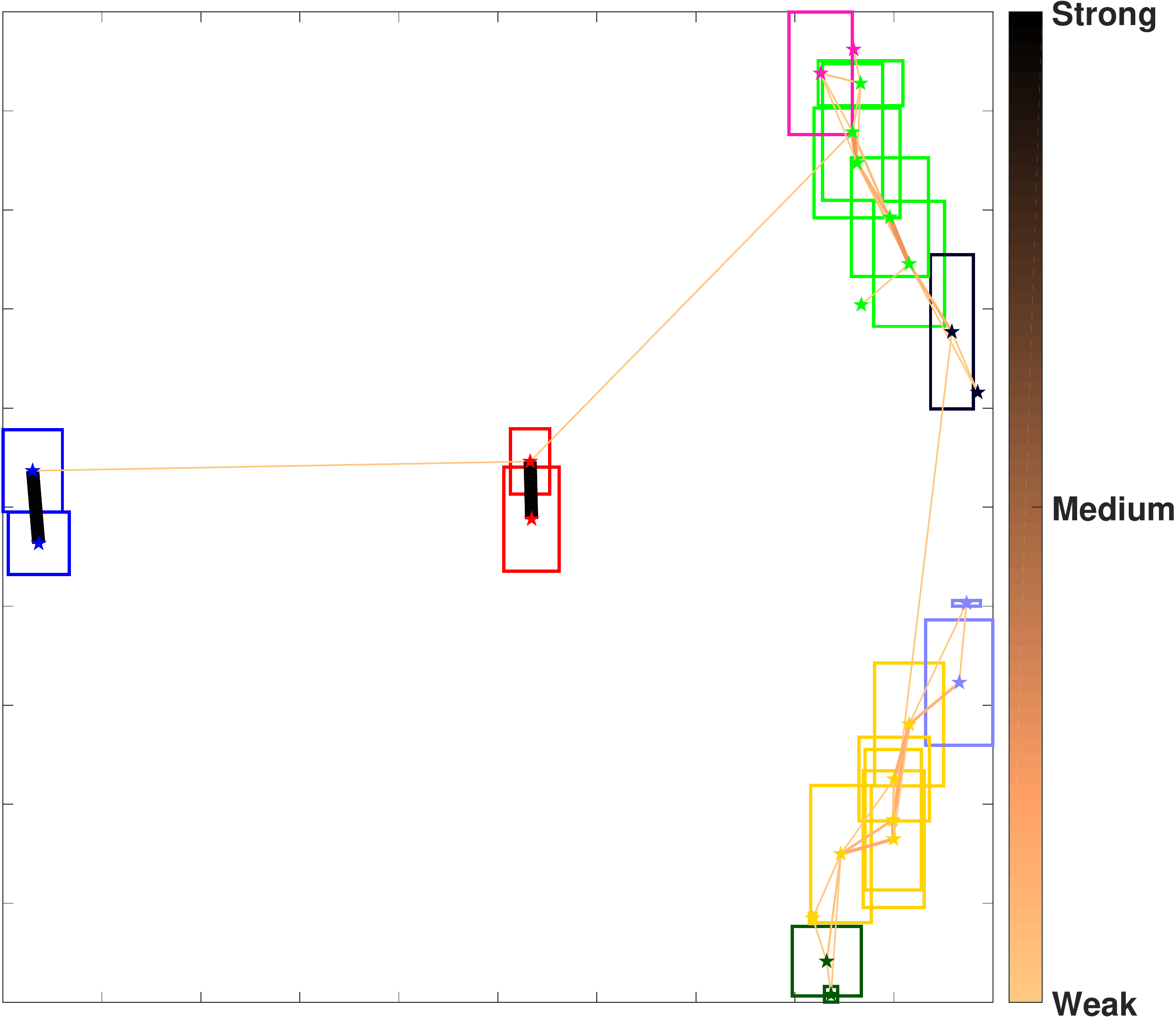}
\label{Fig:SMART_A_4}}
\hfil
\subfloat[iConn\_Index]{\includegraphics[width=\szxb\columnwidth]{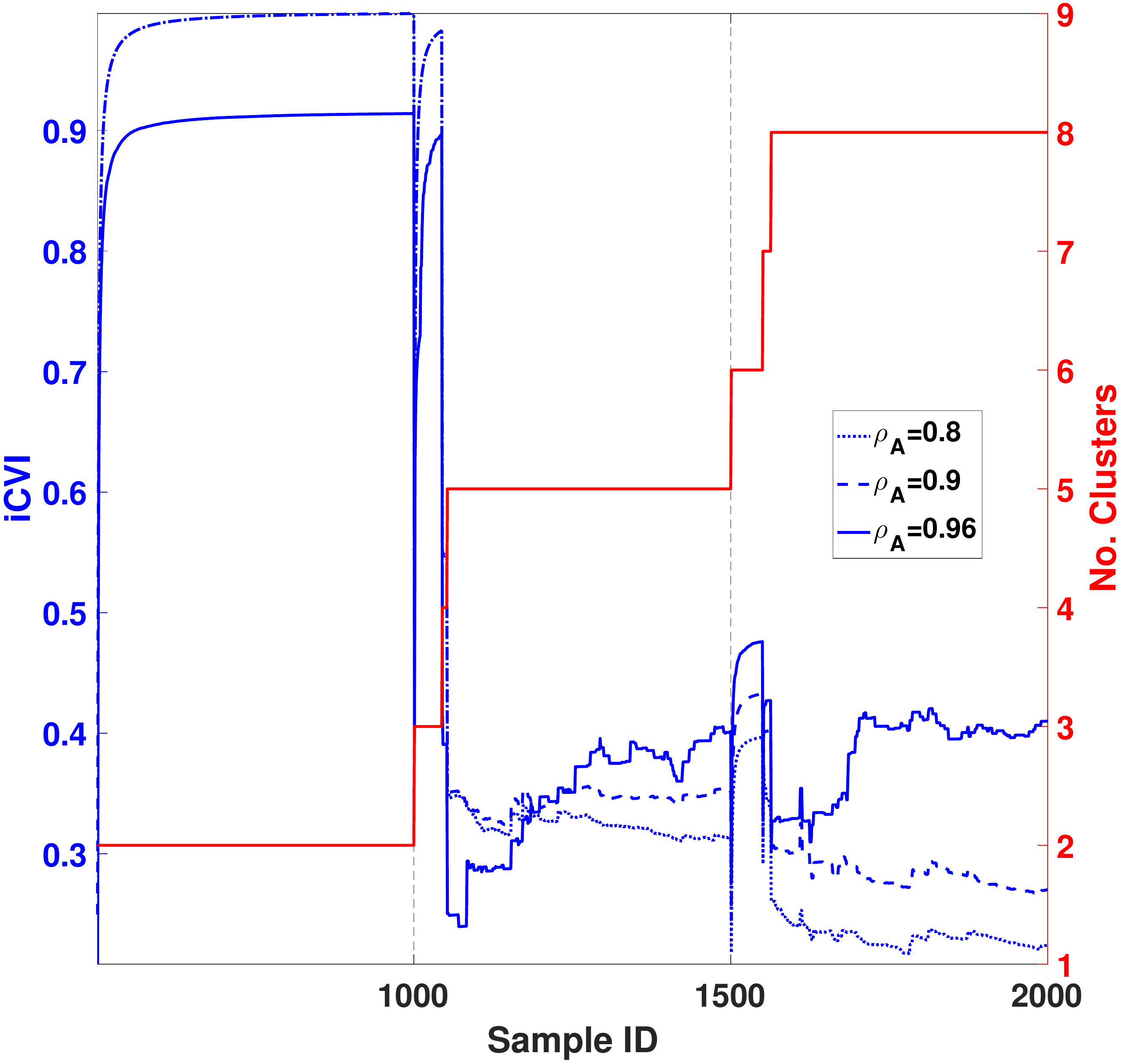}
\label{Fig:iConn4}}
}
\centerline{
\subfloat[iCH]{\includegraphics[width=\szxb\columnwidth]{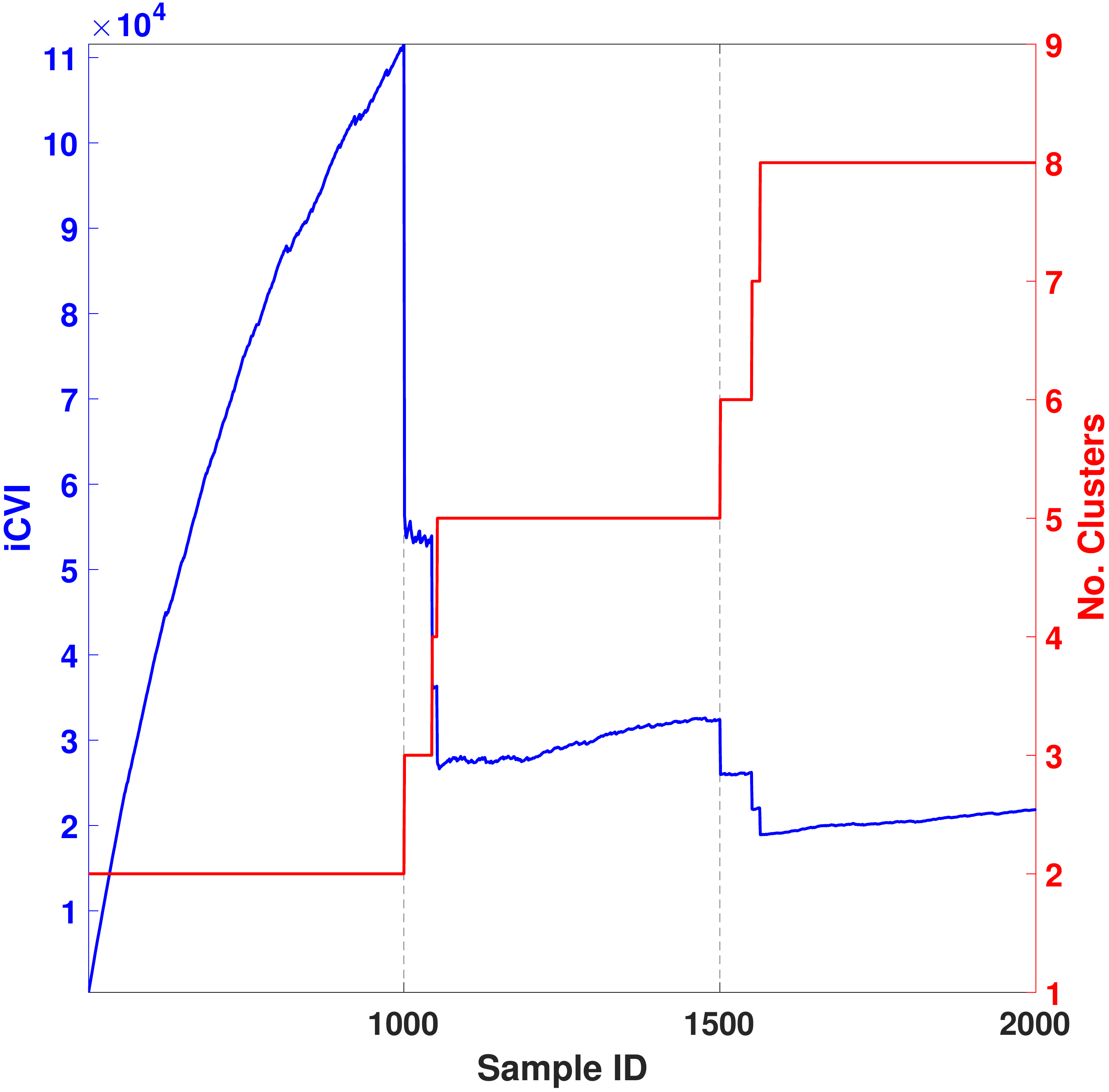}
\label{Fig:iCH4}}
\hfil
\subfloat[iSIL]{\includegraphics[width=\szxb\columnwidth]{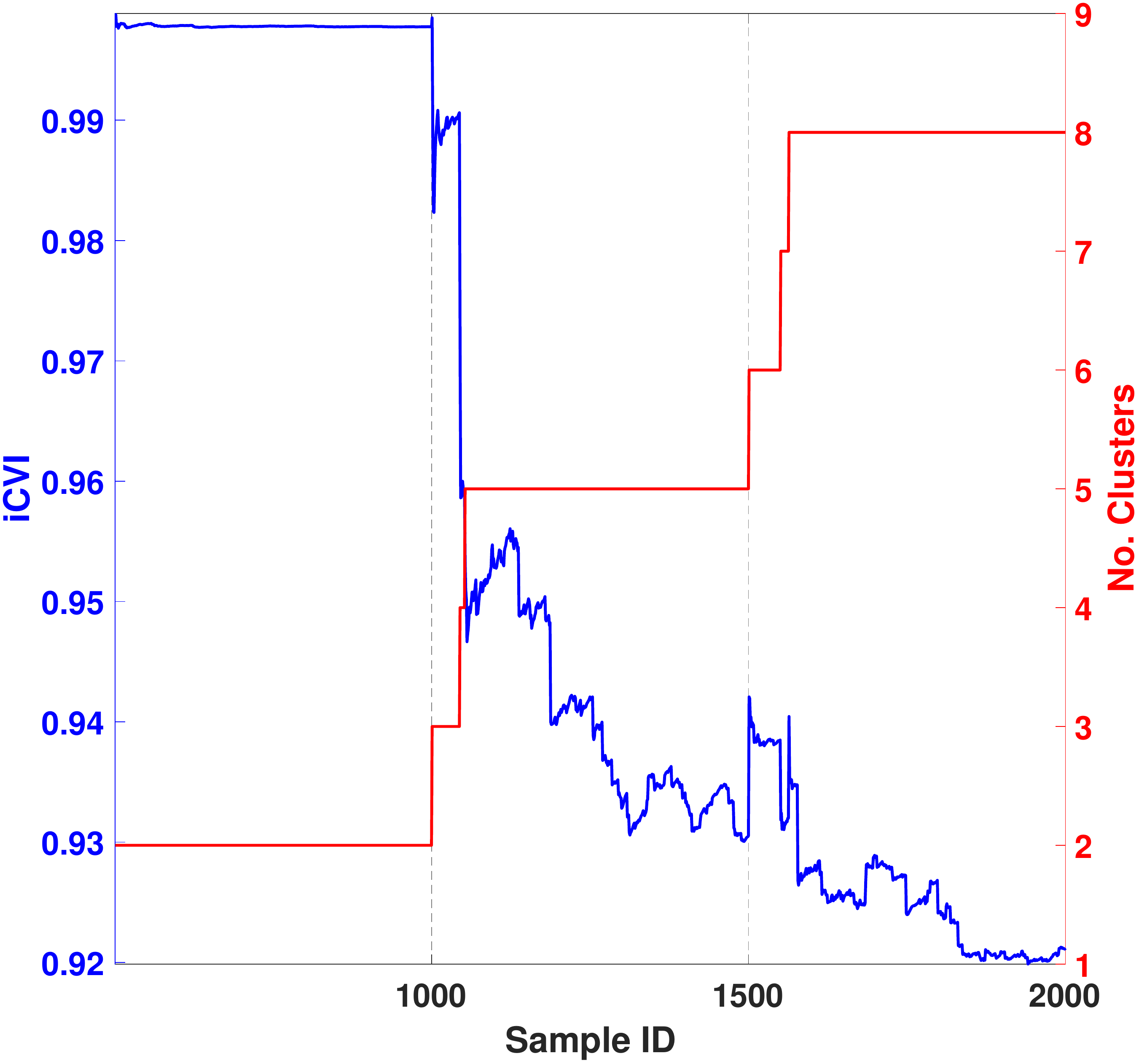}
\label{Fig:iSIL4}}
\hfil
\subfloat[iPBM]{\includegraphics[width=\szxb\columnwidth]{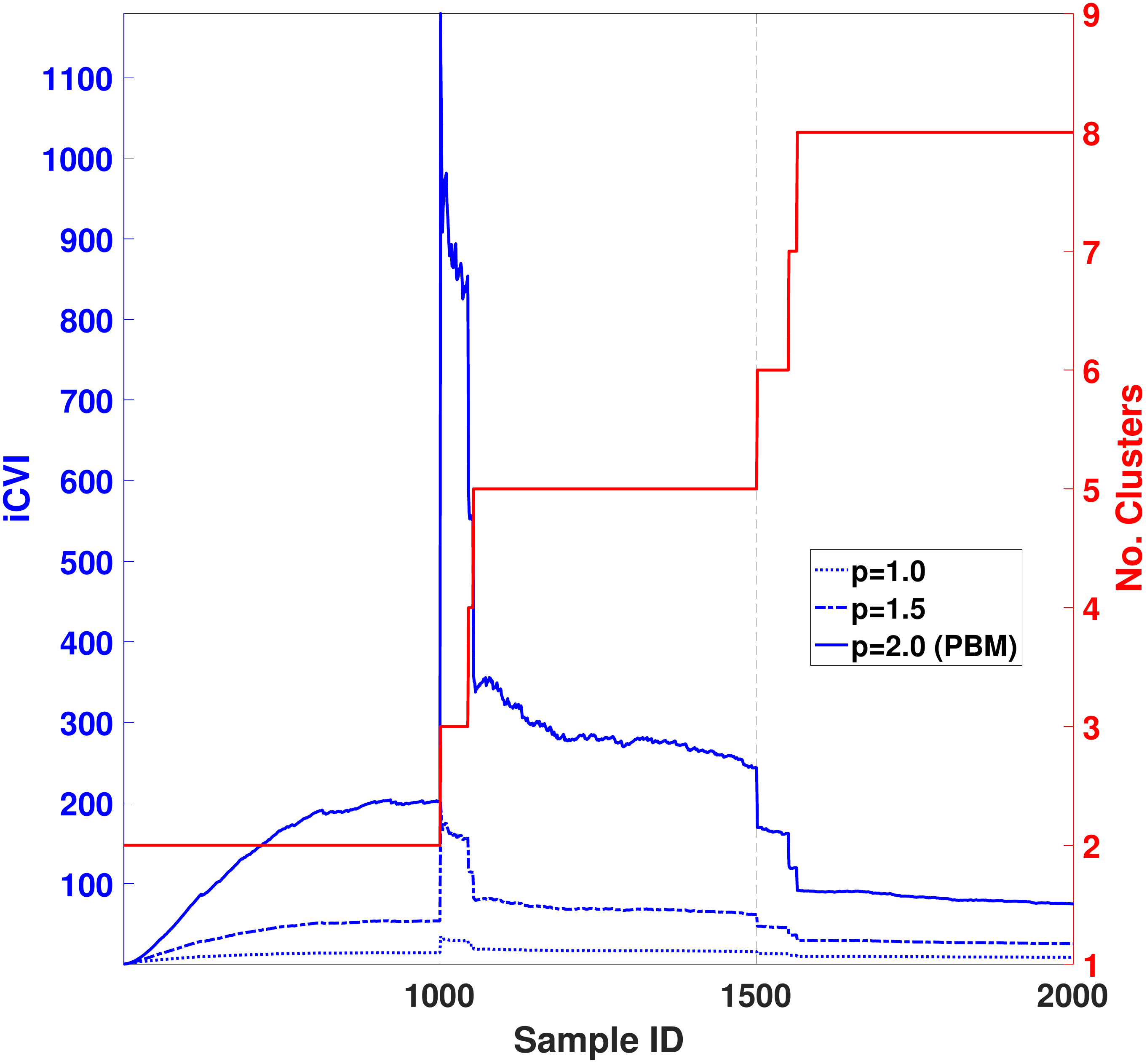}
\label{Fig:iPBM4}}
}
\centerline{
\subfloat[irCIP]{\includegraphics[width=\szxb\columnwidth]{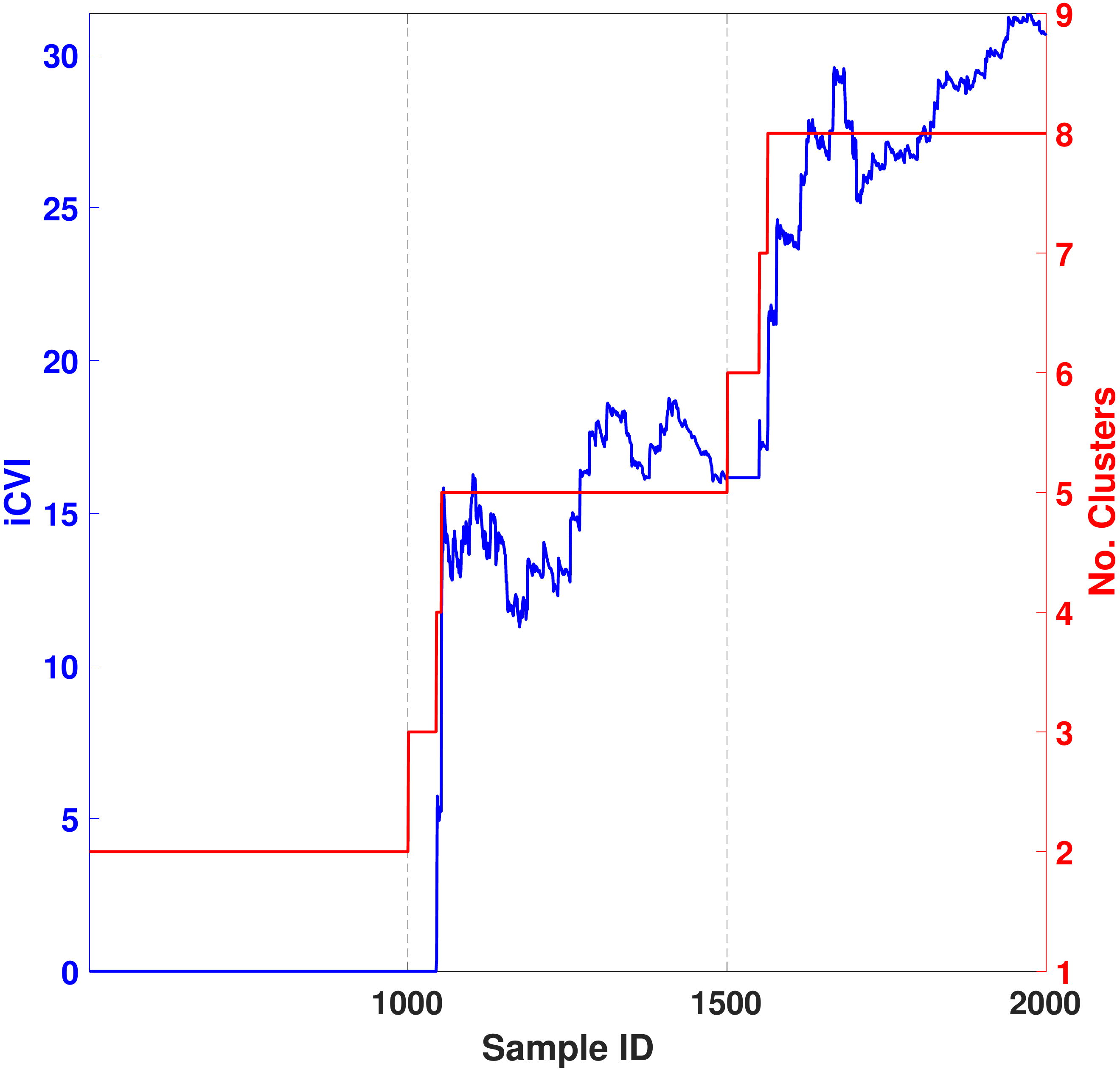}
\label{Fig:irCIP4}}
\hfil
\subfloat[irH]{\includegraphics[width=\szxb\columnwidth]{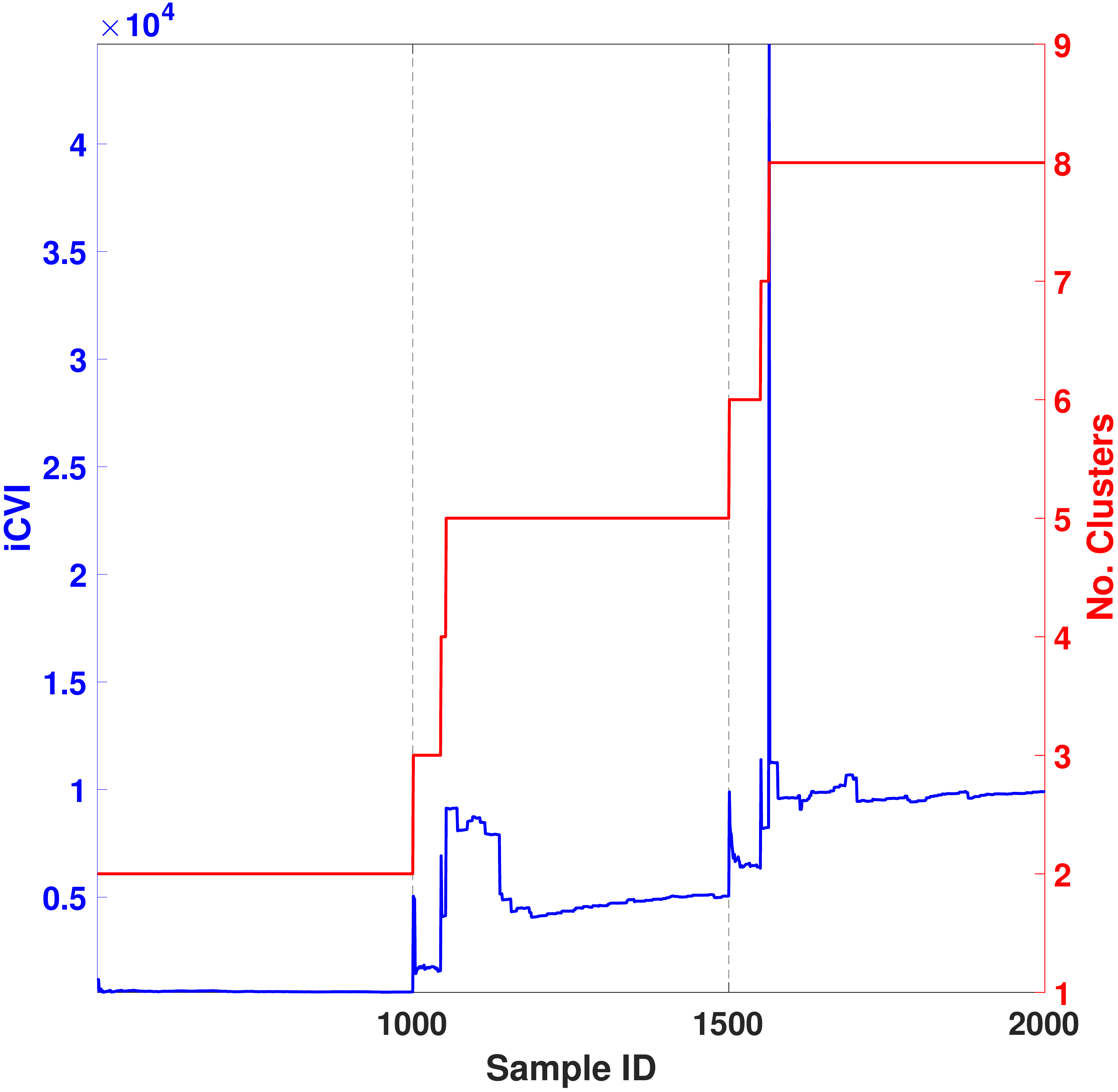}
\label{Fig:irH4}}
\hfil
\subfloat[iNI]{\includegraphics[width=\szxb\columnwidth]{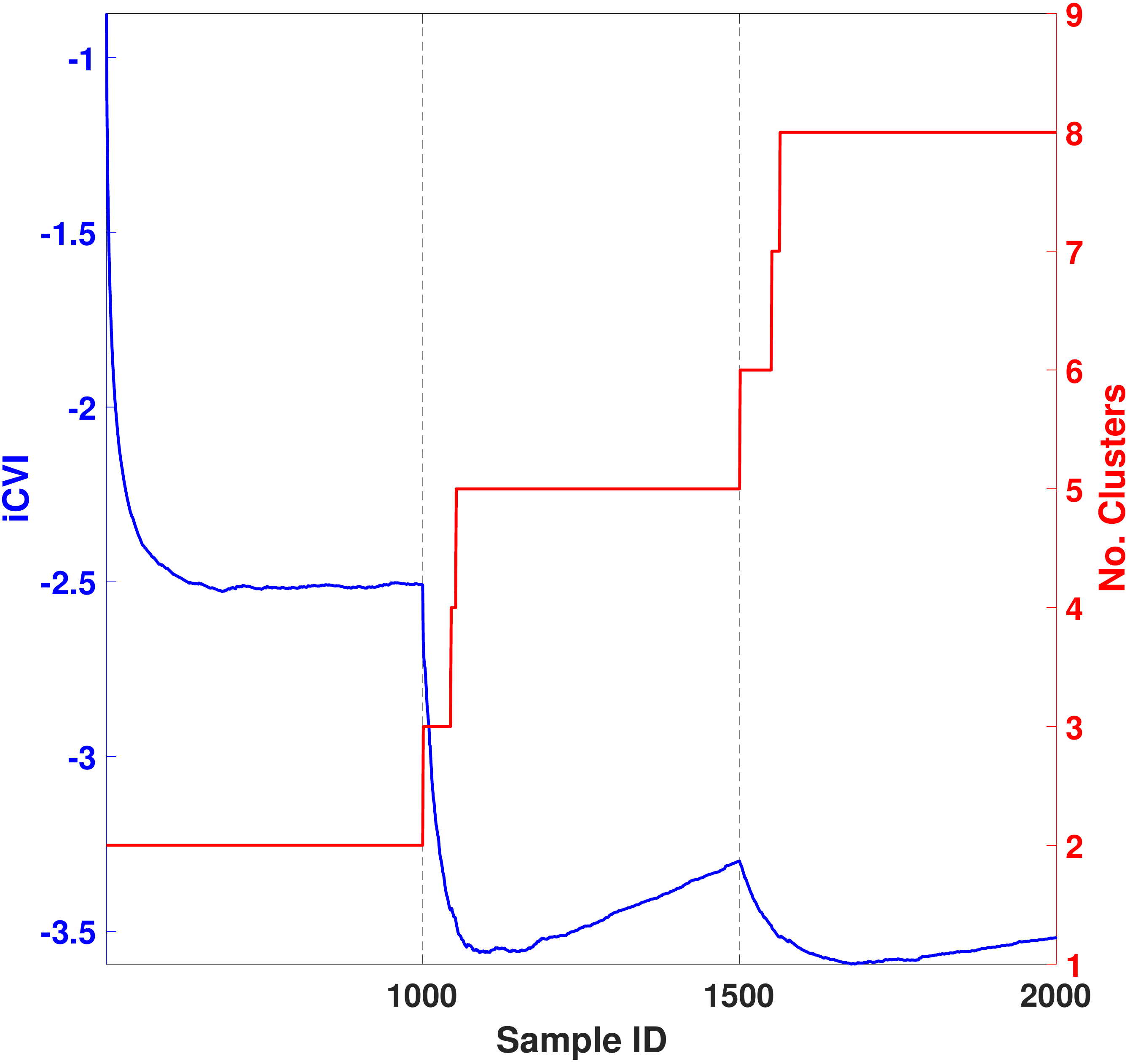}
\label{Fig:iNI4}}
}
\centerline{
\subfloat[iXB]{\includegraphics[width=\szxb\columnwidth]{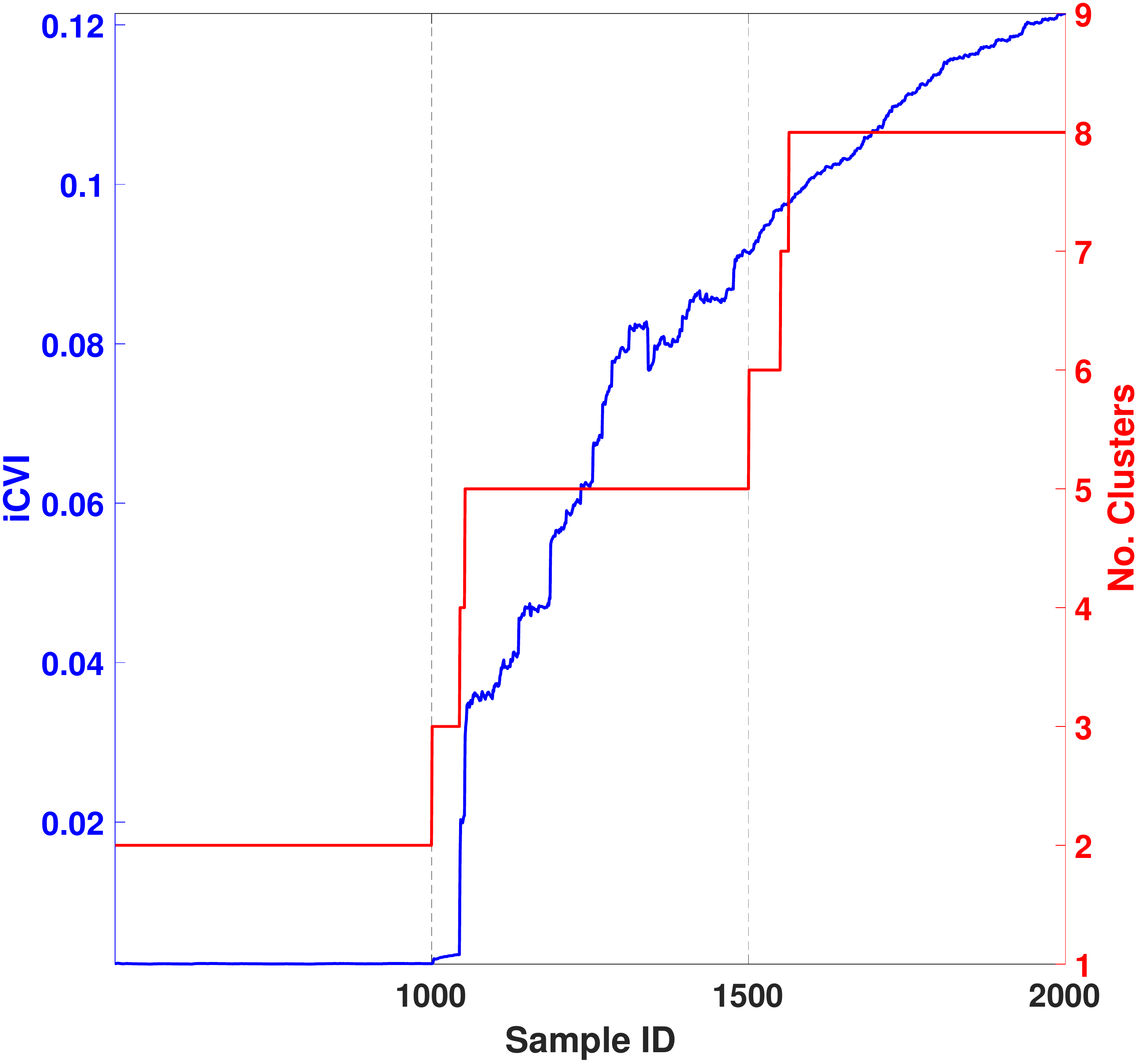}
\label{Fig:iXB4}}
\hfil
\subfloat[iDB]{\includegraphics[width=\szxb\columnwidth]{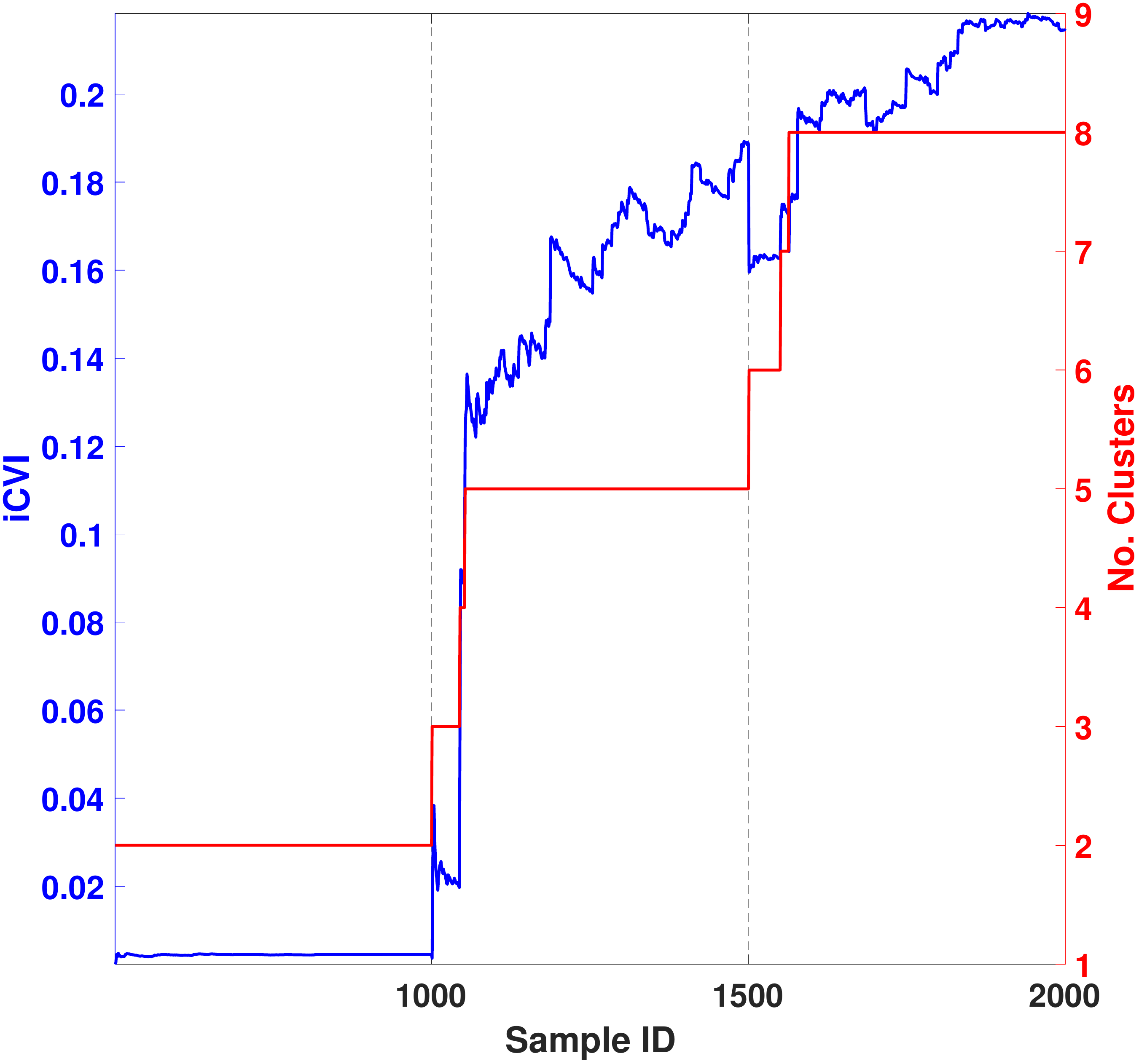}
\label{Fig:iDB4}}
\hfil
\subfloat[PS]{\includegraphics[width=\szxb\columnwidth]{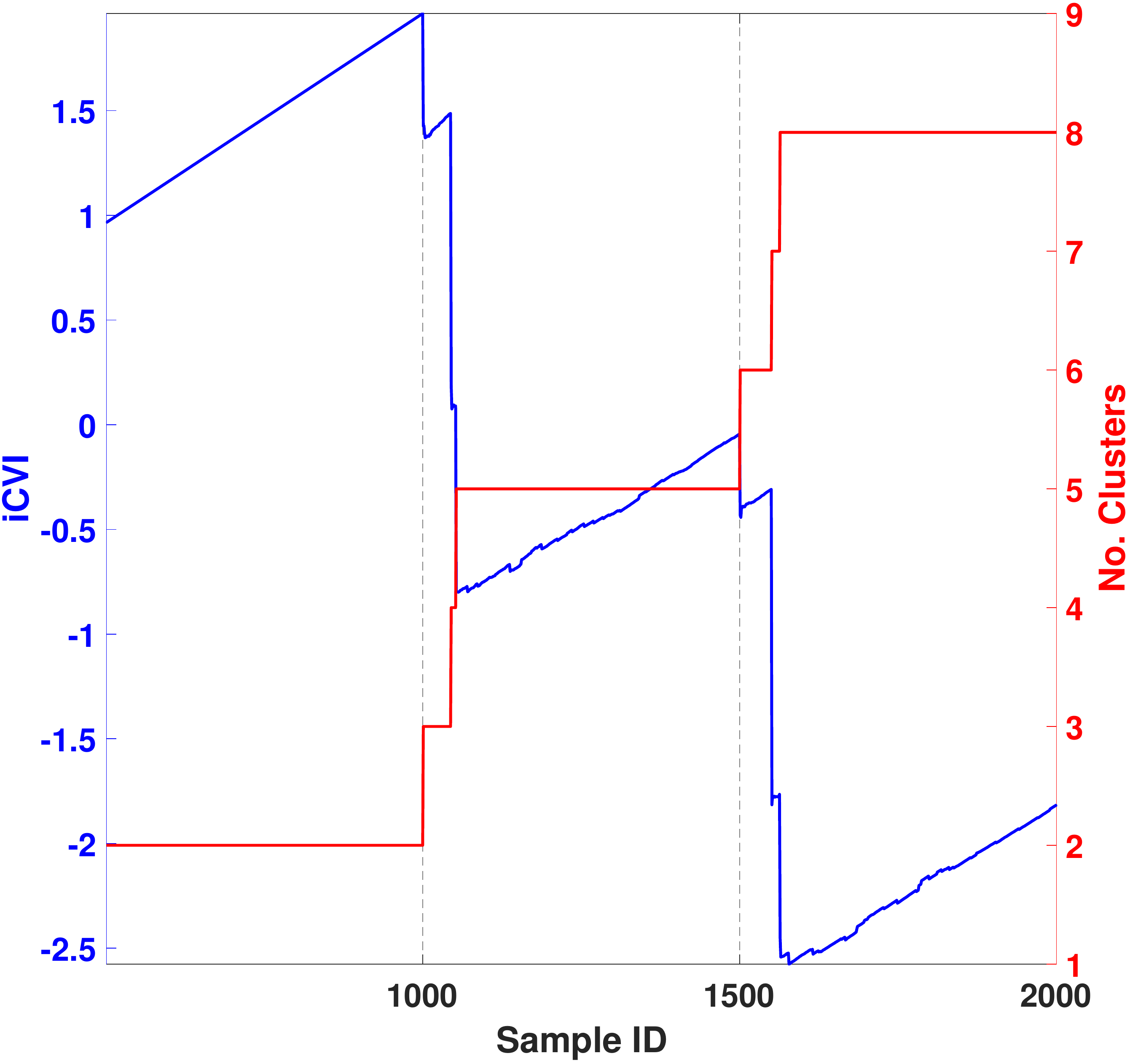}
\label{Fig:PS4}}
}
\caption{(a) An over-partition of the \textit{D4} data set by fuzzy ART-based clustering algorithms ($ARI = 0.9315$, $\rho=0.80$). (b) Fuzzy SMART's module A categories ($\rho_A = 0.9$) and CONNvis~\cite{tasdemir2009} (thicker and darker lines indicate stronger connections). (c)-(l) Behavior of the iCVIs (blue curve) for the partition in (a). The number of clusters is tracked by the step-like red curve. The dashed vertical lines represent the limits between two consecutive clusters (ground truth), \ie~samples before a line belong to one cluster whereas samples after it belong to another.}
\label{Fig:D4OP}
\end{figure}
\newcommand{\iconnsize}{.71}
\begin{figure}[!hp]
\centerline{
\subfloat[Conn\_Index and iConn\_Index behaviors ($\rho_A=0.92$)] {\includegraphics[width=\iconnsize\columnwidth] {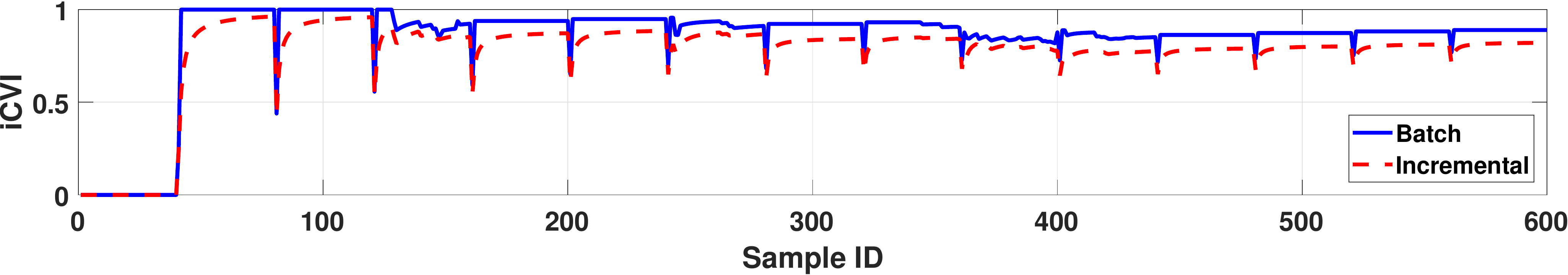}}
}
\centerline{
\subfloat[$Error = Conn\_Index - iConn\_Index $] {\includegraphics[width=\iconnsize\columnwidth] {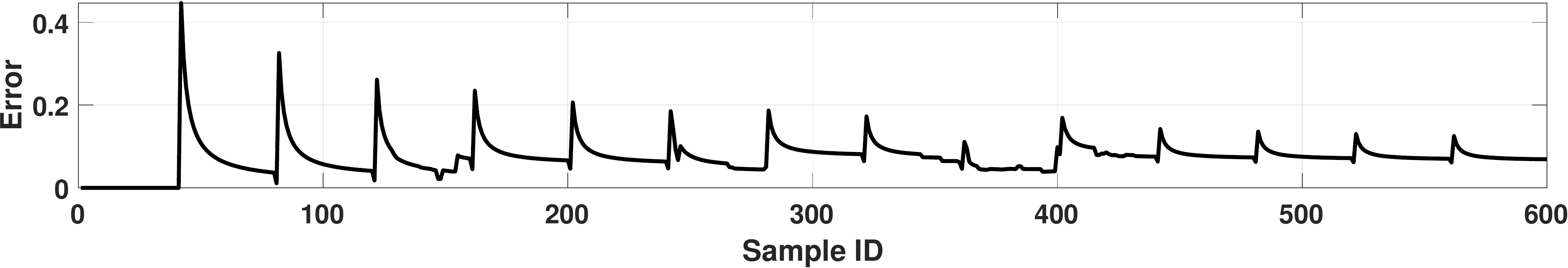}}
}
\centerline{
\subfloat[Correlation coefficient and MSE] {\includegraphics[width=\iconnsize\columnwidth] {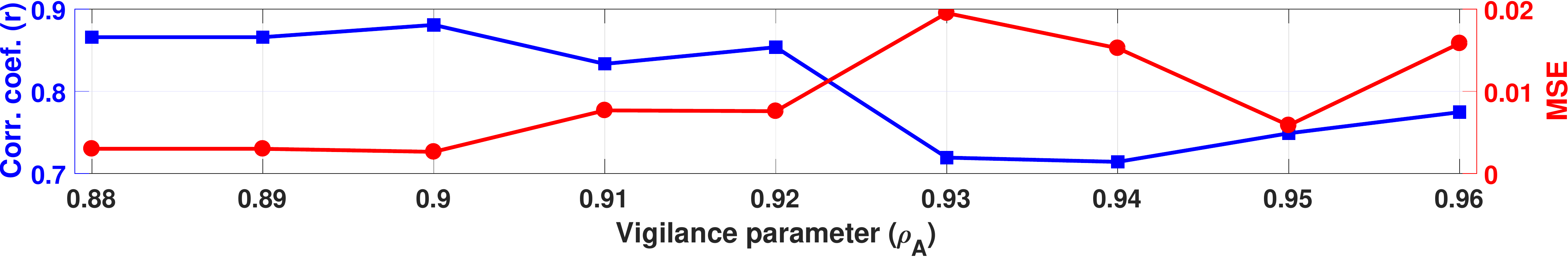} \label{Fig:rcoef1}}
}
\caption{(a) Behaviors of Conn\_Index (continuous blue line) and iConn\_Index (dashed red line) for the high-quality partitioning of the \textit{R15} data set (Fig.~\ref{Fig:R15HQ}). (b) Error between the batch and incremental versions in (a). (c)~Correlation coefficients and MSE between the batch and incremental versions for fuzzy SMART's module A vigilance parameter $\rho_A \in [\rho_B, 0.96]$.}
\label{Fig:iConn_R15_A}
\end{figure}
\begin{figure}[!hp]
\centerline{
\subfloat[Conn\_Index and iConn\_Index behaviors ($\rho_A=0.92$)] {\includegraphics[width=\iconnsize\columnwidth] {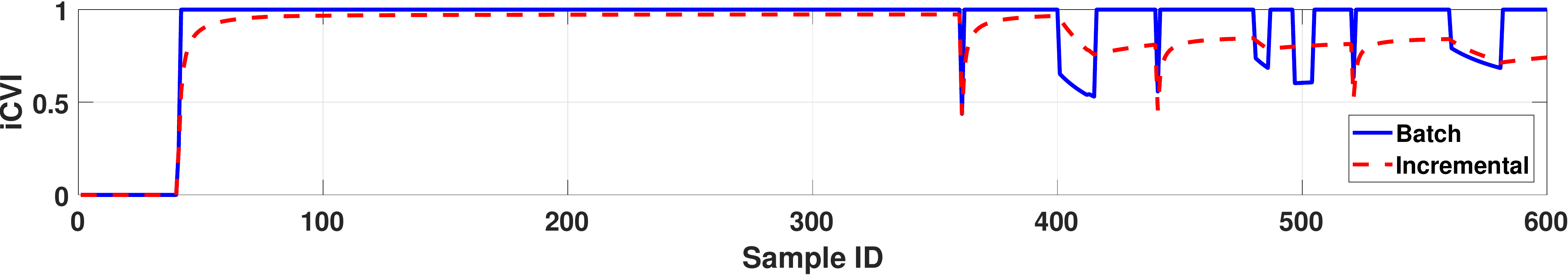}}
}
\centerline{
\subfloat[$Error = Conn\_Index - iConn\_Index$] {\includegraphics[width=\iconnsize\columnwidth] {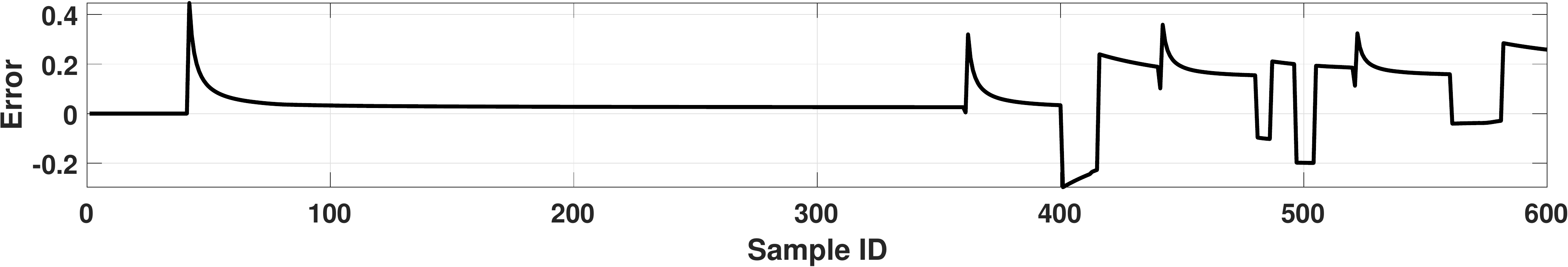}}
}
\centerline{
\subfloat[Correlation coefficient and MSE] {\includegraphics[width=\iconnsize\columnwidth] {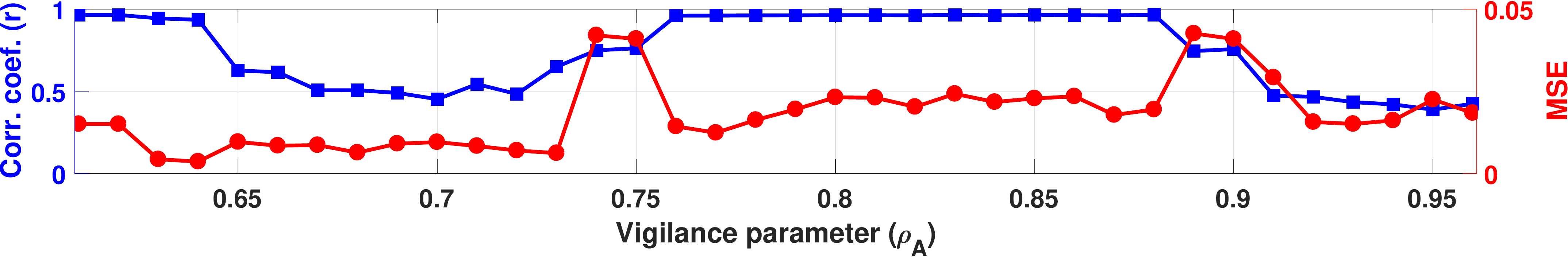} \label{Fig:rcoef2}}
}
\caption{(a) Behaviors of Conn\_Index (continuous blue line) and iConn\_Index (dashed red line) for the under-partitioning of the \textit{R15} data set (Fig.~\ref{Fig:R15UP}). (b) Error between the batch and incremental versions in (a). (c)~Correlation coefficients and MSE between the batch and incremental versions for fuzzy SMART's module A vigilance parameter $\rho_A \in [\rho_B, 0.96]$.}
\label{Fig:iConn_R15_B}
\end{figure}
\begin{figure}[!hp]
\centerline{
\subfloat[Conn\_Index and iConn\_Index behaviors ($\rho_A=0.92$)] {\includegraphics[width=\iconnsize\columnwidth] {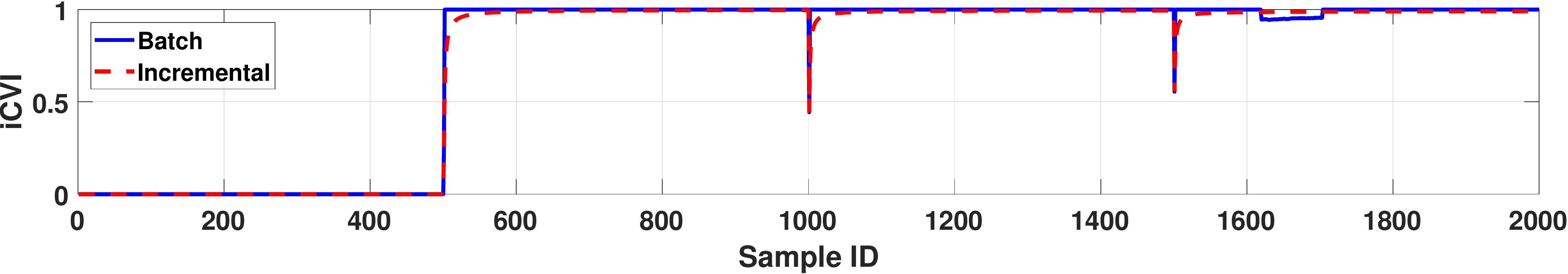}}
}
\centerline{
\subfloat[$Error = Conn\_Index - iConn\_Index$] {\includegraphics[width=\iconnsize\columnwidth] {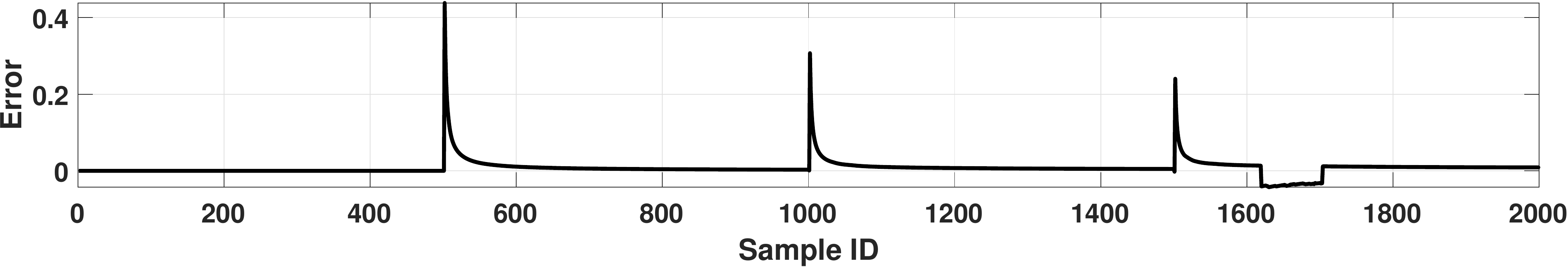}}
}
\centerline{
\subfloat[Correlation coefficient and MSE] {\includegraphics[width=\iconnsize\columnwidth] {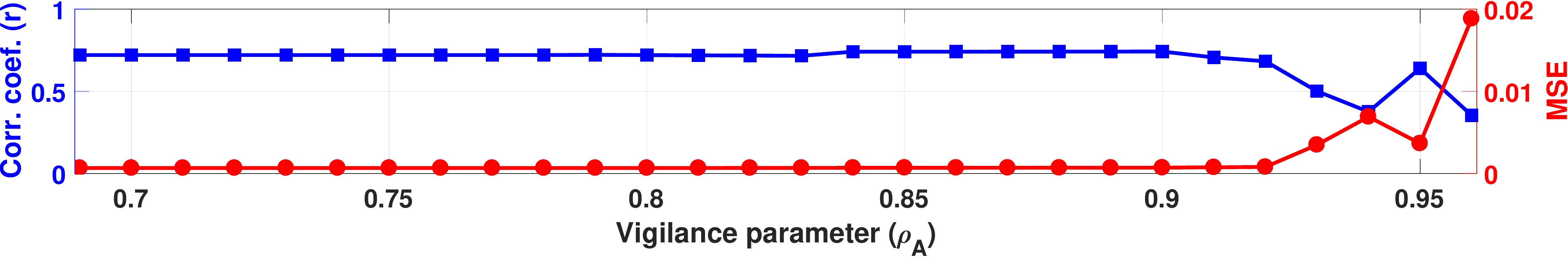} \label{Fig:rcoef3}}
}
\caption{(a) Behaviors of Conn\_Index (continuous blue line) and iConn\_Index (dashed red line) for the high-quality partitioning of the \textit{D4} data set (Fig.~\ref{Fig:D4HQ}). (b) Error between the batch and incremental versions in (a). (c)~Correlation coefficients and MSE between the batch and incremental versions for fuzzy SMART's module A vigilance parameter $\rho_A \in [\rho_B, 0.96]$.}
\label{Fig:iConn_D4_A}
\end{figure}
\begin{figure}[!hp]
\centerline{
\subfloat[Conn\_Index and iConn\_Index behaviors ($\rho_A=0.92$)] {\includegraphics[width=\iconnsize\columnwidth] {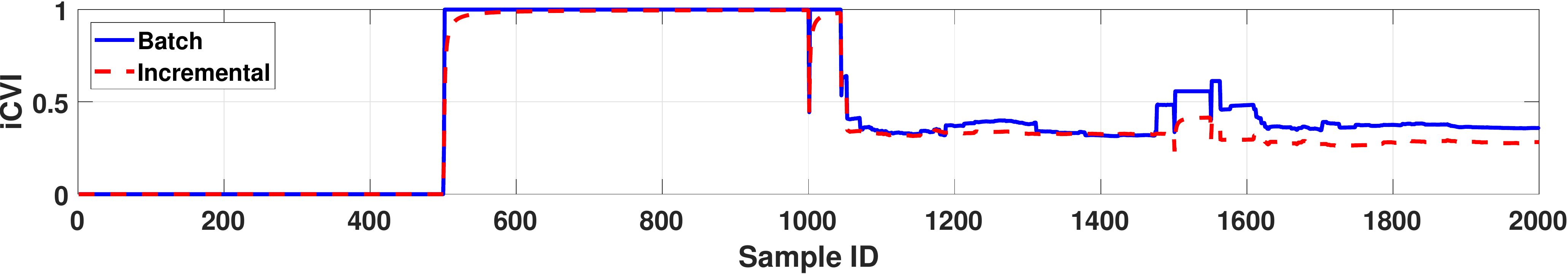}}
}
\centerline{
\subfloat[$Error = Conn\_Index - iConn\_Index$] {\includegraphics[width=\iconnsize\columnwidth] {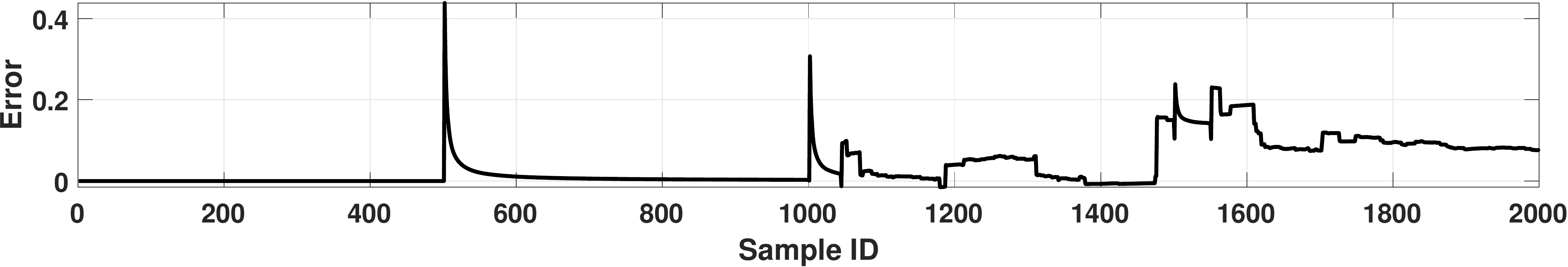}}
}
\centerline{
\subfloat[Correlation coefficient and MSE] {\includegraphics[width=\iconnsize\columnwidth] {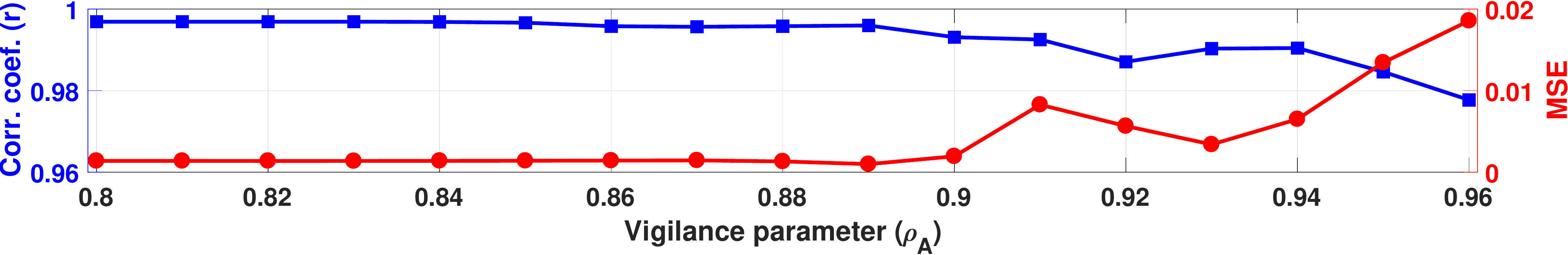} \label{Fig:rcoef4}}
}
\caption{(a) Behaviors of Conn\_Index (continuous blue line) and iConn\_Index (dashed red line) for the over-partitioning of the \textit{D4} data set (Fig.~\ref{Fig:D4OP}). (b) Error between the batch and incremental versions in (a). (c)~Correlation coefficients and MSE between the batch and incremental versions for fuzzy SMART's module A vigilance parameter $\rho_A \in [\rho_B, 0.96]$.}
\label{Fig:iConn_D4_B}
\end{figure}

Except for the iConn\_Index, none of the iCVIs provided distinctive insights on the over-partition problem: there is a noticeable decrease of iConn\_Index values (due to a large increase of $Inter\_Conn$ and decrease of $Intra\_Conn$), especially considering that this iCVI's value is bounded to the interval $[0,1]$. More importantly, following the over-partition, it does not exhibit the general behavior previously observed in Figs.~\ref{Fig:iConn1} and~\ref{Fig:iConn3}, and it maintains its poor assessment of the clustering solution, thus indicating that there is an issue with the partition found by the clustering algorithm. 

\section{Incremental versus batch implementations}  \label{Sec:inc_vs_batch}

When evaluated over time, most iCVIs discussed in this study yield the same values as their batch counterparts (\eg~the the recursive formulation of compactness is an exact computation, not an approximation~\cite{Moshtaghi2018, Moshtaghi2018b}). The only exception is the iConn\_Index, which is the subject of analysis of this section. Figs.~\ref{Fig:iConn_R15_A} to~\ref{Fig:iConn_D4_B} illustrate the evolution of both Conn\_Index and iConn\_Index for all four experiments described in Section~\ref{Sec:results}. These figures also show the error (difference) between the batch and incremental implementations of the Conn\_Index after the presentation of each sample. To obtain the batch Conn\_Index values, fuzzy SMART was set to evaluation mode and all first and second winning prototypes were recomputed after the presentation of each sample. 

Notably, error spikes consistently occur on the appearance of new clusters. In general, the error gradually diminishes over time, as samples within a given cluster are continuously presented to the system. These trends are particularly clear when fuzzy SMART yield high quality partitions (Figs.~\ref{Fig:iConn_R15_A} and~\ref{Fig:iConn_D4_A}). Regarding the cases of under- and over-partitioning (Figs.~\ref{Fig:iConn_R15_B} and~\ref{Fig:iConn_D4_B}), the errors are more pronounced. However, iConn\_Index still smoothly follows the overall trends of its batch counterpart (which has a more jagged behavior).

Finally, the effect of fuzzy SMART module A's quantization level on the similarity of the batch and incremental implementations was investigated. This was done by varying its vigilance parameter $\rho_A$ in the closed interval $[\rho_B, 0.96]$ (larger values of $\rho_A$ produce finer granularity of cluster prototypes). The Pearson correlation coefficients~\cite{Bain1992} and the mean squared error (MSE) depicted in Figs.~\ref{Fig:rcoef1}, \ref{Fig:rcoef2}, \ref{Fig:rcoef3}, and~\ref{Fig:rcoef4} show that 
the behavior of iConn\_Index is consistent with Conn\_Index across wide ranges of fuzzy SMART module A's vigilance. Interestingly, their dissimilarity tends to increase with very large vigilance values. These results support the original assumption, stated in Section~\ref{Sec:iConn}, that both versions of the Conn\_Index would behave similarly. Therefore, iConn\_Index is suitable for monitoring the performance online clustering methods.

\section{Conclusion}  \label{Sec:conclusion}

This paper extended six cluster validity indices (CVIs) to incremental versions, namely, incremental Calinski-Harabasz (iCH), incremental I index and incremental Pakhira-Bandyopadhyay-Maulik (iI and iPBM), incremental Silhouette (iSIL), incremental Negentropy Increment (iNI), incremental Representative Cross Information Potential (irCIP) and Cross Entropy (irH), and incremental Conn\_Index (iConn\_Index). Furthermore, using fuzzy adaptive resonance theory (ART)-based clustering algorithms, three different scenarios were analyzed: detection of the correct number of clusters in high-quality partitions, under- and over-partitioning. In such scenarios, a comparative study was performed among the presented incremental cluster validity indices (iCVIs), the Partition Separation (PS) index, the incremental Xie-Beni (iXB), and the incremental Davies-Bouldin~(iDB). 

As expected from previous studies, most iCVIs undergo abrupt changes following the creation of a new cluster. When samples from the same cluster are presented, however, each iCVI exhibits a particular behavior, which was taken as a reference to compare the cases of under- and over-partitioning a data set. In these experiments, the least visually informative iCVIs (\ie~that provided less useful visual cues/hints in their behavior) were irCIP and iXB. Particularly, most iCVIs detected under-partitioning in at least one stage of the incremental clustering process, whereas only the iConn\_Index provided some insight to indicate over-partitioning problems. Nonetheless, the iConn\_Index failed in identifying one of the under-partitioning cases. Therefore, the usual recommendation regarding batch CVIs also applies to iCVIs: this research highlights the importance of monitoring a number of iCVI dynamics at any given time, rather than relying on the assessment of only one. Finally, it was shown that, although not equal to its batch counterpart, the iConn\_Index follows the same general trends. It is expected that the observations from the study presented here will assist in incremental clustering applications such as data streams.

\section*{Acknowledgment}

This research was sponsored by the Missouri University of Science and Technology Mary K. Finley Endowment and Intelligent Systems Center; the Coordena\c{c}\~{a}o de Aperfei\c{c}oamento de Pessoal de N\'{i}vel Superior - Brazil (CAPES) - Finance code BEX 13494/13-9; the U.S. Dept. of Education Graduate Assistance in Areas of National Need program; and the Army Research Laboratory (ARL), and it was accomplished under Cooperative Agreement Number W911NF-18-2-0260. The views and conclusions contained in this document are those of the authors and should not be interpreted as representing the official policies, either expressed or implied, of the Army Research Laboratory or the U.S. Government. The U.S. Government is authorized to reproduce and distribute reprints for Government purposes notwithstanding any copyright notation herein. The authors would also like to thank Prof. James M. Keller and his coauthors for providing an early copy of reference~\cite{Keller2018}.

\bibliographystyle{IEEEtran}
\bibliography{IEEEabrv,bib/references}

\end{document}